\documentclass{article} %
\usepackage{iclr2025_conference,times}

\usepackage{arydshln}

\usepackage{subcaption}

\usepackage{graphicx} 
\usepackage{algorithm}
\usepackage{algorithmic}

\usepackage[utf8]{inputenc} %
\usepackage[T1]{fontenc}    %
\usepackage{hyperref}       %
\usepackage{url}            %
\usepackage{booktabs}       %
\usepackage{amsfonts}       %
\usepackage{nicefrac}       %
\usepackage{microtype}      %
\usepackage{xcolor}         %

\usepackage{amsmath}
\usepackage{amssymb}

\usepackage{cleveref}

\newcommand{\E}{\mathbb{E}}

\newcommand{\ba}[1]{\begin{align}#1\end{align}}

\newcommand{\cdotv}{\boldsymbol{\cdot}}

\usepackage{stackengine}

\makeatletter
\newcommand{\distas}[1]{\mathbin{\overset{#1}{\kern\z@\sim}}}%

\newcommand{\cL}{\mathcal{L}}

\newcommand{\cN}{\mathcal{N}}

\newcommand{\beqs}{\vspace{0mm}\begin{eqnarray}}
\newcommand{\eeqs}{\vspace{0mm}\end{eqnarray}}
\newcommand{\barr}{\begin{array}}
\newcommand{\earr}{\end{array}}

\def\gL{{\mathcal{L}}}

\usepackage{wrapfig}
\usepackage{multirow}

\newcommand{\cv}[0]{{\boldsymbol{c}}}

\newcommand{\xv}{\boldsymbol{x}}

\newcommand{\zv}{\boldsymbol{z}}

\newcommand{\epsilonv}{\boldsymbol{\epsilon}}

\newcommand{\thetav}{\boldsymbol{\theta}}

\newcommand{\given}{\,|\,}

\title{
Guided Score identity Distillation for Data-Free One-Step Text-to-Image Generation
}

\author{Mingyuan Zhou$^{1,2,}$\thanks{The majority of the work was done while the authors were at Google.}\,\,\,,  Zhendong Wang$^1$, Huangjie Zheng$^1$, and Hai Huang$^{3,*}$ \\ %
\\
$^1$The University of Texas at Austin, $^2$Google DeepMind, and $^3$Atlassian
}

\iclrfinalcopy %
\begin{document}

\maketitle

\begin{figure}[!h]
    \centering
    \begin{minipage}[b]{0.192\textwidth}
        \centering
        \includegraphics[width=\textwidth]{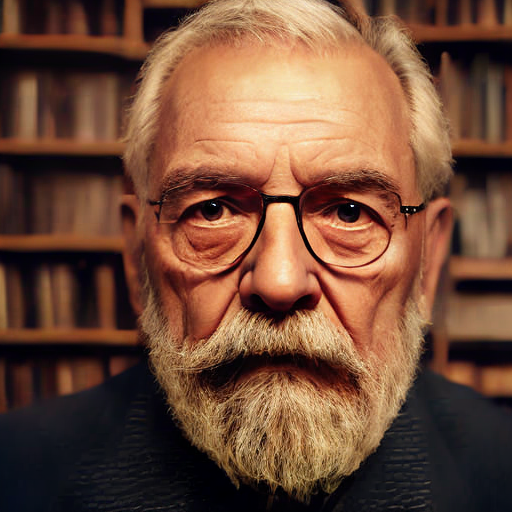}
     \\\vspace{-3mm}\caption*{\tiny distinguished older gentleman in a vintage
study, surrounded by books and dim lighting [...]
}
    \end{minipage}~
    \begin{minipage}[b]{0.192\textwidth}
        \centering
        \includegraphics[width=\textwidth]{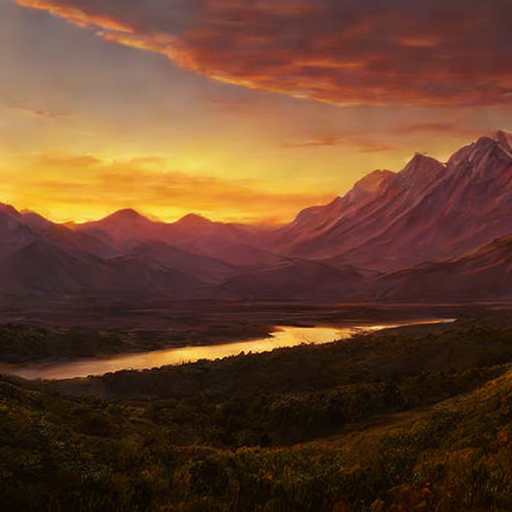}
     \\\vspace{-3mm}\caption*{\tiny saharian landscape at sunset , 4k ultra realism,
BY Anton Gorlin, trending on artstation, [...]
}
    \end{minipage}~
    \begin{minipage}[b]{0.192\textwidth}
        \centering
        \includegraphics[width=\textwidth]{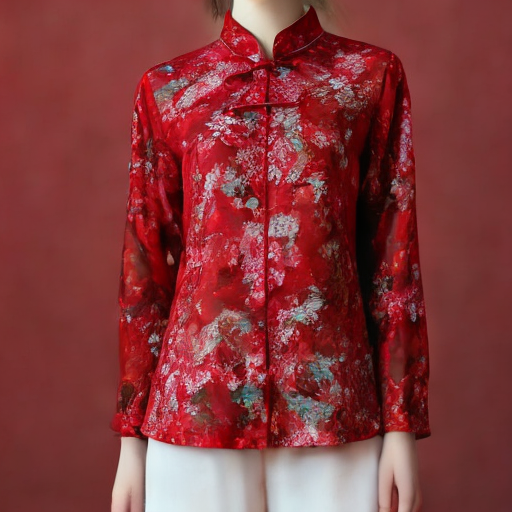}
     \\\vspace{-3mm}\caption*{\tiny chinese red blouse, in the style of dreamy and
romantic compositions, floral explosions. [...]
}
    \end{minipage}~
    \begin{minipage}[b]{0.192\textwidth}
        \centering
        \includegraphics[width=\textwidth]{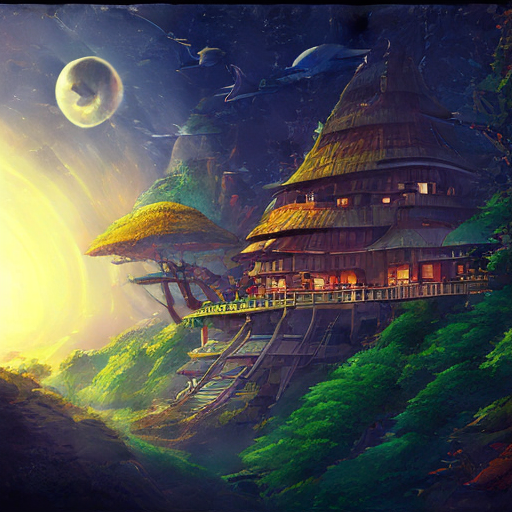}
     \\\vspace{-3mm}\caption*{\tiny futuristic simple multilayered architecture, habitation cabin in the trees, dramatic soft light [...]
}
    \end{minipage}~
    \begin{minipage}[b]{0.1931\textwidth}
        \centering
        \includegraphics[width=\textwidth]{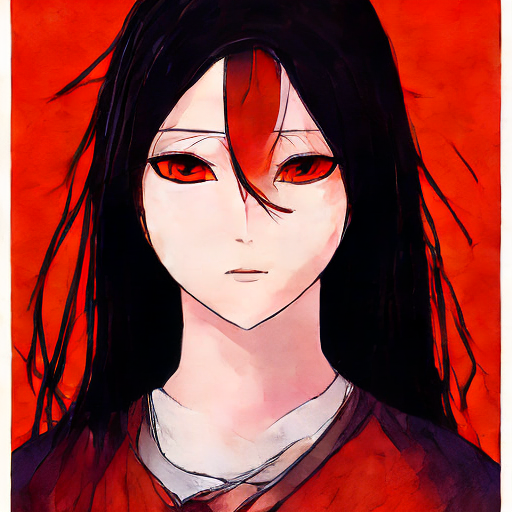}
     \\\vspace{-3mm}\caption*{\tiny poster art for the collection of the asian woman,
dynamic anime [...], mysterious realism, [...]
}
    \end{minipage}
    \\
    \begin{minipage}[b]{0.192\textwidth}
        \centering
        \includegraphics[width=\textwidth]{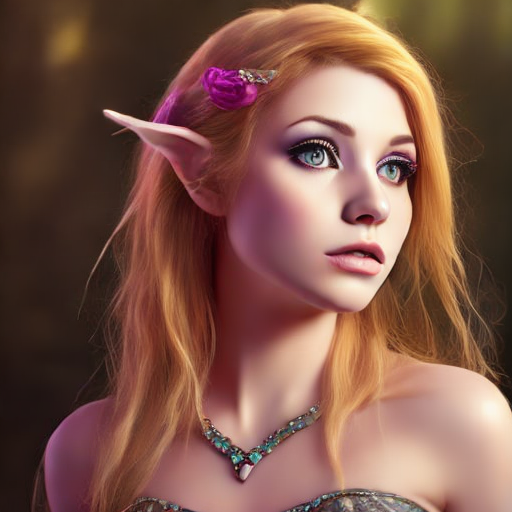}
     \\\vspace{-3mm}\caption*{\tiny A fantasy-themed portrait of a female elf with
golden hair and violet eyes, her attire  [...]
}
    \end{minipage}~
    \begin{minipage}[b]{0.192\textwidth}
        \centering
        \includegraphics[width=\textwidth]{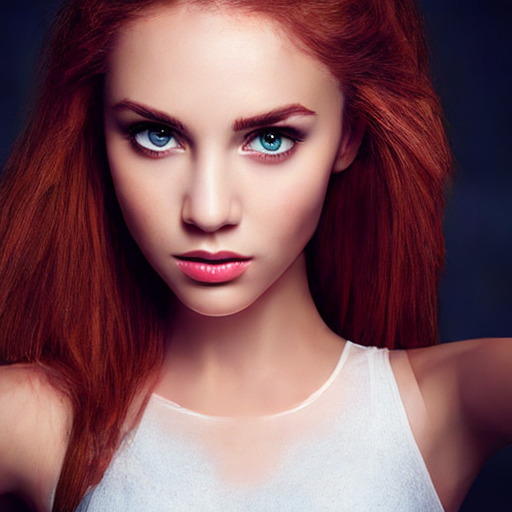}
     \\\vspace{-3mm}\caption*{\tiny 
very beautiful girl in [...], white short top,
charismatic personality, professional photo, [...]
}
    \end{minipage}~
    \begin{minipage}[b]{0.192\textwidth}
        \centering
        \includegraphics[width=\textwidth]{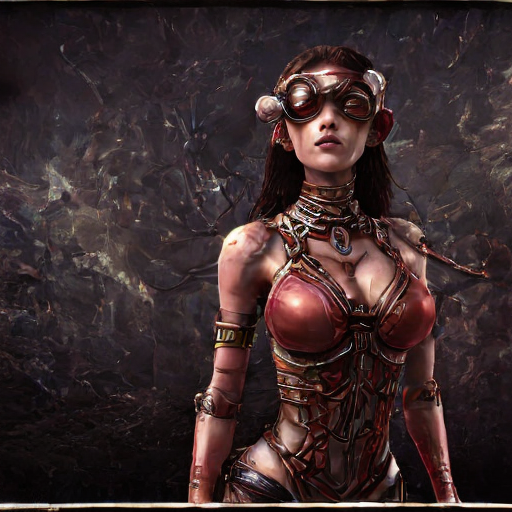}
     \\\vspace{-3mm}\caption*{\tiny 
     steampunk atmosphere, a stunning girl with a mecha musume aesthetic, adorned in [...]
}
    \end{minipage}~
    \begin{minipage}[b]{0.192\textwidth}
        \centering
        \includegraphics[width=\textwidth]{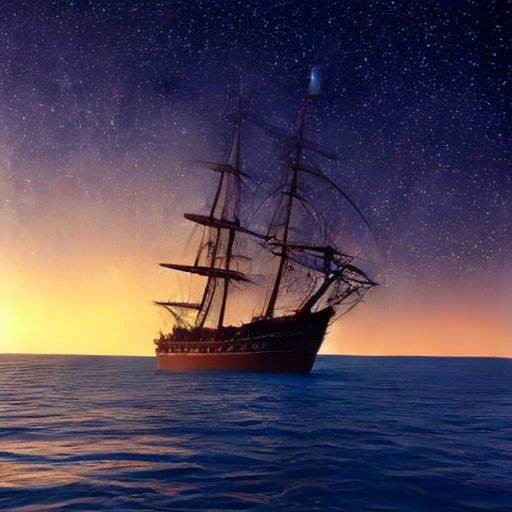}
     \\\vspace{-3mm}\caption*{\tiny (Pirate ship sailing into a bioluminescence sea with a galaxy in the sky), epic, 4k, ultra.
}
    \end{minipage}~
    \begin{minipage}[b]{0.1931\textwidth}
        \centering
        \includegraphics[width=\textwidth]{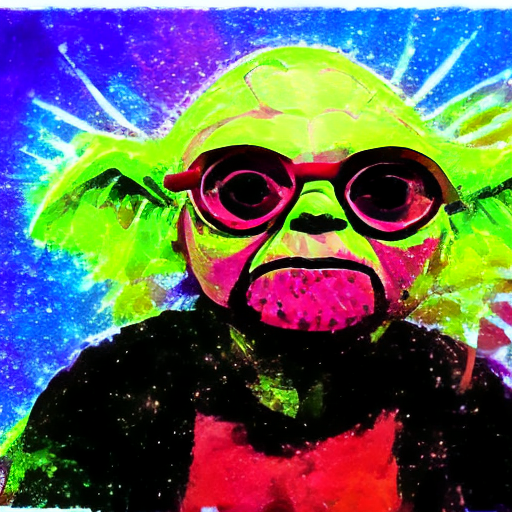}
     \\\vspace{-3mm}\caption*{\tiny 
     tshirt design, colourful, no background, yoda with sun glasses, dancing at a festival [...] 8k.
}
    \end{minipage}
    \vspace{-3mm}
    \caption{\small Example generation results of resolution 512x512  from the one-step generator distilled from Stable Diffusion 2.1-base using the proposed method: Score identity Distillation with Long-Short Guidance.\normalsize}
    \label{fig:teaser}
\end{figure}

\begin{abstract}
Diffusion-based text-to-image generation models trained on extensive text-image pairs have demonstrated the ability to produce photorealistic images aligned with textual descriptions. However, a significant limitation of these models is their slow sample generation process, which requires iterative refinement through the same network. To overcome this, we introduce a data-free guided distillation method that enables the efficient distillation of pretrained Stable Diffusion models without access to the real training data, often restricted due to legal, privacy, or cost concerns. This method enhances Score identity Distillation (SiD) with Long and Short Classifier-Free Guidance (LSG), an innovative strategy that applies Classifier-Free Guidance (CFG) not only to the evaluation of the pretrained diffusion model but also to the training and evaluation of the fake score network. We optimize a model-based explicit score matching loss using a score-identity-based approximation alongside our proposed guidance strategies for practical computation. By exclusively training with synthetic images generated by its one-step generator, our data-free distillation method rapidly improves FID and CLIP scores, achieving state-of-the-art FID performance while maintaining a competitive CLIP score. Notably, the one-step distillation of Stable Diffusion 1.5 achieves an FID of \textbf{8.15} on the COCO-2014 validation set, a record low value under the data-free setting. Our code and checkpoints are available at \href{https://github.com/mingyuanzhou/SiD-LSG}{https://github.com/mingyuanzhou/SiD-LSG}.

\end{abstract}

\section{Introduction}
The pursuit of generating high-resolution, photorealistic images that matches the textual descriptions has %
driven the machine learning community in developing powerful text-to-image (T2I) generative models. 
T2I diffusion models \citep{nichol2022glide, ramesh2022hierarchical, saharia2022photorealistic, rombach2022high,podell2024sdxl} are currently leading the way in delivering unprecedentedly visual quality, diverse generation, and accurate text-image correspondences. They are renowned for their straightforward implementation and stability during optimization, and they receive substantial acclaim for the robust support from the open-source community \citep{rombach2022high,podell2024sdxl}. 

Despite these advantages, a significant limitation of diffusion models, including those used for T2I tasks, is their slow sampling process, which involves iterative refinement through repeated passes of the generation network.
Originally, this required thousands of stochastic sampling steps \citep{song2019generative,ho2020denoising,dhariwal2021diffusion,song2021denoising}. Recent advancements in ODE-based deterministic samplers %
have significantly reduced the required number of sampling steps to just tens or hundreds \citep{song2021denoising, liu2022pseudo, lu2022dpmsolver, karras2022elucidating}.
To further reduce the number of steps below ten, or even down to one, %
the focus has shifted toward distilling the iterative-refinement based multi-step T2I generative progress, %
using a wide variety of acceleration techniques \citep{zheng2023truncated,meng2023distillation,liu2022flow,luo2023latentlora,thuan2024swiftbrush,Sauer2023AdversarialDD,Xu2023UFOGenYF,yin2023onestepDW}.  However, they often result in clearly reduced ability to match the original data distribution, reflected as clearly worsening FIDs. All of them, with the exception of SwiftBrush \citep{thuan2024swiftbrush}, also require access to real images or the assistance of extra regression or adversarial losses.

It is commonly believed that student models used for distillation sacrifice performance for increased speed. However, recent findings from Score identity Distillation (SiD) \citep{zhou2024score} present a notable discovery. The SiD-based single-step student model, although trained in a data-free manner, not only simplifies the multi-step generation process required by the teacher diffusion model, EDM by \citet{karras2022elucidating}, but also excels in performance. It surpasses the teacher model in terms of Fr\'echet inception distance (FID) \citep{heusel2017gans} on the CIFAR10-32x32, FFHQ-64x64, and AFHQ-v2 64x64 datasets. It only slightly underperforms in FID in comparison on ImageNet 64x64.
The success of SiD in distilling  EDM diffusion models for non-T2I generation in the pixel space has inspired us to adapt it to open-source T2I latent diffusion models, specifically Stable Diffusion (SD) versions 1.5 and 2.1-base, aiming to significantly enhance their generation speed while maintaining performance. However, adapting SiD poses several notable challenges: first,  SiD-EDM does not incorporate classifier-free guidance (CFG) \citep{ho2022classifier}, which is integral to SD; second, SiD has primarily been applied to distill pre-trained EDM models, which utilize noise scheduling and preconditioning methods markedly different from the DDPM noise scheduling employed in SD; third, both the complexities and sizes of the data and model in EDM are much smaller than those in~SD.

To address these challenges, we explore the integration of CFG and SiD for T2I diffusion  distillation. In addition to testing the conventional approach of enhancing CFG on the pretrained score network, we introduce a novel strategy of reducing CFG on the fake score network, as well as a combined approach that employs both strategies, which we refer to as long and short guidance (LSG). This new method  efficiently distills SD models into one-step T2I generators without requiring training data. Surprisingly, it achieves a new benchmark on the COCO-2014 validation set with a zero-shot FID score of \textbf{8.15}, the lowest to date for one-step, data-free distillation, despite not relying on additional regression or adversarial losses, nor real data—key components of recent distillation techniques.

The development of our SiD-LSG approach builds significantly on prior work in generative models, CFG, and acceleration methods. A comprehensive review of related work is provided in Appendix~\ref{sec:relatedwork}.

\begin{figure*}[!t]
\begin{center}
\includegraphics[width=0.17\linewidth]{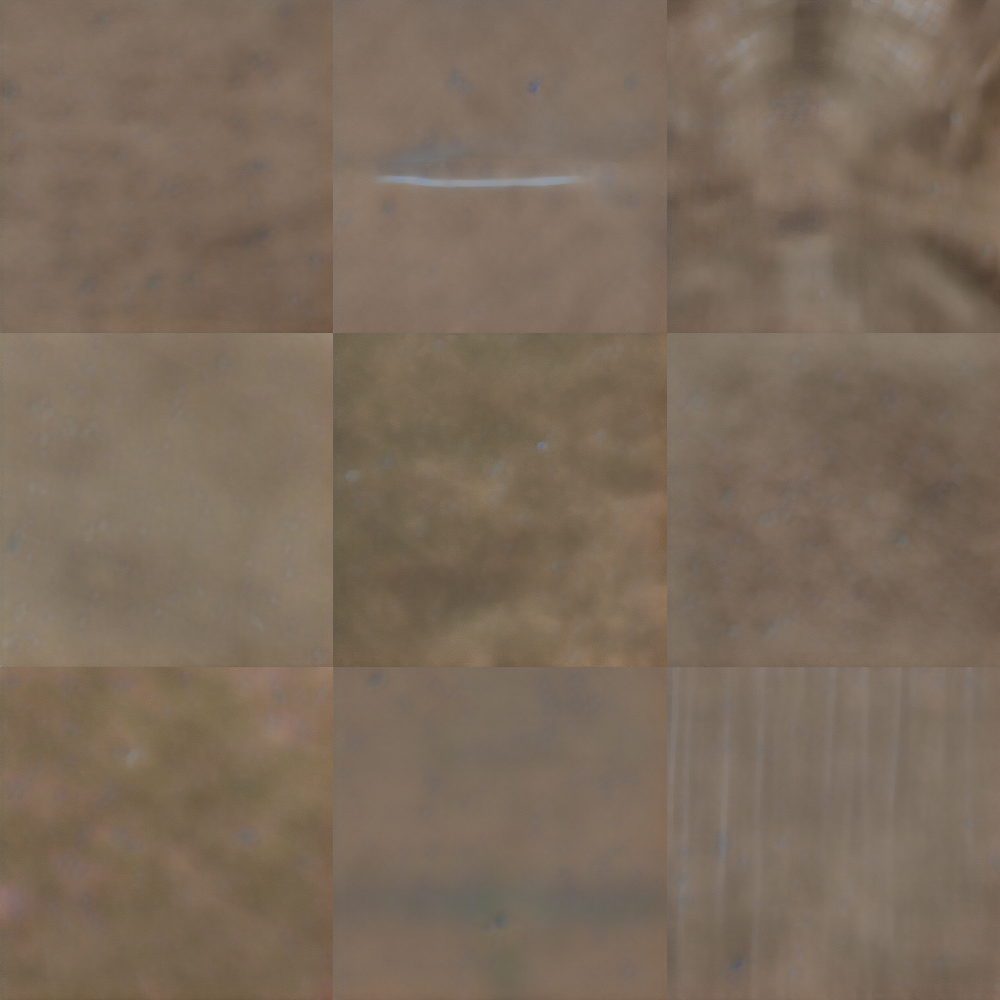}
\includegraphics[width=0.17\linewidth]{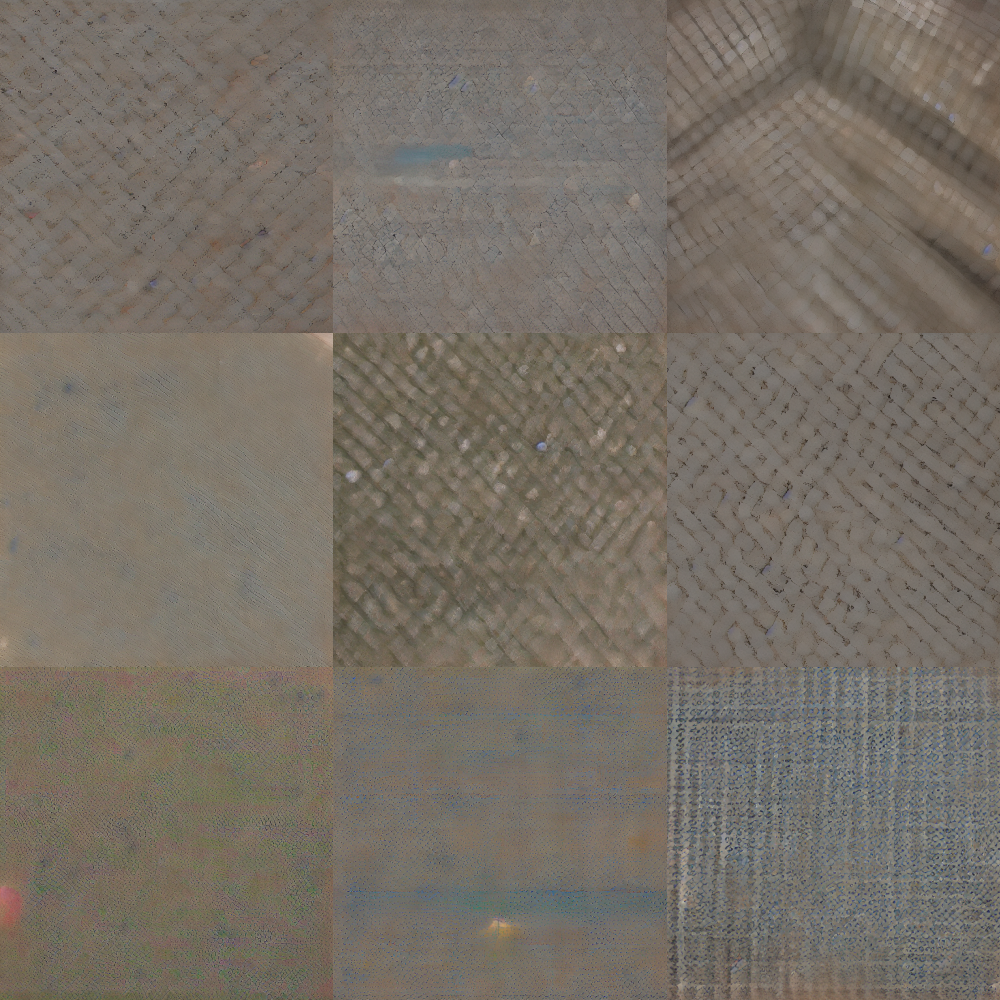}
\includegraphics[width=0.17\linewidth]{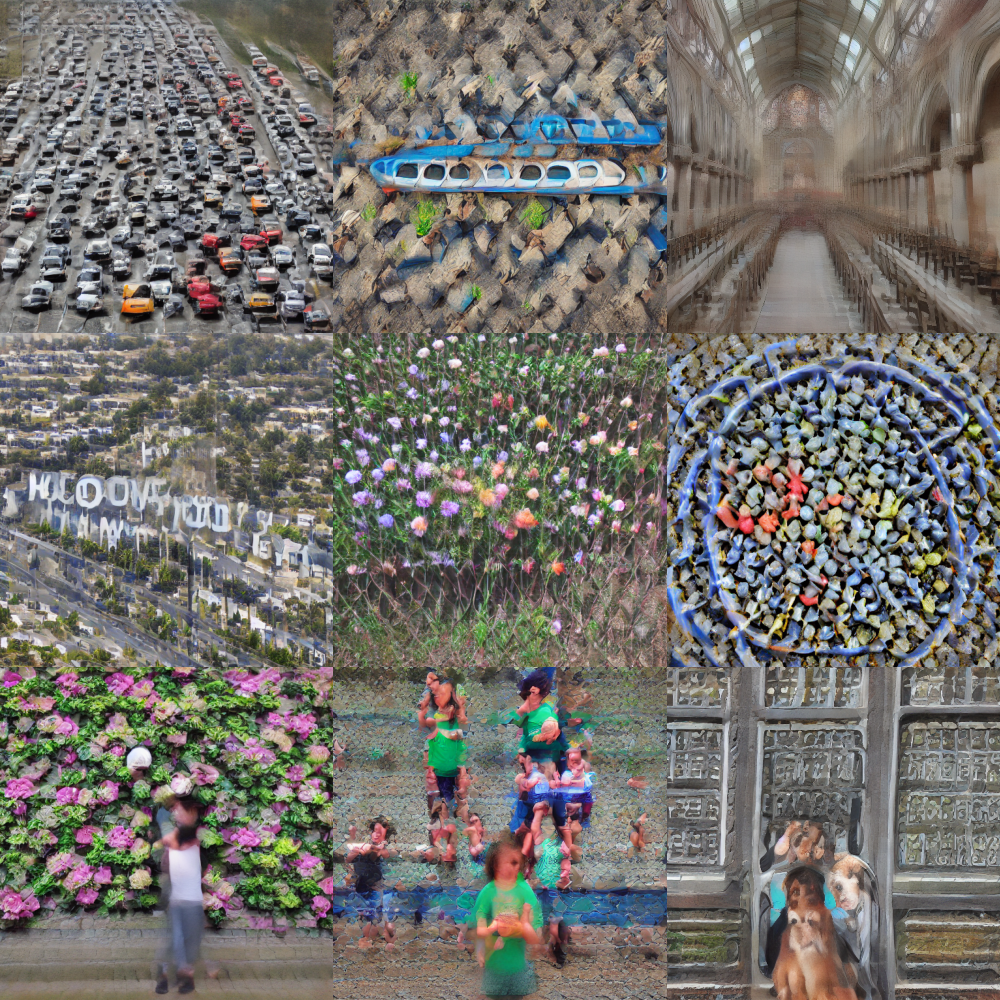}
\includegraphics[width=0.17\linewidth]{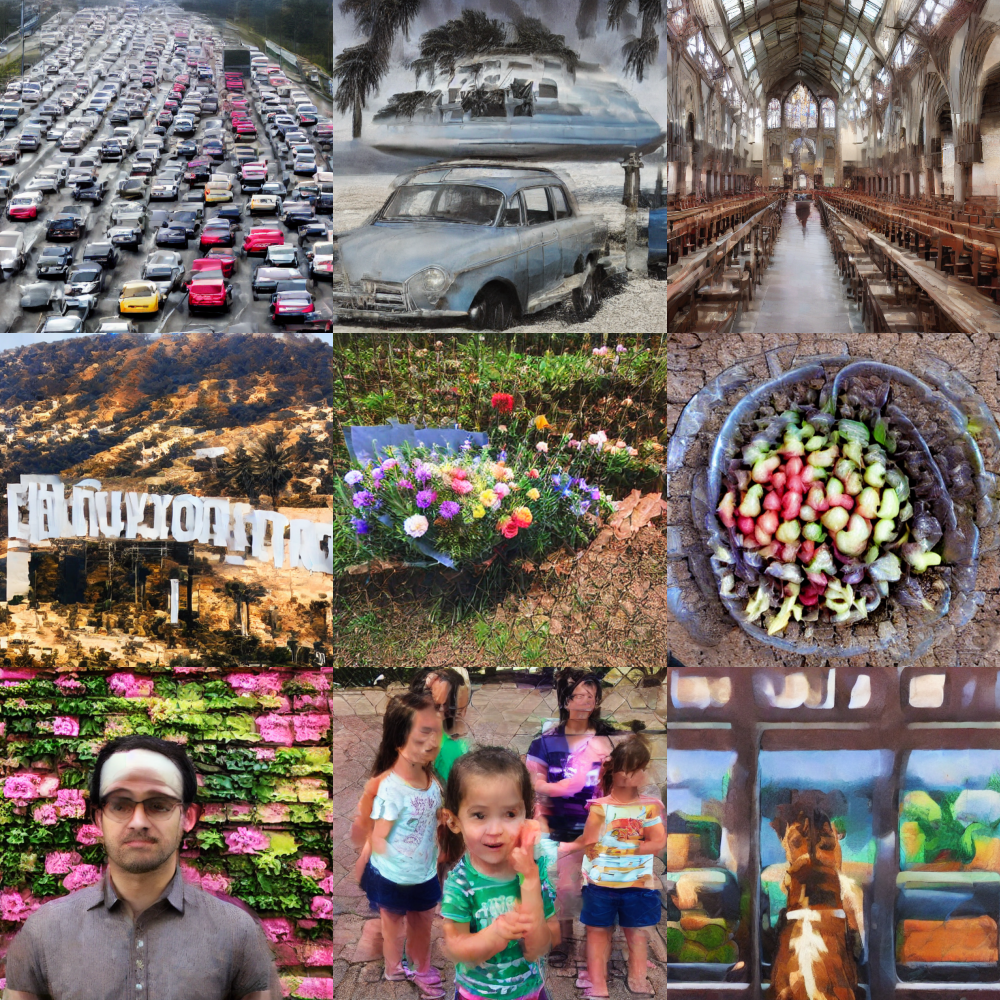}
\includegraphics[width=0.17\linewidth]{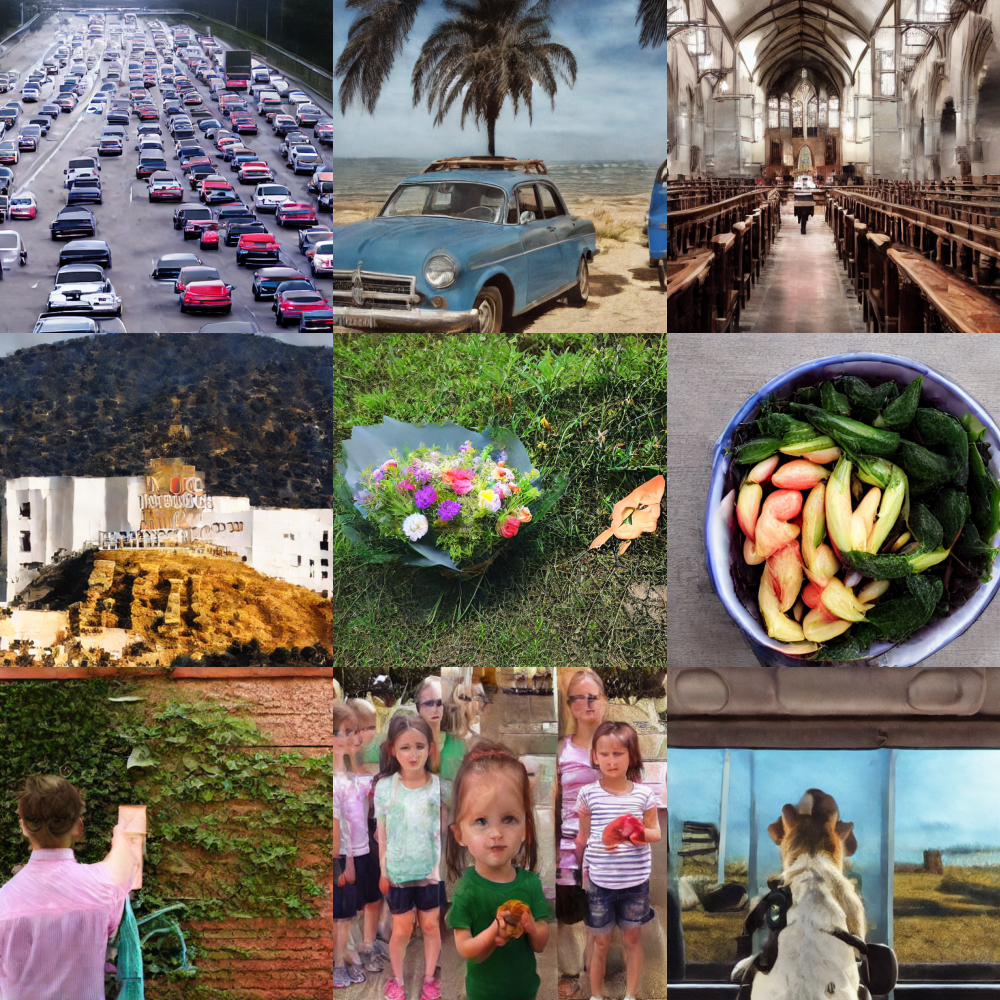}\\
\includegraphics[width=0.17\linewidth]{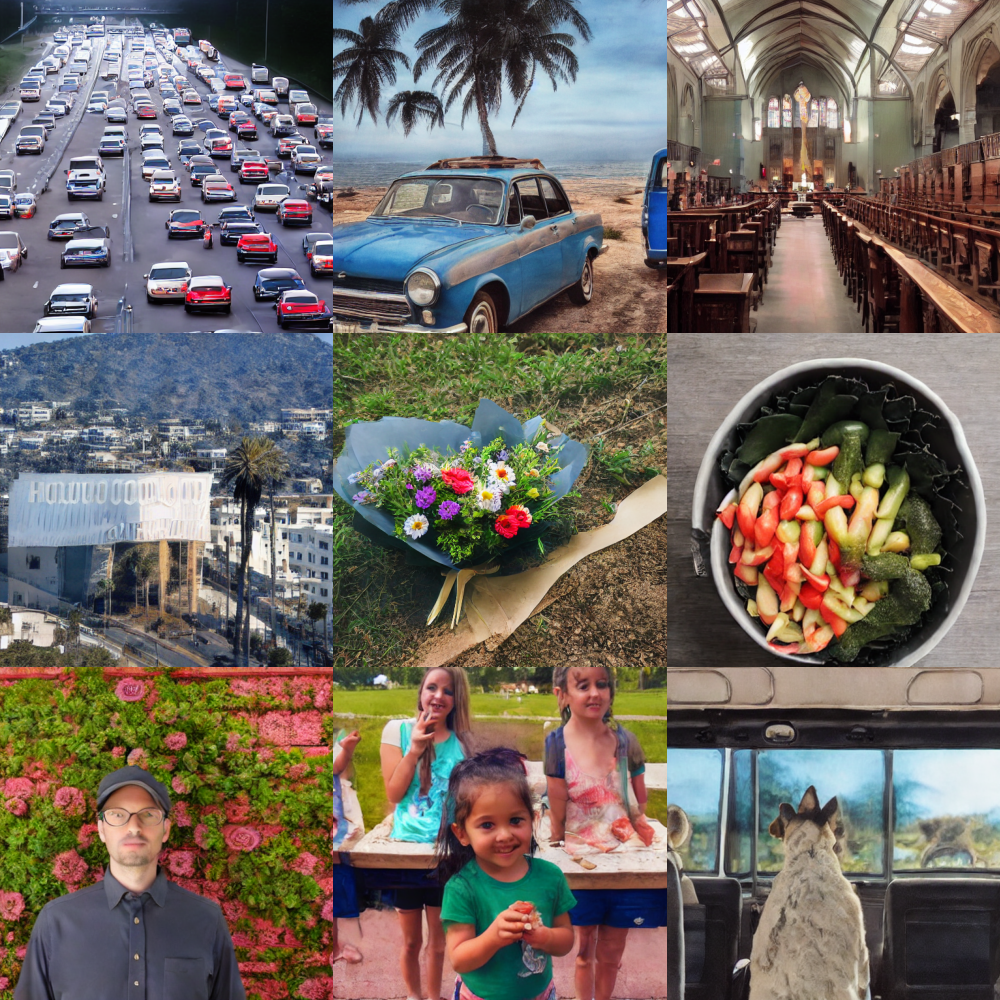}
\includegraphics[width=0.17\linewidth]{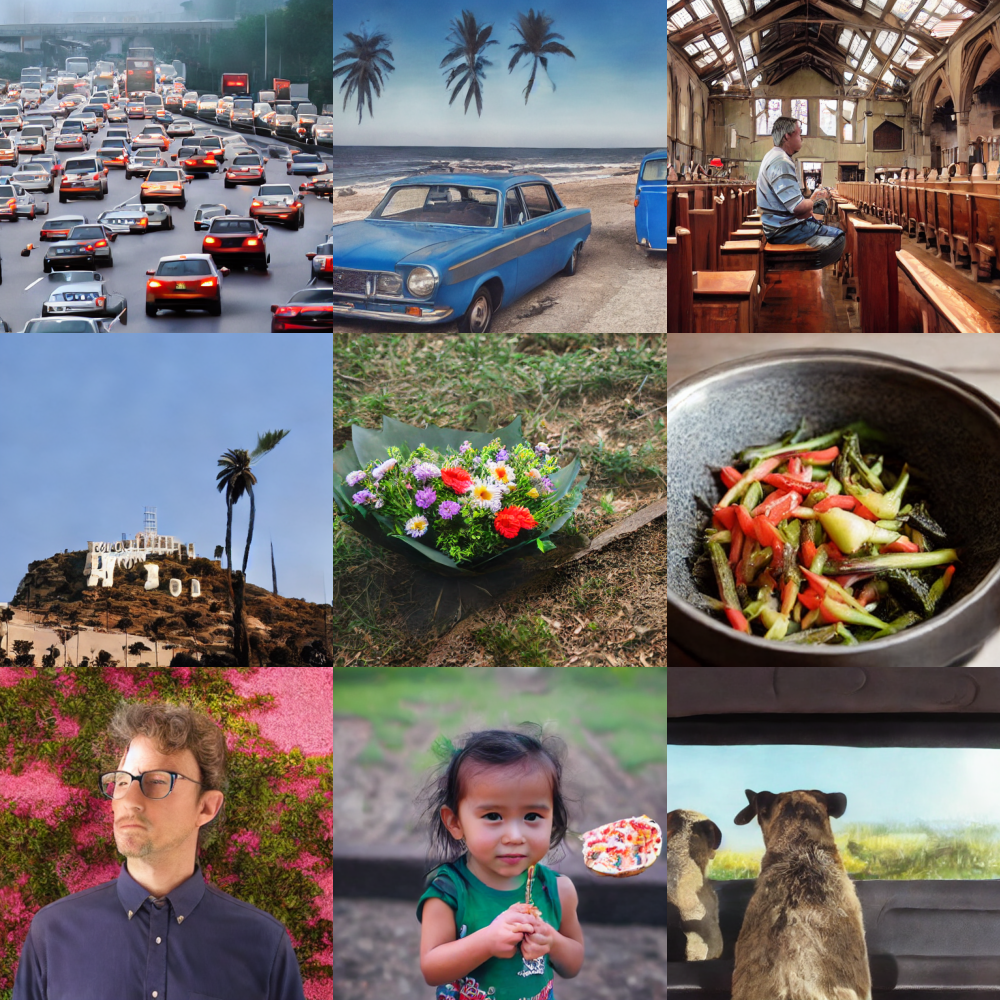}
\includegraphics[width=0.17\linewidth]{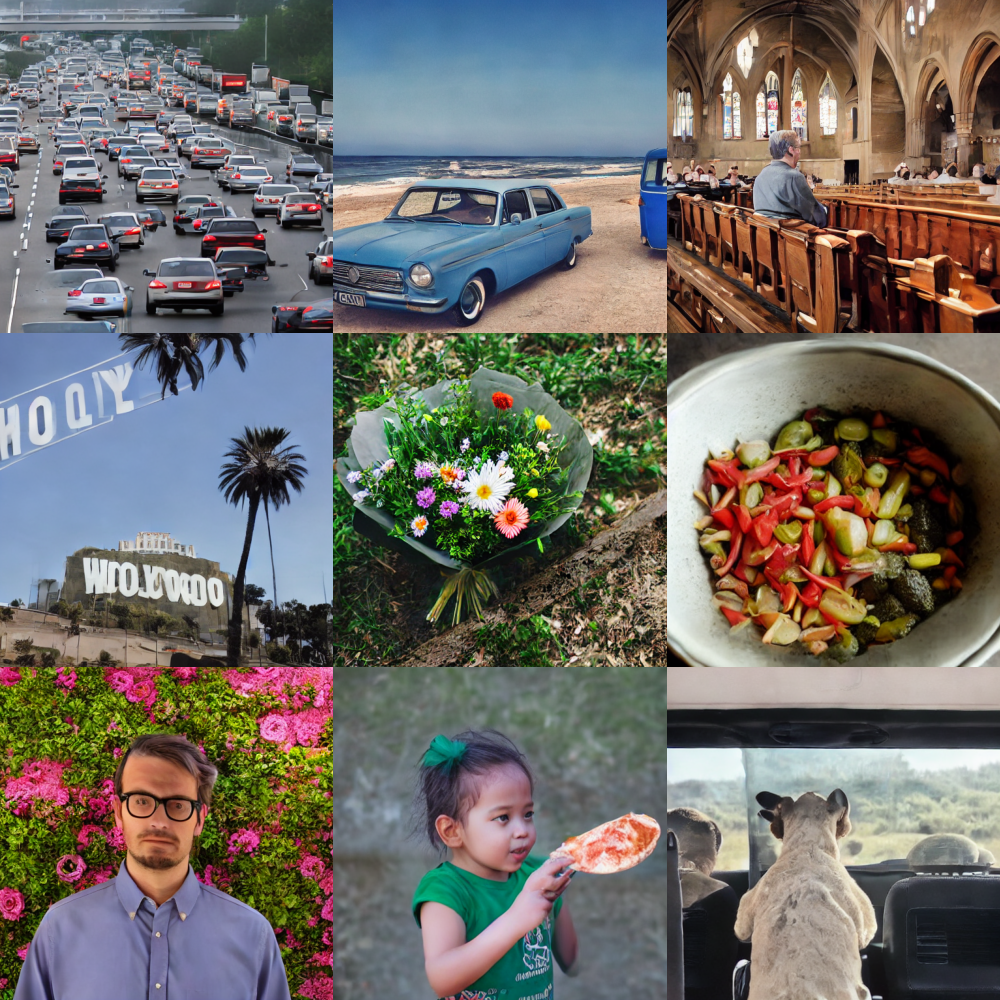}
\includegraphics[width=0.17\linewidth]{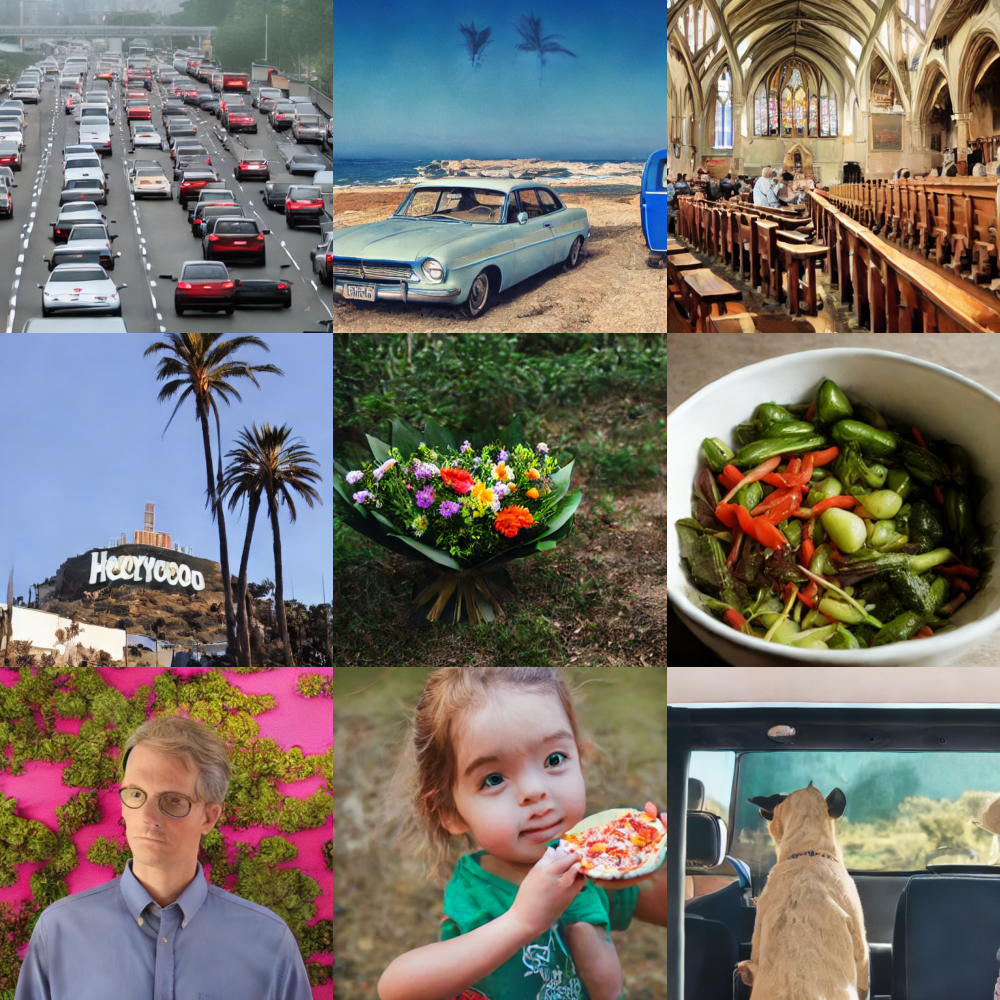}
\includegraphics[width=0.17\linewidth]{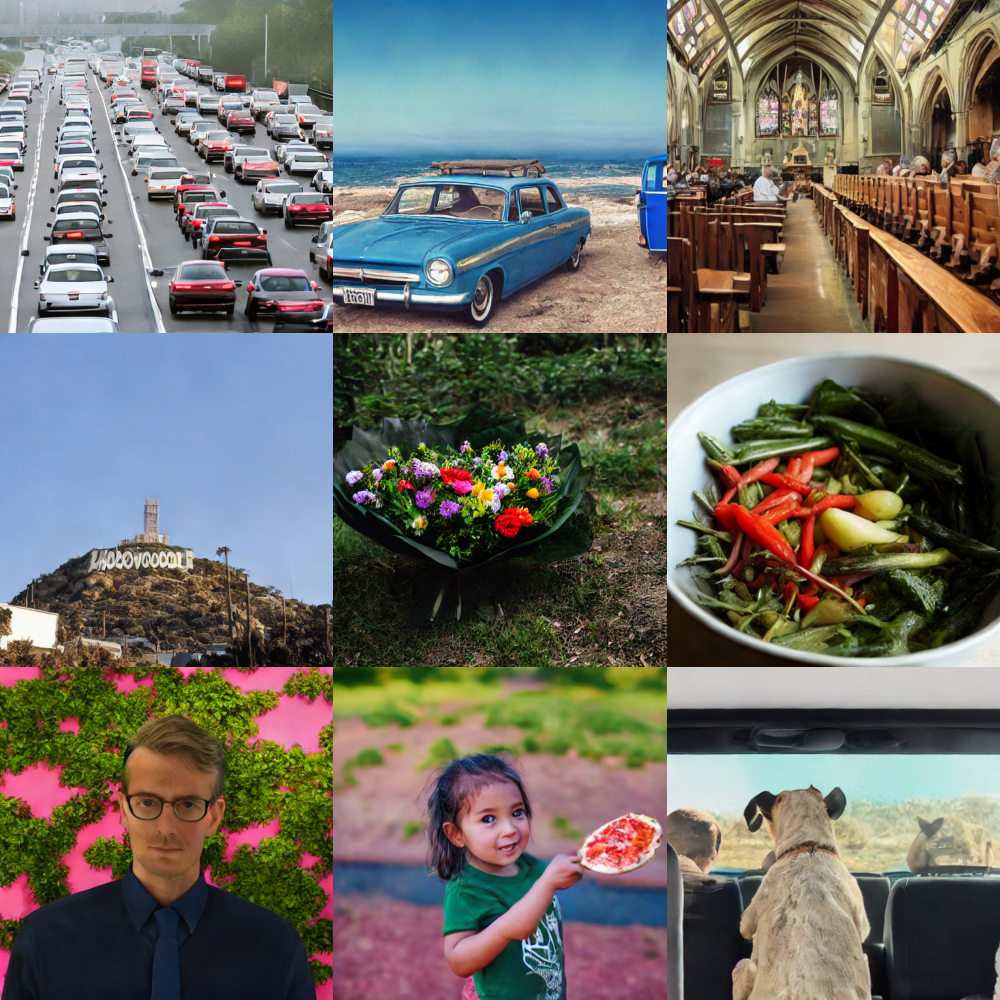}
\end{center}
 \vspace{-3.5mm}
 \caption{\small 
Rapid advancements in distilling Stable Diffusion 1.5 are showcased by the proposed SiD method that incorporates long-short guidance (LSG). Key parameters include a batch size of 512, a learning rate of 1e-6, and an LSG scale of 2. This data-free approach achieves a zero-shot FID of \textbf{9.56} on the COCO-2014 validation set, along with a competitive CLIP score of 0.313. By reducing the LSG scale to 1.5, the FID can be further lowered to a record \textbf{8.15} among data-free diffusion distillation models, with a corresponding CLIP score of 0.304.
The series of images, generated from the same set of random noises post-training the SiD generator with varying counts of synthesized images, illustrates progressions at 0, 0.02, 0.1, 0.2, 0.5, 1, 2, 3, 4, and 5 million images. These are equivalent to 0, 40, 200, 400, 1k, 2k, 4K, 6K, 8k, and 10k  training iterations respectively, organized from the top left to the bottom right. The progression of FIDs and CLIPs is detailed in the orange solid curves in the left plot of Fig.~\ref{fig:lsg}. The corresponding COCO-2014 validation text prompts are listed in Appendix \ref{sec:prompts}.
 }
 \label{fig:imagenet_progress}
 \vspace{-2.5mm}
\end{figure*}

\section{Data-Free Guided Score identity Distillation}

We explore the use of SiD and CFG to distill SD, the leading  open-source platform for T2I diffusion models that operate on the latent space of an image encoder-decoder \citep{rombach2022high}, with a specific focus on the data-free setting. Our focus is specifically on SD-v1-5 (SD1.5) and SD-v2-1-base (SD2.1-base), which are two versions extensively benchmarked for diffusion~distillation.

 A significant hurdle  involves incorporating CFG~\citep{ho2022classifier}, 
 essential for T2I diffusion models to enhance their photorealism and text alignment, into the SiD loss functions. The second challenge is adapting SiD, initially tested with EDM noise scheduling, to the  DDPM scheduling employed by SD. Addressing these challenges primarily requires modifying the derivation and loss functions of SiD. A third challenge arises because SD models are significantly larger and trained on bigger, more complex, and higher resolution data. Overcoming this challenge requires addressing numerous technical details, such as establishing the minimum hardware requirements 
and 
configuring the appropriate software settings to align with the constraints of the available computing~platforms.

\subsection{Preliminaries on Score identity Distillation}
We denote \(\cv\) as the text representation of a pretrained text encoder, such as CLIP~\citep{radford2021learning}. 
Our objective is to distill a student model \( p_{\theta}(\xv_g\given \cv) \) from a pretrained T2I diffusion model, such as SD1.5, which can generate text-guided random samples in a single step as:
$
\xv_g = G_{\theta}(\zv, \cv), ~\zv \sim p(\zv),
$
where \( G_{\theta} \) is a neural network parameterized by \(\theta\) that deterministically transforms noise \( \zv \sim p(\zv) \) into generated data \( \xv_g \) under the guidance of text \(\cv\). The distribution of \(\xv_g\) is often implicit \citep{mohamed2016learning,tran2017hierarchical,yin2018semi}, lacking an analytic probability density function (PDF) but is straightforward to sample from. 
The marginals of the real and generated data under the forward diffusion process can be expressed as:
\[
\textstyle p_{\text{data}}(\xv_t\given \cv) = \int q(\xv_t \given \xv_0)p_{\text{data}}(\xv_0\given \cv) \, \mathrm{d}\xv_0, \quad p_{\theta}(\xv_t\given \cv) = \int q(\xv_t \given \xv_g)p_{\theta}(\xv_g\given \cv) \, \mathrm{d}\xv_g.
\]
This structure, characterized by explicit conditional layers but implicit marginals, exemplifies a semi-implicit distribution \citep{yin2018semi,yu2023hierarchical}. This concept is employed by \citet{zhou2024score} to develop SiD, a method whose single-step data-free distillation capability has so far been demonstrated only on non-T2I %
diffusion models based on EDM.

We define the forward diffusion transition as $
q(\xv_t \given \xv_0) = \mathcal{N}(a_t\xv_0, \sigma_t^2\mathbf{I}),
$
and unlike in SiD, which adheres to EDM noise scheduling where \(a_t=1\), we allow \(a_t\) to vary within \([0,1]\) to align with the DDPM scheduling used by SD. %
This necessitates generalizing the equations used in SiD by permitting \(a_t \neq 1\). We note that other diffusion types, such as categorical \citep{austin2021structured,hoogeboom2021argmax,gu2021vector,hu2022global}, Poisson \citep{chen2023learning}, and beta diffusions \citep{zhou2023beta}, also align with the semi-implicit framework and can potentially be adapted~similarly.

\textbf{Score identities. } The scores \( S(\xv_t) := \nabla_{\xv_t} \ln p_{\text{data}}(\xv_t\given \cv) \) and $\nabla_{\xv_t} \ln p_{\thetav}(\xv_t\given \cv)$ are generally unknown. However, the score of the forward conditional \( q(\xv_t \given \xv) \sim \mathcal{N}(\xv, \sigma_t^2\mathbf{I})\)
is analytic: %
\begin{align}
\text{ If }\xv_t = a_t\xv+\sigma_t\epsilonv_t,~\epsilonv_t\sim\mathcal{N}(0,\mathbf{I}),~~ \text{ then }
\nabla_{\xv_t} \ln q(\xv_t \given \xv) =\sigma_t^{-2}
{(a_t\xv - \xv_t)}= -\sigma_t^{-1}{\epsilonv_t}. %
\notag
\end{align}
Exploiting the semi-implicit constructions, 
we follow SiD to present the following three identities:
\begin{align}
\textstyle
&\E[\xv_0 \given \xv_t,\cv] = \textstyle\int \xv_0 q(\xv_0 \given \xv_t,\cv) \, \mathrm d\xv_0 
= (\xv_t + \sigma_t^2 \nabla_{\xv_t} \ln p_{{\text{data}}}(\xv_t\given \cv))/a_t,
\notag
\\
\textstyle
&\E[\xv_g \given \xv_t,\cv] =\textstyle \int \xv_g q(\xv_g \given \xv_t) \, \mathrm d\xv_g 
= (\xv_t + \sigma_t^2 \nabla_{\xv_t} \ln p_\theta(\xv_t\given \cv))/a_t,
\notag\\
&\E_{p_{\theta}(\xv_t\given \cv)}\left[u^T(\xv_t) \nabla_{\xv_t}\ln p_{\theta}(\xv_t\given \cv)\right]= \E_{q(\xv_t \given \xv_g) p_{\theta}(\xv_g\given \cv)}\left[u^T(\xv_t) \nabla_{\xv_t} \ln q(\xv_t \given \xv_g, \cv)\right]. \notag
\label{projected_score}
\end{align}

\textbf{MESM loss. } A pretrained T2I diffusion model, such as SD, provides a score network $S_{\phi}$ parameterized by \( \phi \) that estimates the true data score as %
\begin{equation}
    -\sigma_t\nabla_{\xv_t} \ln p_\text{data}(\xv_t\given \cv) \approx -\sigma_tS_{\phi}(\xv_t, \cv):= {\sigma_t^{-1}}({\xv_t-a_t f_{\phi}(\xv_t,t,\cv)}) =  \epsilonv_{\phi}(\xv_t,\cv). \notag
\end{equation}
It adopts \( f_{\phi}(\xv_t,t,\cv) \) as the functional approximation of the conditional expectation of the real image $\xv_0$ given noisy image $\xv_t$ and text $\cv$, expressed as $\E[\xv_0\given  \xv_t,\cv]$, adopts $\epsilonv_{\phi}(\xv_t,t,\cv)$ to predict the noise inside $\xv_t$, and adopts $-\sigma_t^{-1}\epsilonv_{\phi}(\xv_t,\cv)$ as the functional approximation of the true score $\nabla_{\xv_t} \ln p_\text{data}(\xv_t\given \cv) $.
Given time step \( t \sim p(t)\) and text $\cv$, we define the model-based explicit score-matching (MESM) distillation loss, which is a form of Fisher divergence \citep{lyu2009interpretation,holmes2017assigning,yang2019variational,yu2023semiimplicit}, as 
\begin{align}\textstyle 
&\gL_{\theta} = 
\E_{\xv_t\sim p_{\theta}(\xv_t)}[
\|S_{\phi}(\xv_t,\cv) - \nabla_{\xv_t}\ln p_{\theta}(\xv_t\given \cv)\|_2^2].\label{eq:MESM} 
\end{align}

\textbf{Loss approximation for distillation. } %
The MESM loss in \eqref{eq:MESM} is in general intractable to compute as 
$\nabla_{\xv_t}\ln p_{\theta}(\xv_t\given \cv)$ is unknown. %
To denoise the noisy fake data $\xv_t$ generated as 
\ba{\xv_t=a_t \xv_g+\sigma_t\epsilonv_t,~~\epsilonv_t\sim \mathcal{N}(0,\mathbf{I}),~~
\xv_g = 
G_{\theta}(\zv,\cv),~ ~\zv\sim p(\zv),
\label{eq:repara}}
there exists an optimal denoising network 
defined as
$
f_{\psi^*(\theta)}(\xv_t,t,\cv)=\E[\xv_g\given \xv_t,\cv]=(\xv_t + \sigma_t^2 \nabla_{\xv_t} \ln p_\theta(\xv_t\given \cv))/a_t.
$ Given this optimal denoising network, %
the MESM loss  in \eqref{eq:MESM}  would become 
\begin{align}\textstyle 
&\gL_{\theta} = \E_{\xv_t\sim p_{\theta}(\xv_t)}[
\|a_t\sigma_t^{-2} (f_{\phi}(\xv_t,t,\cv)-
f_{\psi^*(\theta)}(\xv_t,t,\cv)\|_2^2].
\end{align}
As $\psi^*(\theta)$ is unknown in practice, %
we follow SiD to %
alternates between optimizing $\psi$ and $\theta$ using
\ba{
\min_{\psi}\hat{\cL}_{\psi} (\xv_t,c,t)&=\textstyle\frac{a^2_t}{\sigma^2_t} %
\|f_{\psi}(\xv_t, t,\cv)-\xv_g\|_2^2 =  \|\epsilonv_{\psi}(\xv_t, t,\cv)-\epsilonv_t\|_2^2, \label{eq:psi_optimize}\\
\min_{\theta}\textstyle \tilde{L}_{\theta}(\xv_t,t,\phi,\psi)
&\textstyle=
\omega(t)\frac{a_t^2}{{\sigma_t^4} } (f_{\phi}(\xv_t,t,\cv)-f_{\psi}(\xv_t,t,\cv))^T(f_{\psi}(\xv_t,t,\cv)-\xv_g),
\label{eq:obj-theta}\\
&=\textstyle\omega(t)\frac{1}{{\sigma_t^2} } (\epsilonv_{\psi}(\xv_t, t,\cv)-\epsilonv_{\phi}(\xv_t, t,\cv))^T(\epsilonv_t-\epsilonv_{\psi}(\xv_t, t,\cv)),\notag
}
where $\xv_t$ is generated as in \eqref{eq:repara} and $\omega(t)$ are weighted coefficients that will be specified. 

\subsection{SiD with classifier-free guidance}

An essential practice for enhancing photorealism and alignment with text instructions in T2I diffusion models involves incorporating CFG into %
reverse diffusion. %
This principle also applies to distillation methods for these models, where CFG must be integrated into the appropriate terms of their distillation loss functions. Therefore, a key distinction in the distillation of SD, compared to previous unconditional and label-conditional diffusion models, lies in the need to introduce CFG.

 SiD presents a unique opportunity to apply CFG to enhance its T2I generation performance. First, we note CFG enhances text guidance by modifying the distribution of $\xv_t$ given text $\cv$ as
$$
 \textstyle p(\xv_t\given \cv,\kappa) \propto p( \cv\given \xv_t)^{\kappa}p(\xv_t) \propto \big(\frac{p(\xv_t\given \cv)}{p(\xv_t)}\big)^\kappa p(\xv_t),
$$
which means $\nabla_{\xv_t}\ln p(\xv_t\given \cv,\kappa) = \nabla_{\xv_t}\ln p(\xv_t) +  \kappa[\nabla_{\xv_t}\ln p(\xv_t\given\cv)-\nabla_{\xv_t}\ln p(\xv_t)]$.
Second, the score network, when reaching its optimal, is related to the true score as $
f(\xv_t, t,\cv)={a_t^{-1}}({\xv_t+\sigma_t^2\nabla_{\xv_t}\ln p(\xv_t\given c)}) = {a_t^{-1}}({\xv_t-\sigma_t\epsilonv(\xv_t, t,\cv)}).
$
Therefore, with ``$\cdotv$'' representing $\phi$~or~$\psi$, we can express the score network $f_{\cdotv}(\xv_t,t,\cv)$ under CFG with scale $\kappa$ as
 \ba{
 f_{\cdotv,\kappa}(\xv_t,t,\cv) =  f_{\cdotv}(\xv_t,t) + \kappa[ f_{\cdotv}(\xv_t,t,c)-f_{\cdotv}(\xv_t,t)].
 }
 Similarly, for noise prediction network, we have $
 \epsilonv_{\cdotv,\kappa}(\xv_t,t,\cv) =   \epsilonv_{\cdotv}(\xv_t,t) + \kappa[  \epsilonv_{\cdotv}(\xv_t,t,c)- \epsilonv_{\cdotv}(\xv_t,t)].
 $
 With CFG, the score and noise prediction networks are related in the same way, which means that  $
\epsilonv_{\cdotv,\kappa}(\xv_t, t,\cv) = {\sigma_t^{-1}}({\xv_t-a_t{f_{\cdotv,\kappa}(\xv_t, t,\cv)}})$ and $f_{\cdotv,\kappa}(\xv_t, t,\cv) = {a_t^{-1}}({\xv_t-\sigma_t\epsilonv_{\cdotv,\kappa}(\xv_t, t,\cv)}).
$

 Inspecting the two losses in \eqref{eq:psi_optimize} and \eqref{eq:obj-theta} suggests four potential places to inject CFG. More specifically, with $\kappa_1,\kappa_2,\kappa_3,\kappa_4\in\mathbb{R}_+$, where $\mathbb{R}_+ =\{x:x\ge 0\}$, we  modify the losses with CFGs as
 \ba{
\hat{\cL}_{\psi} (\xv_t,c,t)&=\textstyle\frac{a^2_t}{\sigma^2_t} %
\|f_{\psi,\kappa_1}(\xv_t, t,\cv)-\xv_g\|_2^2 ,
=  \|\epsilonv_{\psi,\kappa_1}(\xv_t, t,\cv)-\epsilonv_t\|_2^2, 
\label{eq:psi_optimizecfg}\\
\textstyle \tilde{L}_{\theta}(\xv_t,t,\phi,\psi)
&\textstyle=
\omega(t)\frac{a_t^2}{{\sigma_t^4} } (f_{\phi,\kappa_4}(\xv_t,t,\cv)-f_{\psi,\kappa_2}(\xv_t,t,\cv))^T(f_{\psi,\kappa_3}(\xv_t,t,\cv)-\xv_g).
\label{eq:obj-theta_cfg}
}

\subsection{Long and Short Guidance}
Previous works equipped with a fake score network $f_{\psi}$ typically only consider adding CFG when evaluating the pretrained score network $f_{\phi}$, such as in DMD \citep{yin2023onestepDW}. 
In the context of SiD, this corresponds to  setting $\kappa_1=\kappa_2=\kappa_3=1$ and $\kappa_4>1$.
In this paper, we discover that a broad spectrum of combinations of $\kappa_1$, $\kappa_2$, $\kappa_3$, and $\kappa_4$ can all significantly enhance performance compared to not using any CFG at all, which means setting $\kappa_1=\kappa_2=\kappa_3 =\kappa_4=1$. These different combinations are found to
lead to different balances of minimizing FID, which reflects how well the generated data match the training data in distribution, and maximizing the CLIP score~\citep{radford2021learning}, which reflects how well the generated images follow the textual~guidance. 
This flexibility to accommodate various CFG combinations expands the design space for SD distillation, balancing generation quality and text adherence. 
However, given the vastness of the search space, we are motivated to develop strategies that constrain the scope of exploration. We present three such strategies, acknowledging that there may be more effective approaches not explored in this paper.
\textbf{Long strategy: Enhancing CFG of the pretrained score network $f_{\phi}$. }
Aligning with established practices, the most common approach involves enhancing the CFG of the pretrained score network~\( f_{\phi} \). An example setting under this strategy in SiD is: \( \kappa_1=\kappa_2=\kappa_3=1 \) and \( \kappa_4 =3 \). The rationale is that by biasing the teacher \( f_{\phi} \) to favor generations more aligned with \( \cv \) using a CFG scale greater than~1, such as  \( \kappa_4 =3 \), the student generator is compelled to follow suit. We will present experimental results to demonstrate the effectiveness of this strategy while also discussing its limitations.

\textbf{Short strategy: Weakening CFG of the fake score network $f_{\psi}$.}
With the availability of a fake score network \( f_{\psi} \), we have developed a new strategy to enhance SiD's T2I generation capabilities by reducing the CFG of \( f_{\psi} \) in SiD. An exemplary configuration is setting \( \kappa_1=\kappa_4=1 \) while allowing \( \kappa_2=\kappa_3 \) to vary within $(0,1)$. The underlying idea is that by diminishing \( f_{\psi} \)'s capacity to detect generations aligned with \( \cv \) through a reduced CFG scale (\( 0<\kappa_2<1 \)), the student generator is incentivized to produce images that better align with the textual guidance to compensate for this reduction. We will present experiments to assess the efficacy and limitations of this strategy.

\textbf{Long and short CFGs. } We introduce an innovative strategy termed long-short guidance (LSG), where we enhance CFG during the training of \( f_{\psi} \) by setting \( \kappa_1>1 \), and maintain CFGs on \( f_{\psi} \) and \( f_{\phi} \) during the training of \( G_{\theta} \) at the same or a lower scale by setting \( 1 \leq \kappa_2 = \kappa_3 = \kappa_4 \leq \kappa_1 \). The logic behind LSG is that enhancing the CFG of \( f_{\psi} \) during training effectively reduces its CFG at evaluation time when used at the same or a reduced level. To our knowledge, we are the first to incorporate CFG into the training of the fake score network \( f_{\psi} \). LSG has been shown to effectively balance reducing FID scores while increasing CLIP scores, and will therefore be used as the default method. 

{We refer the reader to Appendix \ref{sec:discuss-LSG} for further discussion on the long and short CFG strategies.}

\section{Experiments} \label{sec:experiment}

\begin{figure}[t]
\centering
\includegraphics[width=.4\linewidth]{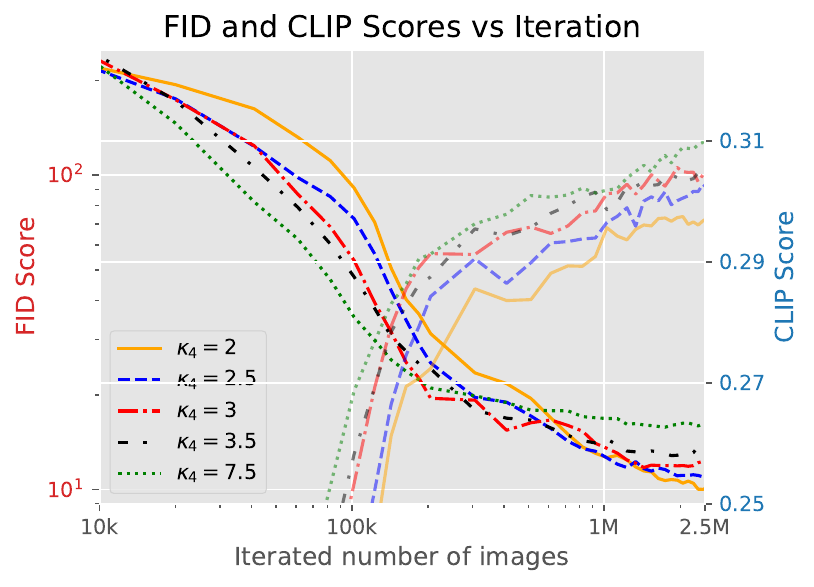}~~
\includegraphics[width=.4\linewidth]{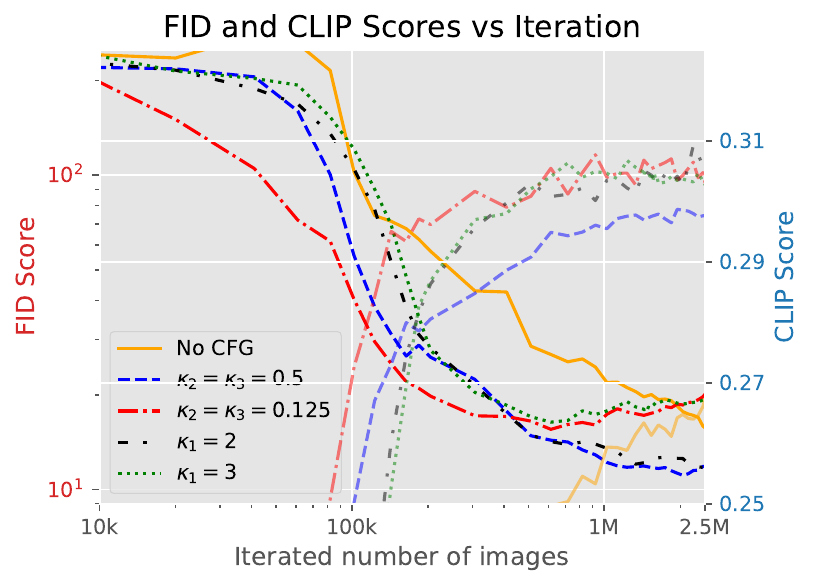}
 \vspace{-3mm}
\caption{\small \textbf{Left (Long CFG of the true score network):} This plot illustrates the {gradual decline in FID and the corresponding rise in CLIP} scores, each influenced by different CFGs applied to the true score network. $\kappa$ values not specified in the legend are set to 1. FID scores %
are plotted on the primary y-axis, %
while CLIP scores %
are displayed on the secondary y-axis in corresponding line styles but with slight transparency. Together, these lines demonstrate how various CFGs impact model performance. \textbf{Right (No CFG; Short CFG of the fake score network with $\kappa_2=\kappa_3\in(0,1)$; a simple form of LSG that sets $\kappa_1>1$):} Analogous plot to the left where the CFGs of the fake score network are not applied, weakened during evaluation, or enhanced during training.
 }
 \label{fig:cfg_1114}
\vspace{2mm}
\centering
\includegraphics[width=.4\linewidth]{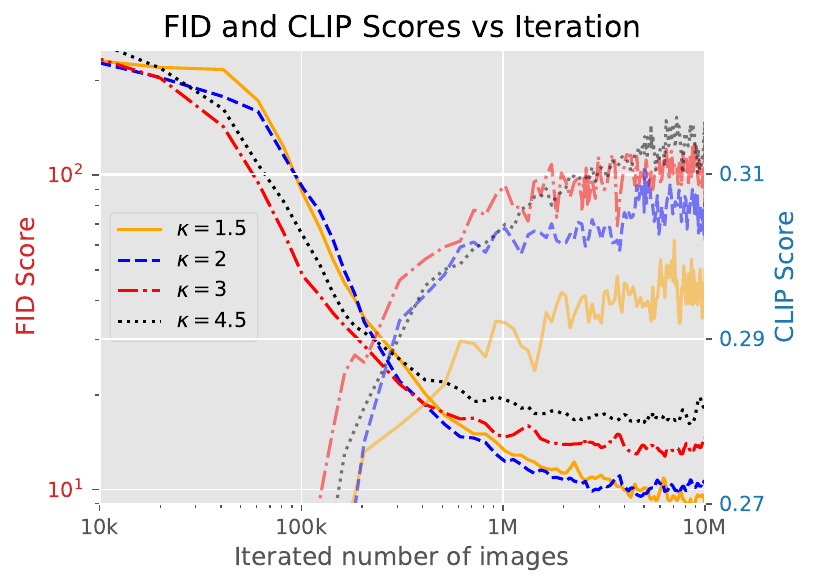}~~
\includegraphics[width=.4\linewidth]{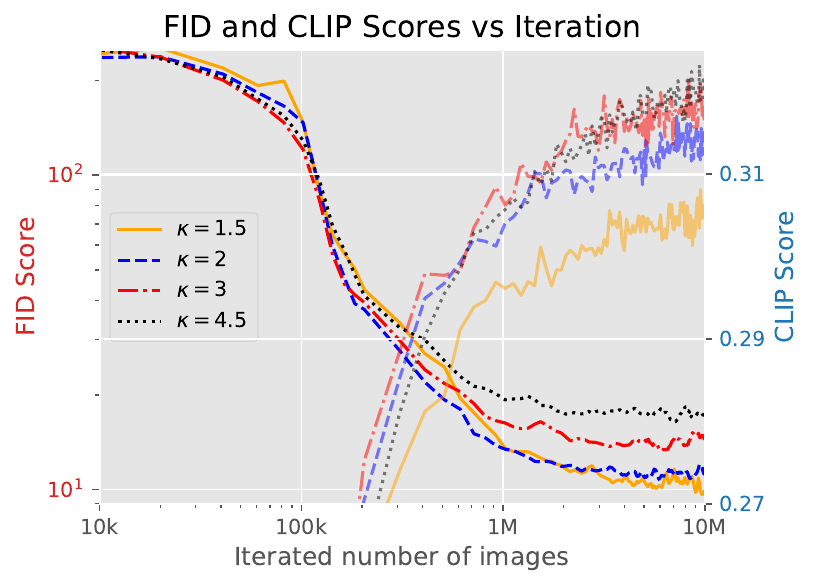}
 \vspace{-3mm}
\caption{\small 
The plot with the proposed long-short guidance (LSG) demonstrates the FID and CLIP progressions of SiD in SD1.5 (left panel) and SD2.1-base (right panel).
 }
 \label{fig:lsg}
 \vspace{-3mm}
\end{figure}

We summarize the parameter settings, such as batch size, learning rate, and optimizer configurations, in Table~\ref{tab:Hyperparameters} in Appendix~\ref{sec:detail}. Unless specified in ablation studies, the settings are uniformly applied across all guidance strategies.
We present the details of our method in Algorithm \ref{alg:sid}, where
we generate an image in one step and generalize the setting in  DMD and SiD to set $\omega(t)$ as
\ba{
\xv_g=G_{\theta}( \zv,\cv)=f_{\theta}(\sigma_{t_{init}} \zv,t_{init},\cv), ~~\zv\sim\mathcal{N}(0,\mathbf{I});~~~~\textstyle\omega(t) = \frac{\sigma_t^4}{a_t^2}\frac{C}{\|\xv_g-f_{\phi,\kappa_4}(\xv_t, t,\cv)\|_{1,sg}}~,\label{eq:weight}
}
where $C$ is the total number of pixels in $\xv_g$. 
We initialize both ${\psi}$ and ${\theta}$ using the pretrained ${\phi}$ from~SD.
We have %
tested 
$t_{init}\in\{354,550, 625,675,800,900,999\}$ and found that the model’s performance under FP32 optimization is not sensitive to these choices.
To align with the setting in SiD, we set $t_{init}=625$, which would result in $\sigma_{t_{init}}/a_{t_{init}}=2.5$  under the DDPM schedule used to train SD1.5.
We conduct a comprehensive study to evaluate the performance of SiD using the proposed LSG for distilling SD1.5. Additionally, we apply the same LSG scales to distill SD2.1-base, further assessing the adaptability and effectiveness of our approach across different model versions.

We consider a standard setting that utilizes the Aesthetics6+ prompt \citep{cherti2023reproducible} for training and evaluates performance by computing zero-shot FID on the COCO-2014 validation set. We adhere to the standard protocol by generating 30k images to compare with the 40,504 images in the COCO-2014 validation set for calculating zero-shot FID. Additionally, we employ the \verb|ViT-g-14-laion2b_s12b_b42k| encoder \citep{ilharco_gabriel_2021_5143773,cherti2023reproducible} to compute the CLIP score \citep{radford2021learning}. 
The FID and CLIP scores presented in the figures are calculated using randomly sampled prompts from the COCO-2014 validation set. %
When reporting the FID and CLIP results of SiD in Table \ref{tab:comparison},  we
use the exact evaluation code\footnote{\href{https://github.com/mingukkang/GigaGAN/tree/main/evaluation}{https://github.com/mingukkang/GigaGAN/tree/main/evaluation}} provided by GigaGAN~\citep{kang2023scaling}, where a pre-defined list of 30k text prompts selected from the COCO-2014 validation set is used to generate the 30k images, which are used for computing FID and CLIP scores with images from the validation set as the reference.

\begin{table}[!t]
\centering
\small
\caption{\small Comparison of image generation methods across various metrics. Inference times are estimated using a Nvidia A100 GPU as reference. The numbers of the methods marked with $^\dagger$ are produced by running the publicly available model checkpoints. Other data are sourced from corresponding scientific papers with comparisons as reported. The value followed by symbol $\sim$  indicates it is estimated based on the plots shown in the paper.}
\label{tab:comparison}
\vspace{-2mm}
\resizebox{.88\textwidth}{!}{
\begin{tabular}{@{}llccccc@{}}
\toprule
\textbf{Method} & \textbf{Res.} & \textbf{Time ($\downarrow$)} & \textbf{\# Steps} & \textbf{\# Param.} & \textbf{FID ($\downarrow$)} & \textbf{CLIP ($\uparrow$)}\\ \midrule
\multicolumn{7}{c}{Autoregressive Models} \\ 
 DALL·E~\citep{ramesh2021zero} & 256 & - & - & 12B & 27.5 & - \\
 CogView2~\citep{ding2021cogview} & 256 & - & - & 6B & 24.0 & - \\
 Parti-750M~\citep{yu2022scaling} & 256 & - & - & 750M & 10.71 & - \\
Parti-3B~\citep{yu2022scaling} & 256 & 6.4s & - & 3B & 8.10 & - \\
Parti-20B~\citep{yu2022scaling} & 256 & - & - & 20B & 7.23 & - \\
Make-A-Scene~\citep{gafni2022make} & 256 & 25.0s & - & - & 11.84 & - \\
\multicolumn{7}{c}{Masked Models} \\ 
Muse~\citep{chang2023muse}  & 256  &  1.3 & 24 & 3B & 7.88 & 0.32\\
\midrule
\multicolumn{7}{c}{Diffusion Models} \\ 
 GLIDE~\citep{nichol2022glide} & 256 & 15.0s & 250+27 & 5B & 12.24 & -\\
 DALL·E 2~\citep{ramesh2022hierarchical} & 256 & - & 250+27 & 5.5B & 10.39 & - \\
LDM-KL-8-G$^*$~\citep{rombach2022high} & 256 & 3.7s & 250 & 1.45B & 12.63& - \\
Imagen~\citep{ho2022imagen} & 256 & 9.1s & - & 3B & 7.27 & - \\
eDiff-I~\citep{balaji2022ediffi} & 256 & 32.0s & 25+10 & 9B & 6.95 & - \\
\midrule
\multicolumn{7}{c}{Generative Adversarial Networks (GANs)} \\
LAFITE~\citep{zhou2022towards} & 256 & 0.02s & 1 & 75M & 26.94 & - \\
 StyleGAN-T~\citep{sauer2023stylegan} & 512 & 0.10s & 1 & 1B & 13.90& $\sim$0.293\\
 GigaGAN~\citep{kang2023scaling} & 512 & 0.13s & 1 & 1B & 9.09 & -\\
  \midrule
   \multicolumn{7}{c}{Distilled Stable Diffusion 2.1}\\
   $^\dagger$ADD (SD-Turbo)~\citep{Sauer2023AdversarialDD} &512 & - & 1 & - &16.25 &0.335\\
     \midrule
   \multicolumn{7}{c}{Distilled Stable Diffusion XL}\\
     $^\dagger$ADD (SDXL-Turbo)~\citep{Sauer2023AdversarialDD} &512 & - & 1 & - & 19.08& 0.343\\
 \midrule
  \multicolumn{7}{c}{Stable Diffusion 1.5 and its accelerated or distilled versions}\\
    SD1.5 (CFG=3)~\citep{rombach2022high} & 512 & 2.59s & 50 & 0.9B & 8.78& - \\
    SD1.5 (CFG=8)~\citep{rombach2022high}&512& 2.59s & 50 & 0.9B &13.45&0.322\\
  \hdashline
  DPM++ (4 step)~\citep{lu2022dpm++} & 512 & 0.26s & 4 & 0.9B & 22.44 & 0.31\\
 UniPC (4 step)~\citep{zhao2023unipc} & 512 & 0.26s & 4 & 0.9B & 22.30 & 0.31 \\
 LCM-LoRA (4 step)~\citep{luo2023latentlora} & 512 & 0.19s & 4 & 0.9B & 23.62& 0.30\\
  LCM-LoRA (1 step)~\citep{luo2023latentlora} & 512 & 0.07s & 1 & 0.9B & 77.90 & 0.24\\
 InstaFlow-0.9B ~\citep{liu2023insta} & 512 & 0.09s & 1 & 0.9B & 13.10 & 0.28\\
  InstaFlow-1.7B ~\citep{liu2023insta} & 512 & 0.12s & 1 & 1.7B & 11.83 & - \\
 UFOGen~\citep{Xu2023UFOGenYF} & 512 & 0.09s & 1 & 0.9B & 12.78 & - \\
 DMD (CFG=3)~\citep{yin2023onestepDW}  & 512 & 0.09s & 1 & 0.9B & 11.49 & -\\
  DMD (CFG=8)~\citep{yin2023onestepDW} & 512 & 0.09s & 1 & 0.9B & 14.93 & 0.32\\
 BOOT~\citep{gu2023boot} & 512 & 0.09s & 1 & 0.9B & 48.20 & 0.26 \\
 Guided Distillation~\citep{meng2023distillation} & 512 & - & 1 & 0.9B & 37.3 &0.27\\
  SiD-LSG ($\kappa=1.5$) &512& 0.09s & 1 & 0.9B & 8.71&0.302\\
   SiD-LSG ($\kappa=1.5$, double the training time) &512& 0.09s & 1 & 0.9B & \textbf{8.15}&0.304\\
  SiD-LSG ($\kappa=2$) &512& 0.09s & 1 & 0.9B & {9.56}&0.313\\
 SiD-LSG ($\kappa=3$) &512& 0.09s & 1 & 0.9B & 13.21 & 0.314\\
 SiD-LSG ($\kappa=4.5$) &512& 0.09s & 1 & 0.9B & 16.59 & \textbf{0.317}\\
   \midrule
  \multicolumn{7}{c}{Stable Diffusion 2.1-base and its  distilled versions}\\
  SD2.1-base~\citep{rombach2022high} & 512 & 0.09s & 1 & 0.9B & 202.14& 0.08\\
    SD2.1 base~\citep{rombach2022high} & 512 & 0.77s & 25 & 0.9B & 13.45& 0.30\\
         \hdashline
 SwiftBrush~\citep{thuan2024swiftbrush}&512& 0.09s & 1 & 0.9B &16.67 &0.29\\
 SiD-LSG ($\kappa=1.5$) &512& 0.09s & 1 & 0.9B & \textbf{9.52} & 0.308\\
 SiD-LSG ($\kappa=2$) &512& 0.09s & 1 & 0.9B &10.97 &0.318\\
 SiD-LSG ($\kappa=3$) &512& 0.09s & 1 & 0.9B & 13.50 &0.321\\
 SiD-LSG ($\kappa=4.5$) &512& 0.09s & 1 & 0.9B & 16.54 & \textbf{0.322}\\
 \bottomrule
\end{tabular}}
\vspace{-4mm}
\end{table}

\begin{table}[t]
\centering
\caption{\small Comparison of HPSv2 score and Precision/Recall on the COCO-2014 validation set. The HPSv2 scores of ADD (SD-Turbo) are produced based on the publicly available model checkpoint. The HPSv2 scores of the other baselines are quoted from SwiftBrush \citep{thuan2024swiftbrush}. The Precision and Recall on COCO-2014 are obtained using the 30K images generated by the corresponding model checkpoints. %
}
\vspace{-2.5mm}
\label{tab:hpsv2}
\resizebox{.95\textwidth}{!}{
\begin{tabular}{cl|cccccc}
\toprule[1.5pt]
\multirow{2}{*}{Teacher} &\multirow{2}{*}{Student} & \multicolumn{4}{c}{Human Preference Score v2 $\uparrow$}     & \multicolumn{2}{c}{COCO-2014 Precision \& Recall $\uparrow$}                                                                      \\ \cline{3-8} 
                 &       & \multicolumn{1}{l}{Anime} & \multicolumn{1}{l}{Photo} & \multicolumn{1}{l}{Concept Art} & \multicolumn{1}{l}{Paintings} & \multicolumn{1}{c}{Precision} &\multicolumn{1}{c}{Recall}  \\ \midrule
SD2.1 &ADD (SD-Turbo)~\citep{Sauer2023AdversarialDD}                & \textbf{27.48}                     & {26.89}                     & \textbf{26.86}                           & 27.46   &  0.65 &  0.35 \\ 
SD1.5 & LCM~\citep{luo2023latentlora}                     & 22.61                     & 22.71                     & 22.74                           & 22.91  &  - & -                     \\
SD1.5 & InstaFlow~\citep{liu2023insta}               & {25.98}                     & {26.32}                     & {25.79}                           & {25.93}   &  0.53 & 0.45                      \\
SD1.5 & BOOT~\citep{gu2023boot}                    & 25.29                     & 25.16                     & 24.40                           & 24.61         &  - & -           \\
SD2.1-base & SwiftBrush~\citep{thuan2024swiftbrush}                    & 26.91                    & 27.21                    & 26.32                         & {26.37}  &   0.47   &  0.46                         \\ \midrule
\multirow{4}{*}{SD1.5} & SiD-LSG ($\kappa=1.5$)                       &    {26.58}                  & 26.80                     & {26.02}                          & 26.02        &     0.59       &      {0.52}           \\
& SiD-LSG ($\kappa=1.5$, double the training time)                       &    {26.58}                  & 26.80                     & 26.01                          & 26.02         &      0.60       &      \textbf{0.53}     \\ 
& SiD-LSG ($\kappa=2$)                       &    {26.94}                  & 27.03                     & {26.35 }                          & 26.27        &     0.64       &      {0.48}           \\ 
&SiD-LSG ($\kappa=3$)                       &    27.10                  & 27.11                   & 26.47                         & 26.46      &      0.65      &        0.40          \\ 
&SiD-LSG ($\kappa=4.5$)                       &    {27.39}                  & {27.30}                     & {26.65}                          & {26.58}      &     \textbf{0.67}      &        0.34       \\  \midrule
\multirow{3}{*}{SD2.1-base} & SiD-LSG ($\kappa=1.5$)                    &       26.65          &       26.87              &         26.19                &     26.14      &     0.60     &      0.49         \\ 
& SiD-LSG ($\kappa=2$)                    &    26.90                 & 27.08                     & {26.43 }                          & {26.47}        &     0.62       &      0.44           \\ 
&SiD-LSG ($\kappa=3$)                    &        27.27             &           27.22          &            {26.75}              &    {26.72}     &      0.64      &        0.38        \\ 
&SiD-LSG ($\kappa=4.5$)                    &        {27.42}             &           \textbf{27.31}          &            {26.81}              &    {26.79}     &      0.63      &        0.34        \\ 
\bottomrule[1.5pt]
\end{tabular}}
\vspace{-4mm}
\end{table}

\vspace{-1mm}
\subsection{Long and/or short guidance strategies}
\vspace{-1mm}
\textbf{No CFG. } 
Figure~\ref{fig:cfg_1114} demonstrates that without CFG, where \(\kappa_1 = \kappa_2 = \kappa_3 = \kappa_4 = 1\), results are underwhelming, highlighting the need for CFG in SiD to distill T2I diffusion models.

\textbf{Long the CFG of the pretrained score network. } 
For the long strategy, 
 we set \(\kappa_1=\kappa_2=\kappa_3=1\) and \(\kappa_4 > 1\), specifically exploring values of \(\kappa_4\) from the set \(\{2, 2.5, 3, 3.5, 7.5\}\). As shown in the left panel of Figure~\ref{fig:cfg_1114}, our experiments results indicate that this setup yields highly competitive performance, even after processing as few as 2.5 million fake images (approximately 5,000 iterations with a mini-batch size of 512). The parameter \(\kappa_4\) plays a crucial role in balancing FID and CLIP scores. For achieving lower FID scores, \(\kappa_4=2\) can yield an FID close to 10, while for higher CLIP scores, \(\kappa_4=7.5\) can result in a CLIP score around 0.31. However, our primary objective is to devise a strategy that lowers FID while minimizing CLIP degradation. %

\textbf{Short the CFG of the fake score network. }
For the short strategy, we set $\kappa_1=\kappa_4=1$ and explore $\kappa_2=\kappa_3\in \{0.5,0.125\}$. As illustrated in the right panel of Figure~\ref{fig:cfg_1114}, this approach delivers competitive performance, with $\kappa_2$ dictating the balance between FID and CLIP scores. However, compared to the long strategy, this configuration generally produces inferior results, as suggested by lower CLIP scores when FIDs are controlled to similar levels.

\textbf{LSG: Long and Short classifier-free Guidance. } 
Below, we explore how to effectively integrate the long and short strategies to enhance text guidance in diffusion distillation. Initially, we discovered a ``simplest'' form of LSG strategy 
by amplifying the CFG during the training of $f_{\psi}$. 
Specifically, we set $\kappa_1$ to 2 or 3, while keeping $\kappa_2=\kappa_3=\kappa_4$ at 1. As illustrated in the right panel of Figure~\ref{fig:cfg_1114}, setting $\kappa_1=3$ (green lines) proves as effective as the short strategy with $\kappa_2=\kappa_3=0.125$ (red lines), and $\kappa_1=2$ (black lines) outperforms the short strategy with $\kappa_2=\kappa_3=0.5$ (blue lines) and is comparable to the long strategy with $\kappa_4=3$ (red lines in the left panel). These findings validate this ``simplest'' LSG as an effective guidance strategy in text-guided diffusion distillation.

Nevertheless, merely matching the performance of the best long or short strategy is insufficient to justify adopting this ``simplest'' LSG. The efficacy of LSG is notably enhanced when the CFG scale applied to the fake score network during its training exceeds 1 and is maintained throughout the generator's training. This strategy strikes an effective balance between minimizing FID and maximizing CLIP. Specifically, for this recommended LSG configuration, we evaluated $\kappa_1=\kappa_2=\kappa_3=\kappa_4$ values within $\{1.5, 2, 3, 4.5\}$, presenting the results in Figure~\ref{fig:lsg} for distilling both SD 1.5 and 2.1-base models.
Comparisons between SD1.5 outcomes in the left panel of Figure~\ref{fig:lsg} and those in Figure~\ref{fig:cfg_1114} demonstrate superior performance with this LSG setting, indicated by higher CLIP scores at controlled FID levels, and lower FID scores at controlled CLIP levels. Generally, within the range of 1.5 to 4.5, a lower guidance scale correlates with better FID but worse CLIP, and vice versa, as evidenced by the curves for both SD 1.5 and 2.1-base in Figure~\ref{fig:lsg}.

\vspace{-1mm}
\subsection{Ablation study}
\vspace{-1mm}

\begin{figure}[!t]
\centering
\includegraphics[width=.325\linewidth]{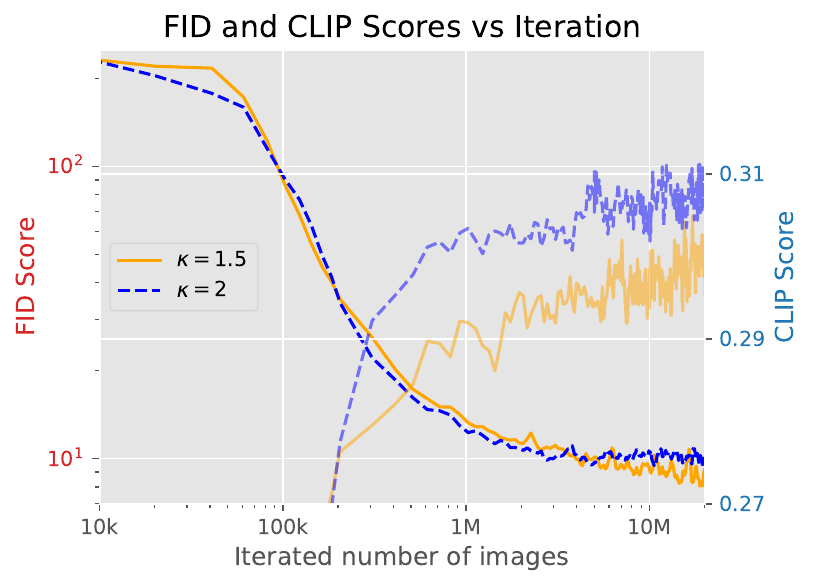}
\includegraphics[width=.325\linewidth]{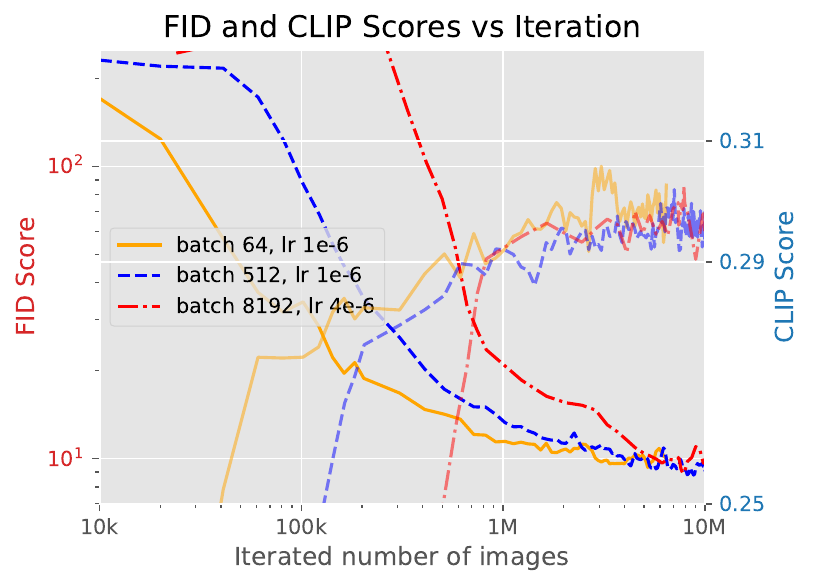}
\includegraphics[width=.325\linewidth]{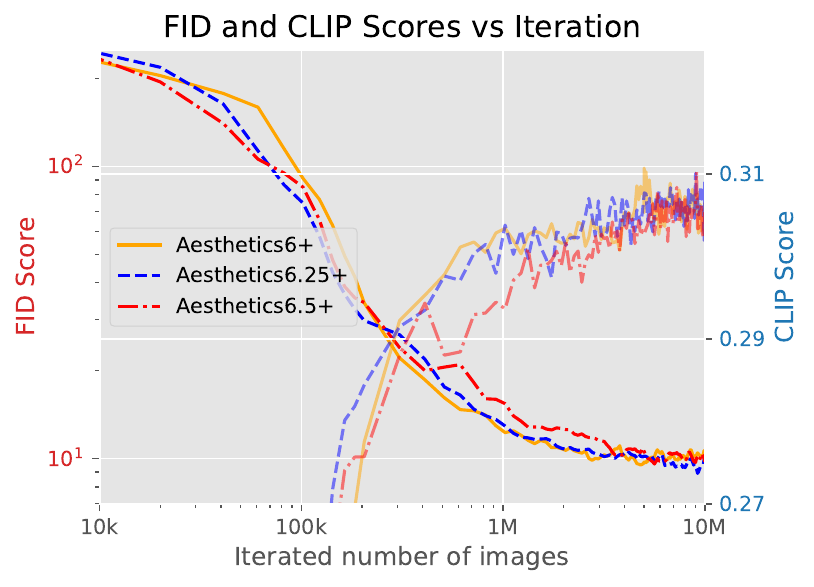}
 \vspace{-2mm}
\caption{\small This figure illustrates the progression of FID and CLIP scores during an ablation study of distilling SD1.5 using SiD-LSG. The default settings of batch size 512, learning rate 1e-6, LSG scale 2, and Prompt Aesthetics6+ are maintained unless specified otherwise. \textbf{Left}: The number of training fake images is doubled from 10M to 20M under LSG scales of 1.5 and 2.0. \textbf{Middle}: Variations in batch size and learning rate settings under LSG 1.5. \textbf{Right}: Comparison of training prompts Aesthetics6+, Aesthetics6.25+, and Aesthetics6.5+. 
 }
 \label{fig:ablation}
  \vspace{-3mm}
\end{figure}

We investigate the impact of extended training durations under two different guidance scales, variations in batch size, and the selection of training prompts. Initially, we doubled the number of fake images used to train the generator from 10M to 20M and monitored the evolution of the FID and CLIP scores. From the left panel of Figure \ref{fig:ablation}, we observe continuous improvements in both FID and CLIP scores for an LSG of 1.5 when training is extended beyond 10M fake images, and sustained enhancements in CLIP scores for an LSG of 2.0. Notably, by doubling the training input from 10M to 20M fake images, the FID under LSG 1.5 decreased from 8.71 to a record low of \textbf{8.15} among diffusion distillation methods in the data-free setting.

We assessed the effects of batch size by considering two additional settings: a batch size of 8192 with a learning rate of 4e-6, a configuration used in SiD to distill the EDM model pretrained on ImageNet 64x64 \citep{zhou2024score}, and a batch size of 64 with a learning rate of 1e-6. The middle panel of Figure \ref{fig:ablation} demonstrates that while there are initial differences in the convergence speed in terms of the number of fake images processed, the performances eventually converge to similar levels. We note that while smaller batch sizes may seem to converge faster, they require more time to process the same number of images. This is due to more frequent model parameter updates, higher communication costs between GPUs, and additional overheads typically associated with smaller batch sizes.

Lastly, we explored the effect of changing training prompts, shifting from the Aesthetics6+ prompts to Aesthetics6.25+ and Aesthetics6.5+ prompts \citep{cherti2023reproducible}. The performance, as shown in the right panel of Figure \ref{fig:ablation}, appears comparable across these variations. Specifically, under an LSG of 2.0, switching from Aesthetics6+ to Aesthetics6.25+ enabled us to further reduce the FID from the 9.56 reported in Table \ref{tab:comparison} to 9.21, although the CLIP score decreased slightly from 0.313 to 0.311, indicating no significant performance disparity between them.

The results of the ablation study show that SiD-LSG has low sensitivity to variations in batch size and training prompts and its performance could potentially be further enhanced with extended training.

\vspace{-1mm}
\subsection{Quantitative and qualitative evaluations}
\vspace{-1mm}

We present comprehensive results from prior studies across various experimental settings, including both one-step and multi-step generation methods. When evaluation results are available in existing literature~\citep{yin2023onestepDW,liu2023insta,kang2023scaling,thuan2024swiftbrush}, we directly cite them; otherwise, if model checkpoints are accessible, either publicly or provided upon request by the authors, we utilize the evaluation code from GigaGAN to produce the reported results. For our SiD-LSG, we select $\kappa_1=\kappa_2=\kappa_3=\kappa_4 \in \{1.5,2.0, 3.0, 4.5\}$. 

For comparisons of FID and CLIP scores, the results are detailed in Table~\ref{tab:comparison}. Among all one-step distillation methods, our approach notably excels in zero-shot text-conditioned image generation on the COCO-2014 dataset, as reflected by both FID-30K and CLIP scores. Specifically, with the guidance scale set as 2, our method attains FID scores as low as 9.56 and 10.97, and a CLIP score above 0.31 and around 0.32, using SD 1.5 and 2.1-base as the pretrained backbones, respectively. Notably, by setting the guidance scale to 1.5 and doubling the training time, our method achieves a record-low data-free  FID of \textbf{8.15}, along with a CLIP score of 0.304, when distilling SD1.5. These results remain highly competitive when compared to other generative approaches, such as autoregressive models and GANs, and are even comparable to previous multi-step diffusion-based sampling methods. Analyzing different $\kappa$ values, we observe a trade-off between FID and CLIP scores: smaller $\kappa$ values generally yield better FID metrics, while larger values enhance CLIP scores, aligning with past findings on the impact of guidance scale.

Beyond FID and CLIP scores, we also assess Precision and Recall~\citep{kynkaanniemi2019improved} as well as Human Preference Score (HPSv2)~\citep{wu2023human}, which are presented in Table~\ref{tab:hpsv2}. We reuse the same 30k images from previous evaluations for the Precision and Recall calculations. For HPSv2, we follow their established protocol, generating images from 800 text prompts per category. Except for ADD trained with real data and adversarial loss,  our SiD-LSG models outperform other baselines in HPSv2 scores across all categories, as well as in Precision and Recall metrics. Notably, with $\kappa=4.5$, SiD-LSG reaches peak performance in distilling SD2.1-base. Regarding Precision and Recall, higher $\kappa$ values lead to improved Precision, while lower values result in better Recall.

\begin{figure}[!t]
    \centering
    \includegraphics[width=0.9\textwidth]{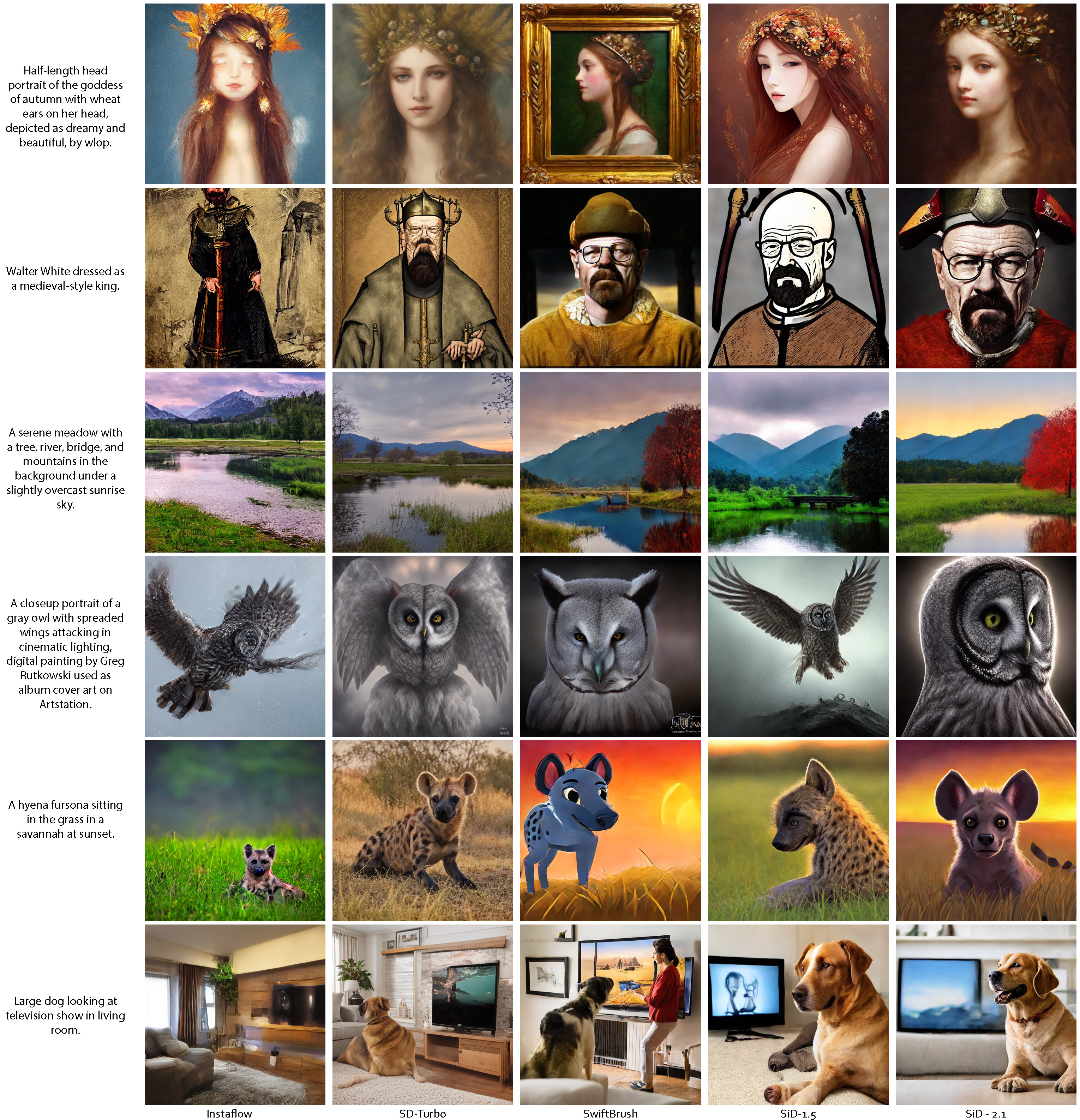}
    \vspace{-2mm}
    \caption{\small Qualitative comparison of one-step distillation methods using identical text prompts and random~seeds.}
    \label{fig:qualitative}
    \vspace{-2mm}
\end{figure}

For qualitative verification, we utilize our models with $\kappa=4.5$, selecting six prompts from across all HPSv2 categories to generate images. To ensure a fair comparison, we maintain the same random seed for image generation across all methods. The visual results are illustrated in Figure \ref{fig:qualitative}, where SiD consistently shows superior text-image alignment and visual fidelity.

To contextualize the numerical differences in metrics such as FID, CLIP, and HPSv2, 
we present Figures \ref{fig:qualitative_1} and \ref{fig:qualitative_2} to illustrate 
their implications %
for visual perception: the model with the best FID excels in diversity, while the model with the best CLIP stands out in text alignment and aesthetic~quality.

For broader impact, please refer to the discussion in Appendix~\ref{sec:broaderimpact}.
For limitations and computational requirements, please refer to a detailed discussion in Appendix~\ref{sec:3.3}.

\section{Conclusion and Future Work} \label{sec:conclusion}

This paper introduces a novel data-free method combining Classifier-Free Guidance (CFG) with Score identity Distillation (SiD) to efficiently distill Stable Diffusion models into effective one-step generators. By leveraging our innovative Long and Short CFG strategies (LSG), we distilled these models using only synthetic images generated by the one-step generator. This approach not only validates the practical potential of SiD but also sets new benchmarks for data-free one-step diffusion distillation, achieving remarkable zero-shot FID scores on the COCO-2014 validation set. Our method enhances efficiency while maintaining generation performance, allowing learning from the teacher model without the need for real images or the inclusion of additional regression or adversarial losses. We will make our code and distilled models publicly available to facilitate further research.

In the data-free setting, we are exploring the use of SiD-LSG for privacy and security-sensitive tasks where access to actual training data is not feasible. While we have advanced the capabilities of data-free diffusion distillation, our baseline methods such as ADD and DMD, typically require the use of real or teacher-synthesized images. Moving forward, we plan to lift the data-free constraint and integrate SiD-LSG with Diffusion GAN-based adversarial training, which has successfully transformed pretrained unconditional and label-conditional diffusion models into one-step generators, achieving state-of-the-art generation performance without CFG \citep{zhou2025adversarial}.
 This transition will involve the use of real images to not only further enhance photo-realism and improve text alignment but also adapt SiD-LSG distilled one-step generators to domains that differ from those used to train the teacher. This approach aims to broaden the applicability and effectiveness of our SiD-LSG distilled diffusion models.

\section*{Acknowledgments}
M. Zhou, Z. Wang, and H. Zheng  acknowledge the support of NSF-IIS 2212418 and NIH-R37 CA271186.

\small

\bibliographystyle{iclr2025_conference}
\bibliography{reference.bib,ref.bib,References052016}

\begin{thebibliography}{123}
\providecommand{\natexlab}[1]{#1}
\providecommand{\url}[1]{\texttt{#1}}
\expandafter\ifx\csname urlstyle\endcsname\relax
  \providecommand{\doi}[1]{doi: #1}\else
  \providecommand{\doi}{doi: \begingroup \urlstyle{rm}\Url}\fi

\bibitem[Achiam et~al.(2023)Achiam, Adler, Agarwal, Ahmad, Akkaya, Aleman,
  Almeida, Altenschmidt, Altman, Anadkat, et~al.]{achiam2023gpt}
Josh Achiam, Steven Adler, Sandhini Agarwal, Lama Ahmad, Ilge Akkaya,
  Florencia~Leoni Aleman, Diogo Almeida, Janko Altenschmidt, Sam Altman,
  Shyamal Anadkat, et~al.
\newblock {GPT}-4 technical report.
\newblock \emph{arXiv preprint arXiv:2303.08774}, 2023.

\bibitem[Albergo et~al.(2023)Albergo, Boffi, and
  Vanden-Eijnden]{albergo2023stochastic}
Michael~S Albergo, Nicholas~M Boffi, and Eric Vanden-Eijnden.
\newblock Stochastic interpolants: A unifying framework for flows and
  diffusions.
\newblock \emph{arXiv preprint arXiv:2303.08797}, 2023.

\bibitem[Austin et~al.(2021)Austin, Johnson, Ho, Tarlow, and van~den
  Berg]{austin2021structured}
Jacob Austin, Daniel~D Johnson, Jonathan Ho, Daniel Tarlow, and Rianne van~den
  Berg.
\newblock Structured denoising diffusion models in discrete state-spaces.
\newblock \emph{Advances in Neural Information Processing Systems},
  34:\penalty0 17981--17993, 2021.

\bibitem[Balaji et~al.(2022)Balaji, Nah, Huang, Vahdat, Song, Zhang, Kreis,
  Aittala, Aila, Laine, Catanzaro, Karras, and Liu]{balaji2022ediffi}
Yogesh Balaji, Seungjun Nah, Xun Huang, Arash Vahdat, Jiaming Song, Qinsheng
  Zhang, Karsten Kreis, Miika Aittala, Timo Aila, Samuli Laine, Bryan
  Catanzaro, Tero Karras, and Ming-Yu Liu.
\newblock {eDiff-I}: Text-to-image diffusion models with an ensemble of expert
  denoisers.
\newblock \emph{arXiv preprint arXiv:2211.01324}, 2022.

\bibitem[Brown et~al.(2020)Brown, Mann, Ryder, Subbiah, Kaplan, Dhariwal,
  Neelakantan, Shyam, Sastry, Askell, Agarwal, Herbert-Voss, Krueger, Henighan,
  Child, Ramesh, Ziegler, Wu, Winter, Hesse, Chen, Sigler, Litwin, Gray, Chess,
  Clark, Berner, McCandlish, Radford, Sutskever, and Amodei]{brown2020language}
Tom Brown, Benjamin Mann, Nick Ryder, Melanie Subbiah, Jared~D Kaplan, Prafulla
  Dhariwal, Arvind Neelakantan, Pranav Shyam, Girish Sastry, Amanda Askell,
  Sandhini Agarwal, Ariel Herbert-Voss, Gretchen Krueger, Tom Henighan, Rewon
  Child, Aditya Ramesh, Daniel Ziegler, Jeffrey Wu, Clemens Winter, Chris
  Hesse, Mark Chen, Eric Sigler, Mateusz Litwin, Scott Gray, Benjamin Chess,
  Jack Clark, Christopher Berner, Sam McCandlish, Alec Radford, Ilya Sutskever,
  and Dario Amodei.
\newblock Language models are few-shot learners.
\newblock \emph{Advances in neural information processing systems},
  33:\penalty0 1877--1901, 2020.

\bibitem[Chang et~al.(2023)Chang, Zhang, Barber, Maschinot, Lezama, Jiang,
  Yang, Murphy, Freeman, Rubinstein, Li, and Krishnan]{chang2023muse}
Huiwen Chang, Han Zhang, Jarred Barber, Aaron Maschinot, Jose Lezama, Lu~Jiang,
  Ming-Hsuan Yang, Kevin~Patrick Murphy, William~T. Freeman, Michael
  Rubinstein, Yuanzhen Li, and Dilip Krishnan.
\newblock Muse: Text-to-image generation via masked generative transformers.
\newblock In Andreas Krause, Emma Brunskill, Kyunghyun Cho, Barbara Engelhardt,
  Sivan Sabato, and Jonathan Scarlett (eds.), \emph{Proceedings of the 40th
  International Conference on Machine Learning}, volume 202 of
  \emph{Proceedings of Machine Learning Research}, pp.\  4055--4075. PMLR,
  23--29 Jul 2023.
\newblock URL \url{https://proceedings.mlr.press/v202/chang23b.html}.

\bibitem[Changpinyo et~al.(2021)Changpinyo, Sharma, Ding, and
  Soricut]{changpinyo2021cc12m}
Soravit Changpinyo, Piyush Sharma, Nan Ding, and Radu Soricut.
\newblock {Conceptual 12M}: Pushing web-scale image-text pre-training to
  recognize long-tail visual concepts.
\newblock In \emph{CVPR}, 2021.

\bibitem[Chen \& Zhou(2023)Chen and Zhou]{chen2023learning}
Tianqi Chen and Mingyuan Zhou.
\newblock Learning to {J}ump: Thinning and thickening latent counts for
  generative modeling.
\newblock In \emph{ICML 2023: International Conference on Machine Learning},
  July 2023.
\newblock URL \url{http://arxiv.org/abs/2305.18375}.

\bibitem[Cherti et~al.(2023)Cherti, Beaumont, Wightman, Wortsman, Ilharco,
  Gordon, Schuhmann, Schmidt, and Jitsev]{cherti2023reproducible}
Mehdi Cherti, Romain Beaumont, Ross Wightman, Mitchell Wortsman, Gabriel
  Ilharco, Cade Gordon, Christoph Schuhmann, Ludwig Schmidt, and Jenia Jitsev.
\newblock Reproducible scaling laws for contrastive language-image learning.
\newblock In \emph{Proceedings of the IEEE/CVF Conference on Computer Vision
  and Pattern Recognition}, pp.\  2818--2829, 2023.

\bibitem[Chong et~al.(2009)Chong, Blei, and Li]{chong2009simultaneous}
Wang Chong, David Blei, and Fei-Fei Li.
\newblock Simultaneous image classification and annotation.
\newblock In \emph{2009 IEEE Conference on computer vision and pattern
  recognition}, pp.\  1903--1910. IEEE, 2009.

\bibitem[Dao(2023)]{dao2023flashattention}
Tri Dao.
\newblock Flashattention-2: Faster attention with better parallelism and work
  partitioning.
\newblock \emph{arXiv preprint arXiv:2307.08691}, 2023.

\bibitem[Dao et~al.(2022)Dao, Fu, Ermon, Rudra, and
  R{\'e}]{dao2022flashattention}
Tri Dao, Dan Fu, Stefano Ermon, Atri Rudra, and Christopher R{\'e}.
\newblock Flashattention: Fast and memory-efficient exact attention with
  io-awareness.
\newblock \emph{Advances in Neural Information Processing Systems},
  35:\penalty0 16344--16359, 2022.

\bibitem[Devlin et~al.(2018)Devlin, Chang, Lee, and Toutanova]{devlin2018bert}
Jacob Devlin, Ming-Wei Chang, Kenton Lee, and Kristina Toutanova.
\newblock {BERT}: Pre-training of deep bidirectional transformers for language
  understanding.
\newblock \emph{arXiv preprint arXiv:1810.04805}, 2018.

\bibitem[Dhariwal \& Nichol(2021)Dhariwal and Nichol]{dhariwal2021diffusion}
Prafulla Dhariwal and Alexander Nichol.
\newblock Diffusion models beat {GAN}s on image synthesis.
\newblock \emph{Advances in Neural Information Processing Systems},
  34:\penalty0 8780--8794, 2021.

\bibitem[Ding et~al.(2021)Ding, Yang, Hong, Zheng, Zhou, Yin, Lin, Zou, Shao,
  Yang, et~al.]{ding2021cogview}
Ming Ding, Zhuoyi Yang, Wenyi Hong, Wendi Zheng, Chang Zhou, Da~Yin, Junyang
  Lin, Xu~Zou, Zhou Shao, Hongxia Yang, et~al.
\newblock Cogview: Mastering text-to-image generation via transformers.
\newblock \emph{Advances in Neural Information Processing Systems},
  34:\penalty0 19822--19835, 2021.

\bibitem[Esser et~al.(2021)Esser, Rombach, and Ommer]{esser2021taming}
Patrick Esser, Robin Rombach, and Bjorn Ommer.
\newblock Taming transformers for high-resolution image synthesis.
\newblock In \emph{Proceedings of the IEEE/CVF conference on computer vision
  and pattern recognition}, pp.\  12873--12883, 2021.

\bibitem[Fei-Fei \& Perona(2005)Fei-Fei and Perona]{fei2005bayesian}
Li~Fei-Fei and Pietro Perona.
\newblock A {B}ayesian hierarchical model for learning natural scene
  categories.
\newblock In \emph{2005 IEEE computer society conference on computer vision and
  pattern recognition (CVPR'05)}, volume~2, pp.\  524--531. IEEE, 2005.

\bibitem[Gafni et~al.(2022)Gafni, Polyak, Ashual, Sheynin, Parikh, and
  Taigman]{gafni2022make}
Oran Gafni, Adam Polyak, Oron Ashual, Shelly Sheynin, Devi Parikh, and Yaniv
  Taigman.
\newblock Make-a-scene: Scene-based text-to-image generation with human priors.
\newblock In \emph{European Conference on Computer Vision}, pp.\  89--106.
  Springer, 2022.

\bibitem[Goodfellow et~al.(2014)Goodfellow, Pouget-Abadie, Mirza, Xu,
  Warde-Farley, Ozair, Courville, and Bengio]{goodfellow2014generative}
Ian Goodfellow, Jean Pouget-Abadie, Mehdi Mirza, Bing Xu, David Warde-Farley,
  Sherjil Ozair, Aaron Courville, and Yoshua Bengio.
\newblock Generative adversarial nets.
\newblock In \emph{Advances in Neural Information Processing Systems}, pp.\
  2672--2680, 2014.

\bibitem[Gregor et~al.(2015)Gregor, Danihelka, Graves, Rezende, and
  Wierstra]{gregor2015draw}
Karol Gregor, Ivo Danihelka, Alex Graves, Danilo Rezende, and Daan Wierstra.
\newblock {DRAW}: A recurrent neural network for image generation.
\newblock In \emph{International conference on machine learning}, pp.\
  1462--1471. PMLR, 2015.

\bibitem[Gu et~al.(2023)Gu, Zhai, Zhang, Liu, and Susskind]{gu2023boot}
Jiatao Gu, Shuangfei Zhai, Yizhe Zhang, Lingjie Liu, and Joshua~M Susskind.
\newblock {BOOT}: Data-free distillation of denoising diffusion models with
  bootstrapping.
\newblock In \emph{ICML 2023 Workshop on Structured Probabilistic Inference and
  Generative Modeling}, 2023.

\bibitem[Gu et~al.(2022)Gu, Chen, Bao, Wen, Zhang, Chen, Yuan, and
  Guo]{gu2021vector}
Shuyang Gu, Dong Chen, Jianmin Bao, Fang Wen, Bo~Zhang, Dongdong Chen, Lu~Yuan,
  and Baining Guo.
\newblock Vector quantized diffusion model for text-to-image synthesis.
\newblock In \emph{CVPR}, 2022.

\bibitem[He et~al.(2020)He, Liu, Gao, and Chen]{he2020deberta}
Pengcheng He, Xiaodong Liu, Jianfeng Gao, and Weizhu Chen.
\newblock {DeBERTa}: Decoding-enhanced bert with disentangled attention.
\newblock \emph{arXiv preprint arXiv:2006.03654}, 2020.

\bibitem[Heusel et~al.(2017)Heusel, Ramsauer, Unterthiner, Nessler, and
  Hochreiter]{heusel2017gans}
Martin Heusel, Hubert Ramsauer, Thomas Unterthiner, Bernhard Nessler, and Sepp
  Hochreiter.
\newblock {GAN}s trained by a two time-scale update rule converge to a local
  {N}ash equilibrium.
\newblock In \emph{Advances in Neural Information Processing Systems}, pp.\
  6626--6637, 2017.

\bibitem[Hinton et~al.(2006)Hinton, Osindero, and Teh]{hinton2006fast}
G.~Hinton, S.~Osindero, and Y.-W. Teh.
\newblock A fast learning algorithm for deep belief nets.
\newblock \emph{Neural Computation}, 18\penalty0 (7):\penalty0 1527--1554,
  2006.

\bibitem[Ho \& Salimans(2022)Ho and Salimans]{ho2022classifier}
Jonathan Ho and Tim Salimans.
\newblock Classifier-free diffusion guidance.
\newblock \emph{arXiv preprint arXiv:2207.12598}, 2022.

\bibitem[Ho et~al.(2020)Ho, Jain, and Abbeel]{ho2020denoising}
Jonathan Ho, Ajay Jain, and Pieter Abbeel.
\newblock Denoising {D}iffusion {P}robabilistic {M}odels.
\newblock \emph{Advances in Neural Information Processing Systems}, 33, 2020.

\bibitem[Ho et~al.(2022)Ho, Chan, Saharia, Whang, Gao, Gritsenko, Kingma,
  Poole, Norouzi, Fleet, and Salimans]{ho2022imagen}
Jonathan Ho, William Chan, Chitwan Saharia, Jay Whang, Ruiqi Gao, Alexey
  Gritsenko, Diederik~P. Kingma, Ben Poole, Mohammad Norouzi, David~J. Fleet,
  and Tim Salimans.
\newblock Imagen video: High definition video generation with diffusion models.
\newblock \emph{arXiv preprint arXiv:2210.02303}, 2022.

\bibitem[Holmes \& Walker(2017)Holmes and Walker]{holmes2017assigning}
Chris~C Holmes and Stephen~G Walker.
\newblock Assigning a value to a power likelihood in a general {B}ayesian
  model.
\newblock \emph{Biometrika}, 104\penalty0 (2):\penalty0 497--503, 2017.

\bibitem[Hoogeboom et~al.(2021)Hoogeboom, Nielsen, Jaini, Forr{\'e}, and
  Welling]{hoogeboom2021argmax}
Emiel Hoogeboom, Didrik Nielsen, Priyank Jaini, Patrick Forr{\'e}, and Max
  Welling.
\newblock Argmax flows and multinomial diffusion: Learning categorical
  distributions.
\newblock \emph{Advances in Neural Information Processing Systems},
  34:\penalty0 12454--12465, 2021.

\bibitem[Hu et~al.(2022)Hu, Wang, Cham, Yang, and Suganthan]{hu2022global}
Minghui Hu, Yujie Wang, Tat-Jen Cham, Jianfei Yang, and Ponnuthurai~N
  Suganthan.
\newblock Global context with discrete diffusion in vector quantised modelling
  for image generation.
\newblock In \emph{Proceedings of the IEEE/CVF Conference on Computer Vision
  and Pattern Recognition}, pp.\  11502--11511, 2022.

\bibitem[Ilharco et~al.(2021)Ilharco, Wortsman, Wightman, Gordon, Carlini,
  Taori, Dave, Shankar, Namkoong, Miller, Hajishirzi, Farhadi, and
  Schmidt]{ilharco_gabriel_2021_5143773}
Gabriel Ilharco, Mitchell Wortsman, Ross Wightman, Cade Gordon, Nicholas
  Carlini, Rohan Taori, Achal Dave, Vaishaal Shankar, Hongseok Namkoong, John
  Miller, Hannaneh Hajishirzi, Ali Farhadi, and Ludwig Schmidt.
\newblock Openclip, July 2021.
\newblock URL \url{https://doi.org/10.5281/zenodo.5143773}.
\newblock If you use this software, please cite it as below.

\bibitem[Jiang et~al.(2023)Jiang, Sablayrolles, Mensch, Bamford, Chaplot,
  de~las Casas, Bressand, Lengyel, Lample, Saulnier, Lavaud, Lachaux, Stock,
  Scao, Lavril, Wang, Lacroix, and Sayed]{jiang2023mistral}
Albert~Q. Jiang, Alexandre Sablayrolles, Arthur Mensch, Chris Bamford,
  Devendra~Singh Chaplot, Diego de~las Casas, Florian Bressand, Gianna Lengyel,
  Guillaume Lample, Lucile Saulnier, Lélio~Renard Lavaud, Marie-Anne Lachaux,
  Pierre Stock, Teven~Le Scao, Thibaut Lavril, Thomas Wang, Timothée Lacroix,
  and William~El Sayed.
\newblock Mistral 7{B}.
\newblock \emph{arXiv preprint arXiv:2310.06825}, 2023.

\bibitem[Kang et~al.(2023)Kang, Zhu, Zhang, Park, Shechtman, Paris, and
  Park]{kang2023scaling}
Minguk Kang, Jun-Yan Zhu, Richard Zhang, Jaesik Park, Eli Shechtman, Sylvain
  Paris, and Taesung Park.
\newblock Scaling up {GANs} for text-to-image synthesis.
\newblock In \emph{Proceedings of the IEEE/CVF Conference on Computer Vision
  and Pattern Recognition}, pp.\  10124--10134, 2023.

\bibitem[Karras et~al.(2019)Karras, Laine, and Aila]{karras2019style}
Tero Karras, Samuli Laine, and Timo Aila.
\newblock A style-based generator architecture for generative adversarial
  networks.
\newblock In \emph{Proceedings of the IEEE/CVF conference on computer vision
  and pattern recognition}, pp.\  4401--4410, 2019.

\bibitem[Karras et~al.(2022)Karras, Aittala, Aila, and
  Laine]{karras2022elucidating}
Tero Karras, Miika Aittala, Timo Aila, and Samuli Laine.
\newblock Elucidating the design space of diffusion-based generative models.
\newblock In Alice~H. Oh, Alekh Agarwal, Danielle Belgrave, and Kyunghyun Cho
  (eds.), \emph{Advances in Neural Information Processing Systems}, 2022.
\newblock URL \url{https://openreview.net/forum?id=k7FuTOWMOc7}.

\bibitem[Kim et~al.(2023)Kim, Lai, Liao, Murata, Takida, Uesaka, He, Mitsufuji,
  and Ermon]{kim2023consistency}
Dongjun Kim, Chieh-Hsin Lai, Wei-Hsiang Liao, Naoki Murata, Yuhta Takida,
  Toshimitsu Uesaka, Yutong He, Yuki Mitsufuji, and Stefano Ermon.
\newblock Consistency trajectory models: Learning probability flow ode
  trajectory of diffusion.
\newblock \emph{arXiv preprint arXiv:2310.02279}, 2023.

\bibitem[Kingma \& Welling(2014)Kingma and Welling]{kingma2013auto}
Diederik~P. Kingma and Max Welling.
\newblock Auto-encoding variational {B}ayes.
\newblock In \emph{International Conference on Learning Representations}, 2014.

\bibitem[Kynk{\"a}{\"a}nniemi et~al.(2019)Kynk{\"a}{\"a}nniemi, Karras, Laine,
  Lehtinen, and Aila]{kynkaanniemi2019improved}
Tuomas Kynk{\"a}{\"a}nniemi, Tero Karras, Samuli Laine, Jaakko Lehtinen, and
  Timo Aila.
\newblock Improved precision and recall metric for assessing generative models.
\newblock \emph{Advances in neural information processing systems}, 32, 2019.

\bibitem[Lefaudeux et~al.(2022)Lefaudeux, Massa, Liskovich, Xiong, Caggiano,
  Naren, Xu, Hu, Tintore, Zhang, Labatut, Haziza, Wehrstedt, Reizenstein, and
  Sizov]{xFormers2022}
Benjamin Lefaudeux, Francisco Massa, Diana Liskovich, Wenhan Xiong, Vittorio
  Caggiano, Sean Naren, Min Xu, Jieru Hu, Marta Tintore, Susan Zhang, Patrick
  Labatut, Daniel Haziza, Luca Wehrstedt, Jeremy Reizenstein, and Grigory
  Sizov.
\newblock {xFormers}: A modular and hackable {T}ransformer modelling library.
\newblock \url{https://github.com/facebookresearch/xformers}, 2022.

\bibitem[Lipman et~al.(2022)Lipman, Chen, Ben-Hamu, Nickel, and
  Le]{lipman2022flow}
Yaron Lipman, Ricky~TQ Chen, Heli Ben-Hamu, Maximilian Nickel, and Matt Le.
\newblock Flow matching for generative modeling.
\newblock \emph{arXiv preprint arXiv:2210.02747}, 2022.

\bibitem[Liu et~al.(2022{\natexlab{a}})Liu, Ren, Lin, and Zhao]{liu2022pseudo}
Luping Liu, Yi~Ren, Zhijie Lin, and Zhou Zhao.
\newblock Pseudo numerical methods for diffusion models on manifolds.
\newblock In \emph{International Conference on Learning Representations},
  2022{\natexlab{a}}.
\newblock URL \url{https://openreview.net/forum?id=PlKWVd2yBkY}.

\bibitem[Liu et~al.(2022{\natexlab{b}})Liu, Gong, and Liu]{liu2022flow}
Xingchao Liu, Chengyue Gong, and Qiang Liu.
\newblock Flow straight and fast: Learning to generate and transfer data with
  rectified flow.
\newblock \emph{arXiv preprint arXiv:2209.03003}, 2022{\natexlab{b}}.

\bibitem[Liu et~al.(2023)Liu, Zhang, Ma, Peng, and Liu]{liu2023insta}
Xingchao Liu, Xiwen Zhang, Jianzhu Ma, Jian Peng, and Qiang Liu.
\newblock Instaflow: One step is enough for high-quality diffusion-based
  text-to-image generation.
\newblock \emph{arXiv preprint arXiv:2309.06380}, 2023.

\bibitem[Lu et~al.(2022{\natexlab{a}})Lu, Zhou, Bao, Chen, Li, and
  Zhu]{lu2022dpm++}
Cheng Lu, Yuhao Zhou, Fan Bao, Jianfei Chen, Chongxuan Li, and Jun Zhu.
\newblock {DPM-S}olver++: Fast solver for guided sampling of diffusion
  probabilistic models.
\newblock In \emph{arXiv preprint arXiv:2211.01095}, 2022{\natexlab{a}}.

\bibitem[Lu et~al.(2022{\natexlab{b}})Lu, Zhou, Bao, Chen, Li, and
  Zhu]{lu2022dpmsolver}
Cheng Lu, Yuhao Zhou, Fan Bao, Jianfei Chen, Chongxuan Li, and Jun Zhu.
\newblock {DPM}-solver: A fast {ODE} solver for diffusion probabilistic model
  sampling in around 10 steps.
\newblock In Alice~H. Oh, Alekh Agarwal, Danielle Belgrave, and Kyunghyun Cho
  (eds.), \emph{Advances in Neural Information Processing Systems},
  2022{\natexlab{b}}.
\newblock URL \url{https://openreview.net/forum?id=2uAaGwlP_V}.

\bibitem[Luhman \& Luhman(2021)Luhman and Luhman]{luhman2021knowledge}
Eric Luhman and Troy Luhman.
\newblock Knowledge distillation in iterative generative models for improved
  sampling speed.
\newblock \emph{arXiv preprint arXiv:2101.02388}, 2021.

\bibitem[Luo et~al.(2023{\natexlab{a}})Luo, Tan, Huang, Li, and
  Zhao]{Luo2023LatentCM}
Simian Luo, Yiqin Tan, Longbo Huang, Jian Li, and Hang Zhao.
\newblock Latent consistency models: Synthesizing high-resolution images with
  few-step inference.
\newblock \emph{ArXiv}, abs/2310.04378, 2023{\natexlab{a}}.

\bibitem[Luo et~al.(2023{\natexlab{b}})Luo, Tan, Patil, Gu, von Platen, Passos,
  Huang, Li, and Zhao]{luo2023latentlora}
Simian Luo, Yiqin Tan, Suraj Patil, Daniel Gu, Patrick von Platen, Apolinário
  Passos, Longbo Huang, Jian Li, and Hang Zhao.
\newblock Lcm-lora: A universal stable-diffusion acceleration module.
\newblock \emph{arXiv preprint arXiv:2310.04378}, 2023{\natexlab{b}}.

\bibitem[Luo et~al.(2023{\natexlab{c}})Luo, Hu, Zhang, Sun, Li, and
  Zhang]{luo2023diffinstruct}
Weijian Luo, Tianyang Hu, Shifeng Zhang, Jiacheng Sun, Zhenguo Li, and Zhihua
  Zhang.
\newblock {Diff-I}nstruct: A universal approach for transferring knowledge from
  pre-trained diffusion models.
\newblock In \emph{Thirty-seventh Conference on Neural Information Processing
  Systems}, 2023{\natexlab{c}}.
\newblock URL \url{https://openreview.net/forum?id=MLIs5iRq4w}.

\bibitem[Lyu(2009)]{lyu2009interpretation}
Siwei Lyu.
\newblock Interpretation and generalization of score matching.
\newblock In \emph{Proceedings of the Twenty-Fifth Conference on Uncertainty in
  Artificial Intelligence}, pp.\  359--366, 2009.

\bibitem[Lyu et~al.(2022)Lyu, Xu, Yang, Lin, and Dai]{lyu2022accelerating}
Zhaoyang Lyu, Xudong Xu, Ceyuan Yang, Dahua Lin, and Bo~Dai.
\newblock Accelerating diffusion models via early stop of the diffusion
  process.
\newblock \emph{arXiv preprint arXiv:2205.12524}, 2022.

\bibitem[Mansimov et~al.(2015)Mansimov, Parisotto, Ba, and
  Salakhutdinov]{mansimov2015generating}
Elman Mansimov, Emilio Parisotto, Jimmy~Lei Ba, and Ruslan Salakhutdinov.
\newblock Generating images from captions with attention.
\newblock \emph{arXiv preprint arXiv:1511.02793}, 2015.

\bibitem[Meng et~al.(2023)Meng, Rombach, Gao, Kingma, Ermon, Ho, and
  Salimans]{meng2023distillation}
Chenlin Meng, Robin Rombach, Ruiqi Gao, Diederik Kingma, Stefano Ermon,
  Jonathan Ho, and Tim Salimans.
\newblock On distillation of guided diffusion models.
\newblock In \emph{Proceedings of the IEEE/CVF Conference on Computer Vision
  and Pattern Recognition}, pp.\  14297--14306, 2023.

\bibitem[Mohamed \& Lakshminarayanan(2016)Mohamed and
  Lakshminarayanan]{mohamed2016learning}
Shakir Mohamed and Balaji Lakshminarayanan.
\newblock Learning in implicit generative models.
\newblock \emph{arXiv preprint arXiv:1610.03483}, 2016.

\bibitem[Nguyen \& Tran(2024)Nguyen and Tran]{thuan2024swiftbrush}
Thuan~Hoang Nguyen and Anh Tran.
\newblock {SwiftBrush}: One-step text-to-image diffusion model with variational
  score distillation.
\newblock In \emph{IEEE/CVF Conference on Computer Vision and Pattern
  Recognition (CVPR)}, 2024.

\bibitem[Nichol et~al.(2022)Nichol, Dhariwal, Ramesh, Shyam, Mishkin, Mcgrew,
  Sutskever, and Chen]{nichol2022glide}
Alexander~Quinn Nichol, Prafulla Dhariwal, Aditya Ramesh, Pranav Shyam, Pamela
  Mishkin, Bob Mcgrew, Ilya Sutskever, and Mark Chen.
\newblock Glide: Towards photorealistic image generation and editing with
  text-guided diffusion models.
\newblock In \emph{International Conference on Machine Learning}, pp.\
  16784--16804. PMLR, 2022.

\bibitem[Pandey et~al.(2022)Pandey, Mukherjee, Rai, and
  Kumar]{pandey2022diffusevae}
Kushagra Pandey, Avideep Mukherjee, Piyush Rai, and Abhishek Kumar.
\newblock {DiffuseVAE}: Efficient, controllable and high-fidelity generation
  from low-dimensional latents.
\newblock \emph{arXiv preprint arXiv:2201.00308}, 2022.

\bibitem[Papamakarios et~al.(2019)Papamakarios, Nalisnick, Rezende, Mohamed,
  and Lakshminarayanan]{papamakarios2019normalizing}
George Papamakarios, Eric Nalisnick, Danilo~Jimenez Rezende, Shakir Mohamed,
  and Balaji Lakshminarayanan.
\newblock Normalizing flows for probabilistic modeling and inference.
\newblock \emph{arXiv preprint arXiv:1912.02762}, 2019.

\bibitem[Peebles \& Xie(2023)Peebles and Xie]{peebles2023scalable}
William Peebles and Saining Xie.
\newblock Scalable diffusion models with transformers.
\newblock In \emph{Proceedings of the IEEE/CVF International Conference on
  Computer Vision}, pp.\  4195--4205, 2023.

\bibitem[Podell et~al.(2024)Podell, English, Lacey, Blattmann, Dockhorn,
  M{\"u}ller, Penna, and Rombach]{podell2024sdxl}
Dustin Podell, Zion English, Kyle Lacey, Andreas Blattmann, Tim Dockhorn, Jonas
  M{\"u}ller, Joe Penna, and Robin Rombach.
\newblock {SDXL}: Improving latent diffusion models for high-resolution image
  synthesis.
\newblock In \emph{The Twelfth International Conference on Learning
  Representations}, 2024.
\newblock URL \url{https://openreview.net/forum?id=di52zR8xgf}.

\bibitem[Polatkan et~al.(2015)Polatkan, Zhou, Carin, Blei, and
  Daubechies]{polatkan2015bayesian}
G.~Polatkan, M.~Zhou, L.~Carin, D.~Blei, and I.~Daubechies.
\newblock A {B}ayesian nonparametric approach to image super-resolution.
\newblock \emph{PAMI}, 37\penalty0 (2):\penalty0 346--358, 2015.

\bibitem[Poole et~al.(2022)Poole, Jain, Barron, and
  Mildenhall]{Poole2022DreamFusionTU}
Ben Poole, Ajay Jain, Jonathan~T. Barron, and Ben Mildenhall.
\newblock {DreamFusion}: Text-to-3{D} using 2{D} diffusion.
\newblock \emph{ArXiv}, abs/2209.14988, 2022.

\bibitem[Qin et~al.(2023)Qin, Zhang, Yu, Feng, Yang, Zhou, Wang, Niebles,
  Xiong, Savarese, et~al.]{qin2023unicontrol}
Can Qin, Shu Zhang, Ning Yu, Yihao Feng, Xinyi Yang, Yingbo Zhou, Huan Wang,
  Juan~Carlos Niebles, Caiming Xiong, Silvio Savarese, et~al.
\newblock Unicontrol: A unified diffusion model for controllable visual
  generation in the wild.
\newblock \emph{arXiv preprint arXiv:2305.11147}, 2023.

\bibitem[Radford et~al.(2018)Radford, Narasimhan, Salimans, Sutskever,
  et~al.]{radford2018improving}
Alec Radford, Karthik Narasimhan, Tim Salimans, Ilya Sutskever, et~al.
\newblock Improving language understanding by generative pre-training.
\newblock 2018.

\bibitem[Radford et~al.(2019)Radford, Wu, Child, Luan, Amodei, and
  Sutskever]{radford2019language}
Alec Radford, Jeff Wu, Rewon Child, David Luan, Dario Amodei, and Ilya
  Sutskever.
\newblock Language models are unsupervised multitask learners.
\newblock 2019.

\bibitem[Radford et~al.(2021)Radford, Kim, Hallacy, Ramesh, Goh, Agarwal,
  Sastry, Askell, Mishkin, Clark, et~al.]{radford2021learning}
Alec Radford, Jong~Wook Kim, Chris Hallacy, Aditya Ramesh, Gabriel Goh,
  Sandhini Agarwal, Girish Sastry, Amanda Askell, Pamela Mishkin, Jack Clark,
  et~al.
\newblock Learning transferable visual models from natural language
  supervision.
\newblock In \emph{International conference on machine learning}, pp.\
  8748--8763. PMLR, 2021.

\bibitem[Raffel et~al.(2020)Raffel, Shazeer, Roberts, Lee, Narang, Matena,
  Zhou, Li, and Liu]{raffel2020exploring}
Colin Raffel, Noam Shazeer, Adam Roberts, Katherine Lee, Sharan Narang, Michael
  Matena, Yanqi Zhou, Wei Li, and Peter~J Liu.
\newblock Exploring the limits of transfer learning with a unified text-to-text
  transformer.
\newblock \emph{Journal of machine learning research}, 21\penalty0
  (140):\penalty0 1--67, 2020.

\bibitem[Ramesh et~al.(2021)Ramesh, Pavlov, Goh, Gray, Voss, Radford, Chen, and
  Sutskever]{ramesh2021zero}
Aditya Ramesh, Mikhail Pavlov, Gabriel Goh, Scott Gray, Chelsea Voss, Alec
  Radford, Mark Chen, and Ilya Sutskever.
\newblock Zero-shot text-to-image generation.
\newblock In \emph{International Conference on Machine Learning}, pp.\
  8821--8831. PMLR, 2021.

\bibitem[Ramesh et~al.(2022)Ramesh, Dhariwal, Nichol, Chu, and
  Chen]{ramesh2022hierarchical}
Aditya Ramesh, Prafulla Dhariwal, Alex Nichol, Casey Chu, and Mark Chen.
\newblock Hierarchical text-conditional image generation with {CLIP} latents.
\newblock \emph{arXiv preprint arXiv:2204.06125}, 2022.

\bibitem[Reed et~al.(2016)Reed, Akata, Yan, Logeswaran, Schiele, and
  Lee]{reed2016generative}
Scott Reed, Zeynep Akata, Xinchen Yan, Lajanugen Logeswaran, Bernt Schiele, and
  Honglak Lee.
\newblock Generative adversarial text to image synthesis.
\newblock In \emph{International conference on machine learning}, pp.\
  1060--1069. PMLR, 2016.

\bibitem[Rezende et~al.(2014)Rezende, Mohamed, and
  Wierstra]{rezende2014stochastic}
Danilo~Jimenez Rezende, Shakir Mohamed, and Daan Wierstra.
\newblock Stochastic backpropagation and approximate inference in deep
  generative models.
\newblock In \emph{Proceedings of the 31st International Conference on Machine
  Learning}, pp.\  1278--1286, 2014.

\bibitem[Rolfe(2016)]{rolfe2016discrete}
Jason~Tyler Rolfe.
\newblock Discrete variational autoencoders.
\newblock \emph{arXiv preprint arXiv:1609.02200}, 2016.

\bibitem[Rombach et~al.(2022)Rombach, Blattmann, Lorenz, Esser, and
  Ommer]{rombach2022high}
Robin Rombach, Andreas Blattmann, Dominik Lorenz, Patrick Esser, and Bj{\"o}rn
  Ommer.
\newblock High-resolution image synthesis with latent diffusion models.
\newblock In \emph{Proceedings of the IEEE/CVF conference on computer vision
  and pattern recognition}, pp.\  10684--10695, 2022.

\bibitem[Saharia et~al.(2022)Saharia, Chan, Saxena, Li, Whang, Denton,
  Ghasemipour, Gontijo~Lopes, Karagol~Ayan, Salimans, Ho, Fleet, and
  Norouzi]{saharia2022photorealistic}
Chitwan Saharia, William Chan, Saurabh Saxena, Lala Li, Jay Whang, Emily~L
  Denton, Kamyar Ghasemipour, Raphael Gontijo~Lopes, Burcu Karagol~Ayan, Tim
  Salimans, Jonathan Ho, David~J Fleet, and Mohammad Norouzi.
\newblock Photorealistic text-to-image diffusion models with deep language
  understanding.
\newblock \emph{Advances in Neural Information Processing Systems},
  35:\penalty0 36479--36494, 2022.

\bibitem[Salakhutdinov \& Hinton(2009)Salakhutdinov and
  Hinton]{salakhutdinov2009deep}
Ruslan Salakhutdinov and Geoffrey Hinton.
\newblock Deep {B}oltzmann machines.
\newblock In \emph{Artificial intelligence and statistics}, pp.\  448--455.
  PMLR, 2009.

\bibitem[Salimans \& Ho(2022)Salimans and Ho]{salimans2022progressive}
Tim Salimans and Jonathan Ho.
\newblock Progressive distillation for fast sampling of diffusion models.
\newblock In \emph{International Conference on Learning Representations}, 2022.
\newblock URL \url{https://openreview.net/forum?id=TIdIXIpzhoI}.

\bibitem[Sauer et~al.(2022)Sauer, Schwarz, and Geiger]{sauer2022stylegan}
Axel Sauer, Katja Schwarz, and Andreas Geiger.
\newblock {StyleGAN-XL}: Scaling {StyleGAN} to large diverse datasets.
\newblock In \emph{ACM SIGGRAPH 2022 conference proceedings}, pp.\  1--10,
  2022.

\bibitem[Sauer et~al.(2023{\natexlab{a}})Sauer, Karras, Laine, Geiger, and
  Aila]{sauer2023stylegan}
Axel Sauer, Tero Karras, Samuli Laine, Andreas Geiger, and Timo Aila.
\newblock {StyleGAN-T}: Unlocking the power of {GAN}s for fast large-scale
  text-to-image synthesis.
\newblock \emph{arXiv preprint arXiv:2301.09515}, 2023{\natexlab{a}}.

\bibitem[Sauer et~al.(2023{\natexlab{b}})Sauer, Lorenz, Blattmann, and
  Rombach]{Sauer2023AdversarialDD}
Axel Sauer, Dominik Lorenz, A.~Blattmann, and Robin Rombach.
\newblock Adversarial diffusion distillation.
\newblock \emph{ArXiv}, abs/2311.17042, 2023{\natexlab{b}}.

\bibitem[Schuhmann et~al.(2022)Schuhmann, Beaumont, Vencu, Gordon, Wightman,
  Cherti, Coombes, Katta, Mullis, Wortsman, et~al.]{schuhmann2022laion}
Christoph Schuhmann, Romain Beaumont, Richard Vencu, Cade Gordon, Ross
  Wightman, Mehdi Cherti, Theo Coombes, Aarush Katta, Clayton Mullis, Mitchell
  Wortsman, et~al.
\newblock {LAION-5B}: An open large-scale dataset for training next generation
  image-text models.
\newblock \emph{Advances in Neural Information Processing Systems},
  35:\penalty0 25278--25294, 2022.

\bibitem[Sohl-Dickstein et~al.(2015)Sohl-Dickstein, Weiss, Maheswaranathan, and
  Ganguli]{sohl2015deep}
Jascha Sohl-Dickstein, Eric Weiss, Niru Maheswaranathan, and Surya Ganguli.
\newblock Deep unsupervised learning using nonequilibrium thermodynamics.
\newblock In \emph{International Conference on Machine Learning}, pp.\
  2256--2265. PMLR, 2015.

\bibitem[Song et~al.(2021{\natexlab{a}})Song, Meng, and
  Ermon]{song2021denoising}
Jiaming Song, Chenlin Meng, and Stefano Ermon.
\newblock Denoising diffusion implicit models.
\newblock In \emph{International Conference on Learning Representations},
  2021{\natexlab{a}}.
\newblock URL \url{https://openreview.net/forum?id=St1giarCHLP}.

\bibitem[Song \& Dhariwal(2023)Song and Dhariwal]{song2023improved}
Yang Song and Prafulla Dhariwal.
\newblock Improved techniques for training consistency models.
\newblock \emph{arXiv preprint arXiv:2310.14189}, 2023.

\bibitem[Song \& Ermon(2019)Song and Ermon]{song2019generative}
Yang Song and Stefano Ermon.
\newblock Generative {M}odeling by {E}stimating {G}radients of the {D}ata
  {D}istribution.
\newblock In \emph{Advances in Neural Information Processing Systems}, pp.\
  11918--11930, 2019.

\bibitem[Song \& Ermon(2020)Song and Ermon]{song2020improved}
Yang Song and Stefano Ermon.
\newblock Improved {T}echniques for {T}raining {S}core-{B}ased {G}enerative
  {M}odels.
\newblock \emph{Advances in Neural Information Processing Systems}, 33, 2020.

\bibitem[Song et~al.(2021{\natexlab{b}})Song, Durkan, Murray, and
  Ermon]{song2021maximum}
Yang Song, Conor Durkan, Iain Murray, and Stefano Ermon.
\newblock Maximum likelihood training of score-based diffusion models.
\newblock In \emph{Advances in Neural Information Processing Systems
  (NeurIPS)}, 2021{\natexlab{b}}.

\bibitem[Song et~al.(2021{\natexlab{c}})Song, Sohl-Dickstein, Kingma, Kumar,
  Ermon, and Poole]{song2021scorebased}
Yang Song, Jascha Sohl-Dickstein, Diederik~P Kingma, Abhishek Kumar, Stefano
  Ermon, and Ben Poole.
\newblock Score-based generative modeling through stochastic differential
  equations.
\newblock In \emph{International Conference on Learning Representations},
  2021{\natexlab{c}}.
\newblock URL \url{https://openreview.net/forum?id=PxTIG12RRHS}.

\bibitem[Song et~al.(2023)Song, Dhariwal, Chen, and
  Sutskever]{song2023consistency}
Yang Song, Prafulla Dhariwal, Mark Chen, and Ilya Sutskever.
\newblock Consistency models.
\newblock \emph{arXiv preprint arXiv:2303.01469}, 2023.

\bibitem[Thomee et~al.(2016)Thomee, Shamma, Friedland, Elizalde, Ni, Poland,
  Borth, and Li]{thomee2016yfcc100m}
Bart Thomee, David~A Shamma, Gerald Friedland, Benjamin Elizalde, Karl Ni,
  Douglas Poland, Damian Borth, and Li-Jia Li.
\newblock {YFCC100M}: The new data in multimedia research.
\newblock \emph{Communications of the ACM}, 59\penalty0 (2):\penalty0 64--73,
  2016.

\bibitem[Touvron et~al.(2023)Touvron, Martin, Stone, Albert, Almahairi, Babaei,
  Bashlykov, Batra, Bhargava, Bhosale, et~al.]{touvron2023llama}
Hugo Touvron, Louis Martin, Kevin Stone, Peter Albert, Amjad Almahairi, Yasmine
  Babaei, Nikolay Bashlykov, Soumya Batra, Prajjwal Bhargava, Shruti Bhosale,
  et~al.
\newblock Llama 2: Open foundation and fine-tuned chat models.
\newblock \emph{arXiv preprint arXiv:2307.09288}, 2023.

\bibitem[Tran et~al.(2017)Tran, Ranganath, and Blei]{tran2017hierarchical}
Dustin Tran, Rajesh Ranganath, and David Blei.
\newblock Hierarchical implicit models and likelihood-free variational
  inference.
\newblock In \emph{Advances in Neural Information Processing Systems}, pp.\
  5523--5533, 2017.

\bibitem[van~den Oord et~al.(2017)van~den Oord, Vinyals, and
  Kavukcuoglu]{van2017neural}
Aaron van~den Oord, Oriol Vinyals, and Koray Kavukcuoglu.
\newblock Neural discrete representation learning.
\newblock \emph{Advances in neural information processing systems}, 30, 2017.

\bibitem[Vaswani et~al.(2017)Vaswani, Shazeer, Parmar, Uszkoreit, Jones, Gomez,
  Kaiser, and Polosukhin]{vaswani2017attention}
Ashish Vaswani, Noam Shazeer, Niki Parmar, Jakob Uszkoreit, Llion Jones,
  Aidan~N Gomez, {\L}ukasz Kaiser, and Illia Polosukhin.
\newblock Attention is all you need.
\newblock \emph{Advances in neural information processing systems}, 30, 2017.

\bibitem[Vincent et~al.(2010)Vincent, Larochelle, Lajoie, Bengio, Manzagol, and
  Bottou]{vincent2010stacked}
Pascal Vincent, Hugo Larochelle, Isabelle Lajoie, Yoshua Bengio, Pierre-Antoine
  Manzagol, and L{\'e}on Bottou.
\newblock Stacked denoising autoencoders: Learning useful representations in a
  deep network with a local denoising criterion.
\newblock \emph{Journal of machine learning research}, 11\penalty0 (12), 2010.

\bibitem[Wang et~al.(2023{\natexlab{a}})Wang, Jiang, Lu, He, Chen, Wang, Zhou,
  et~al.]{wang2023context}
Zhendong Wang, Yifan Jiang, Yadong Lu, Pengcheng He, Weizhu Chen, Zhangyang
  Wang, Mingyuan Zhou, et~al.
\newblock In-context learning unlocked for diffusion models.
\newblock \emph{Advances in Neural Information Processing Systems},
  36:\penalty0 8542--8562, 2023{\natexlab{a}}.

\bibitem[Wang et~al.(2023{\natexlab{b}})Wang, Zheng, He, Chen, and
  Zhou]{wang2023diffusiongan}
Zhendong Wang, Huangjie Zheng, Pengcheng He, Weizhu Chen, and Mingyuan Zhou.
\newblock Diffusion-{GAN}: Training {GAN}s with diffusion.
\newblock In \emph{The Eleventh International Conference on Learning
  Representations}, 2023{\natexlab{b}}.
\newblock URL \url{https://openreview.net/forum?id=HZf7UbpWHuA}.

\bibitem[Wang et~al.(2023{\natexlab{c}})Wang, Lu, Wang, Bao, Li, Su, and
  Zhu]{wang2023prolificdreamer}
Zhengyi Wang, Cheng Lu, Yikai Wang, Fan Bao, Chongxuan Li, Hang Su, and Jun
  Zhu.
\newblock Prolific{D}reamer: High-fidelity and diverse text-to-{3D} generation
  with variational score distillation, 2023{\natexlab{c}}.

\bibitem[Wu et~al.(2023)Wu, Hao, Sun, Chen, Zhu, Zhao, and Li]{wu2023human}
Xiaoshi Wu, Yiming Hao, Keqiang Sun, Yixiong Chen, Feng Zhu, Rui Zhao, and
  Hongsheng Li.
\newblock {Human Preference Score v2: A Solid Benchmark for Evaluating Human
  Preferences of Text-to-Image Synthesis}.
\newblock \emph{arXiv preprint arXiv:2306.09341}, 2023.

\bibitem[Xiao et~al.(2022)Xiao, Kreis, and Vahdat]{xiao2021tackling}
Zhisheng Xiao, Karsten Kreis, and Arash Vahdat.
\newblock Tackling the generative learning trilemma with denoising diffusion
  {GAN}s.
\newblock In \emph{International Conference on Learning Representations}, 2022.
\newblock URL \url{https://openreview.net/forum?id=JprM0p-q0Co}.

\bibitem[Xing et~al.(2012)Xing, Zhou, Castrodad, Sapiro, and
  Carin]{xing2012dictionary}
Zhengming Xing, Mingyuan Zhou, Alexey Castrodad, Guillermo Sapiro, and Lawrence
  Carin.
\newblock Dictionary learning for noisy and incomplete hyperspectral images.
\newblock \emph{SIAM Journal on Imaging Sciences}, 5\penalty0 (1):\penalty0
  33--56, 2012.

\bibitem[Xu et~al.(2018)Xu, Zhang, Huang, Zhang, Gan, Huang, and
  He]{xu2018attngan}
Tao Xu, Pengchuan Zhang, Qiuyuan Huang, Han Zhang, Zhe Gan, Xiaolei Huang, and
  Xiaodong He.
\newblock {A}ttn{GAN}: Fine-grained text to image generation with attentional
  generative adversarial networks.
\newblock In \emph{CVPR}, pp.\  1316--1324, 2018.

\bibitem[Xu et~al.(2022)Xu, Wang, Zhang, Wang, and Shi]{xu2022versatile}
Xingqian Xu, Zhangyang Wang, Eric Zhang, Kai Wang, and Humphrey Shi.
\newblock Versatile diffusion: Text, images and variations all in one diffusion
  model.
\newblock \emph{arXiv preprint arXiv:2211.08332}, 2022.

\bibitem[Xu et~al.(2023)Xu, Zhao, Xiao, and Hou]{Xu2023UFOGenYF}
Yanwu Xu, Yang Zhao, Zhisheng Xiao, and Tingbo Hou.
\newblock {UFOGen}: You forward once large scale text-to-image generation via
  diffusion {GAN}s.
\newblock \emph{ArXiv}, abs/2311.09257, 2023.

\bibitem[Yang et~al.(2019)Yang, Martin, and Bondell]{yang2019variational}
Yue Yang, Ryan Martin, and Howard Bondell.
\newblock Variational approximations using {F}isher divergence.
\newblock \emph{arXiv preprint arXiv:1905.05284}, 2019.

\bibitem[Yin \& Zhou(2018)Yin and Zhou]{yin2018semi}
Mingzhang Yin and Mingyuan Zhou.
\newblock Semi-implicit variational inference.
\newblock In \emph{International Conference on Machine Learning}, pp.\
  5660--5669, 2018.

\bibitem[Yin et~al.(2023)Yin, Gharbi, Zhang, Shechtman, Durand, Freeman, and
  Park]{yin2023onestepDW}
Tianwei Yin, Michael Gharbi, Richard Zhang, Eli Shechtman, Fr{\'e}do Durand,
  William~T. Freeman, and Taesung Park.
\newblock One-step diffusion with distribution matching distillation.
\newblock \emph{ArXiv}, abs/2311.18828, 2023.

\bibitem[Yu et~al.(2022)Yu, Xu, Koh, Luong, Baid, Wang, Vasudevan, Ku, Yang,
  Ayan, et~al.]{yu2022scaling}
Jiahui Yu, Yuanzhong Xu, Jing~Yu Koh, Thang Luong, Gunjan Baid, Zirui Wang,
  Vijay Vasudevan, Alexander Ku, Yinfei Yang, Burcu~Karagol Ayan, et~al.
\newblock Scaling autoregressive models for content-rich text-to-image
  generation.
\newblock \emph{Transactions on Machine Learning Research}, 2022.

\bibitem[Yu \& Zhang(2023)Yu and Zhang]{yu2023semiimplicit}
Longlin Yu and Cheng Zhang.
\newblock Semi-implicit variational inference via score matching.
\newblock In \emph{The Eleventh International Conference on Learning
  Representations}, 2023.
\newblock URL \url{https://openreview.net/forum?id=sd90a2ytrt}.

\bibitem[Yu et~al.(2023)Yu, Xie, Zhu, Yang, Zhang, and
  Zhang]{yu2023hierarchical}
Longlin Yu, Tianyu Xie, Yu~Zhu, Tong Yang, Xiangyu Zhang, and Cheng Zhang.
\newblock Hierarchical semi-implicit variational inference with application to
  diffusion model acceleration.
\newblock In \emph{Thirty-seventh Conference on Neural Information Processing
  Systems}, 2023.
\newblock URL \url{https://openreview.net/forum?id=ghIBaprxsV}.

\bibitem[Zhang et~al.(2017)Zhang, Xu, Li, Zhang, Wang, Huang, and
  Metaxas]{zhang2017stackgan}
Han Zhang, Tao Xu, Hongsheng Li, Shaoting Zhang, Xiaogang Wang, Xiaolei Huang,
  and Dimitris~N Metaxas.
\newblock Stack{GAN}: Text to photo-realistic image synthesis with stacked
  generative adversarial networks.
\newblock In \emph{Proceedings of the IEEE international conference on computer
  vision}, pp.\  5907--5915, 2017.

\bibitem[Zhang et~al.(2021)Zhang, Yin, Fang, Li, Duan, Wu, Sun, Tian, Wu, and
  Wang]{zhang2021ernie}
Han Zhang, Weichong Yin, Yewei Fang, Lanxin Li, Boqiang Duan, Zhihua Wu,
  Yu~Sun, Hao Tian, Hua Wu, and Haifeng Wang.
\newblock {ERNIE-ViLG}: Unified generative pre-training for bidirectional
  vision-language generation.
\newblock \emph{arXiv preprint arXiv:2112.15283}, 2021.

\bibitem[Zhang et~al.(2020)Zhang, Chen, Tian, Wang, and
  Zhou]{Zhang2020Variational}
Hao Zhang, Bo~Chen, Long Tian, Zhengjue Wang, and Mingyuan Zhou.
\newblock Variational hetero-encoder randomized gans for joint image-text
  modeling.
\newblock In \emph{International Conference on Learning Representations}, 2020.
\newblock URL \url{https://openreview.net/forum?id=H1x5wRVtvS}.

\bibitem[Zhang \& Chen(2023)Zhang and Chen]{zhang2023fast}
Qinsheng Zhang and Yongxin Chen.
\newblock Fast sampling of diffusion models with exponential integrator.
\newblock In \emph{The Eleventh International Conference on Learning
  Representations}, 2023.
\newblock URL \url{https://openreview.net/forum?id=Loek7hfb46P}.

\bibitem[Zhao et~al.(2023)Zhao, Bai, Rao, Zhou, and Lu]{zhao2023unipc}
Wenliang Zhao, Lujia Bai, Yongming Rao, Jie Zhou, and Jiwen Lu.
\newblock Unipc: A unified predictor-corrector framework for fast sampling of
  diffusion models.
\newblock \emph{arXiv preprint arXiv:2302.04867}, 2023.

\bibitem[Zheng et~al.(2023{\natexlab{a}})Zheng, Nie, Vahdat, Azizzadenesheli,
  and Anandkumar]{zheng2023fast}
Hongkai Zheng, Weili Nie, Arash Vahdat, Kamyar Azizzadenesheli, and Anima
  Anandkumar.
\newblock Fast sampling of diffusion models via operator learning.
\newblock In \emph{International Conference on Machine Learning}, pp.\
  42390--42402. PMLR, 2023{\natexlab{a}}.

\bibitem[Zheng et~al.(2023{\natexlab{b}})Zheng, He, Chen, and
  Zhou]{zheng2023truncated}
Huangjie Zheng, Pengcheng He, Weizhu Chen, and Mingyuan Zhou.
\newblock Truncated diffusion probabilistic models and diffusion-based
  adversarial auto-encoders.
\newblock In \emph{The Eleventh International Conference on Learning
  Representations}, 2023{\natexlab{b}}.
\newblock URL \url{https://openreview.net/forum?id=HDxgaKk956l}.

\bibitem[Zheng et~al.(2024)Zheng, Wang, Yuan, Ning, He, You, Yang, and
  Zhou]{zheng2024learning}
Huangjie Zheng, Zhendong Wang, Jianbo Yuan, Guanghan Ning, Pengcheng He,
  Quanzeng You, Hongxia Yang, and Mingyuan Zhou.
\newblock Learning stackable and skippable {LEGO} bricks for efficient,
  reconfigurable, and variable-resolution diffusion modeling.
\newblock In \emph{The Twelfth International Conference on Learning
  Representations}, 2024.
\newblock URL \url{https://openreview.net/forum?id=qmXedvwrT1}.

\bibitem[Zhou et~al.(2009)Zhou, Chen, Paisley, Ren, Sapiro, and
  Carin]{zhou2009non}
Mingyuan Zhou, Haojun Chen, John Paisley, Lu~Ren, Guillermo Sapiro, and
  Lawrence Carin.
\newblock Non-parametric {B}ayesian dictionary learning for sparse image
  representations.
\newblock \emph{Advances in neural information processing systems}, 22, 2009.

\bibitem[Zhou et~al.(2023)Zhou, Chen, Wang, and Zheng]{zhou2023beta}
Mingyuan Zhou, Tianqi Chen, Zhendong Wang, and Huangjie Zheng.
\newblock Beta diffusion.
\newblock In \emph{Neural Information Processing Systems}, 2023.
\newblock URL \url{https://arxiv.org/abs/2309.07867}.

\bibitem[Zhou et~al.(2024)Zhou, Zheng, Wang, Yin, and Huang]{zhou2024score}
Mingyuan Zhou, Huangjie Zheng, Zhendong Wang, Mingzhang Yin, and Hai Huang.
\newblock Score identity distillation: Exponentially fast distillation of
  pretrained diffusion models for one-step generation.
\newblock In \emph{Forty-first International Conference on Machine Learning},
  2024.
\newblock URL \url{https://openreview.net/forum?id=QhqQJqe0Wq}.

\bibitem[Zhou et~al.(2025)Zhou, Zheng, Gu, Wang, and
  Huang]{zhou2025adversarial}
Mingyuan Zhou, Huangjie Zheng, Yi~Gu, Zhendong Wang, and Hai Huang.
\newblock Adversarial score identity distillation: Rapidly surpassing the
  teacher in one step.
\newblock In \emph{International Conference on Learning Representations}, 2025.

\bibitem[Zhou et~al.(2022)Zhou, Zhang, Chen, Li, Tensmeyer, Yu, Gu, Xu, and
  Sun]{zhou2022towards}
Yufan Zhou, Ruiyi Zhang, Changyou Chen, Chunyuan Li, Chris Tensmeyer, Tong Yu,
  Jiuxiang Gu, Jinhui Xu, and Tong Sun.
\newblock Towards language-free training for text-to-image generation.
\newblock In \emph{Proceedings of the IEEE/CVF Conference on Computer Vision
  and Pattern Recognition}, pp.\  17907--17917, 2022.

\end{thebibliography}
\normalsize

\newpage
\appendix
\begin{center}
    \Large\textbf{%
    Guided Score Identity Distillation for Data-Free One-Step Text-to-Image Generation:
    Appendix}
\end{center}

\begin{figure}[!ht]
    \centering
    SD1.5 distilled with SiD-LSG ($\kappa=1.5$): FID $=$ \textbf{8.15}, CLIP $=$ 0.304 \begin{minipage}[b]{0.193\textwidth}
        \centering
        \includegraphics[width=\textwidth]{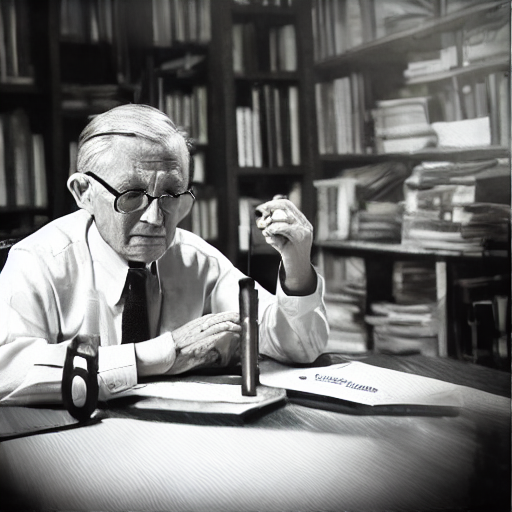}
     \\
    \end{minipage}~
    \begin{minipage}[b]{0.193\textwidth}
        \centering
        \includegraphics[width=\textwidth]{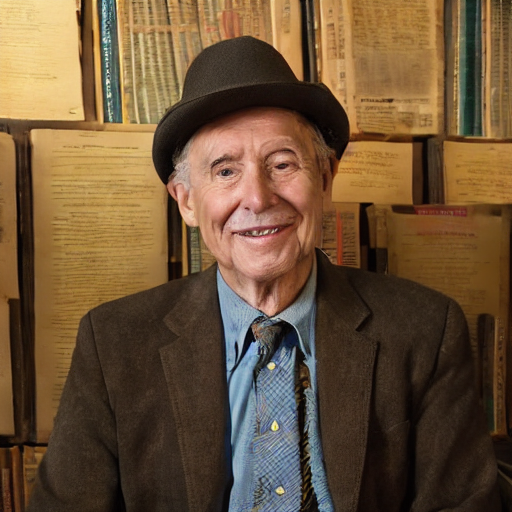}
     \\
    \end{minipage}~
    \begin{minipage}[b]{0.193\textwidth}
        \centering
        \includegraphics[width=\textwidth]{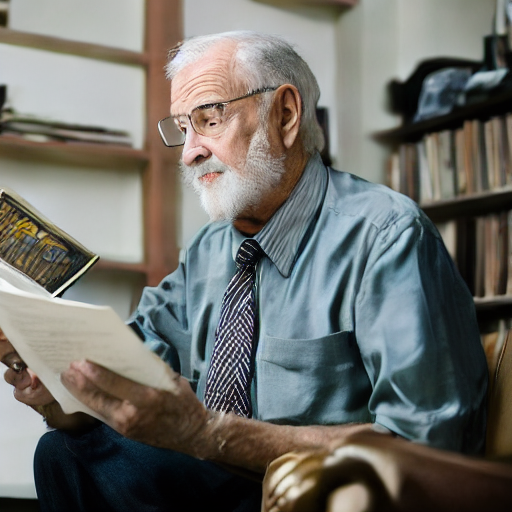}
     \\
    \end{minipage}~
    \begin{minipage}[b]{0.195\textwidth}
        \centering
        \includegraphics[width=\textwidth]{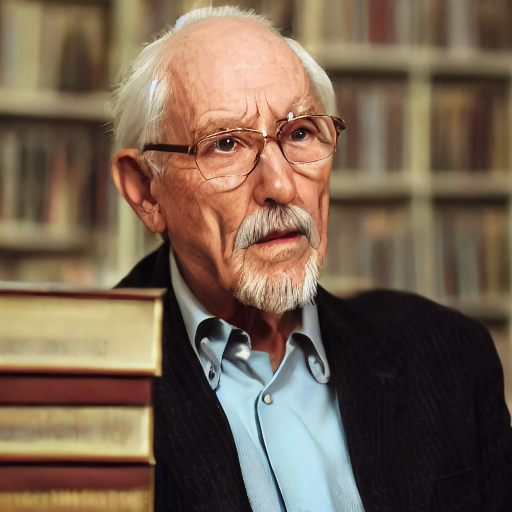}
     \\
    \end{minipage}~
    \begin{minipage}[b]{0.193\textwidth}
        \centering
        \includegraphics[width=\textwidth]{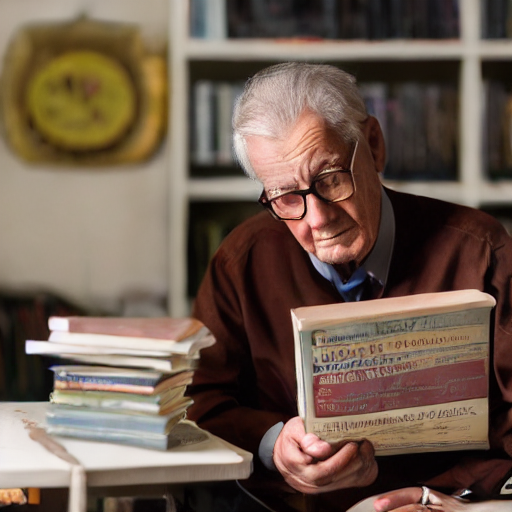}
     \\
    \end{minipage}
    \\
    SD2.1-base distilled with SiD-LSG ($\kappa=4.5$): FID = 16.54, CLIP = \textbf{0.322} \begin{minipage}[b]{0.193\textwidth}
        \centering
        \includegraphics[width=\textwidth]{sid_sd_images/figure1/000000.png}
     \\
    \end{minipage}~
    \begin{minipage}[b]{0.193\textwidth}
        \centering
        \includegraphics[width=\textwidth]{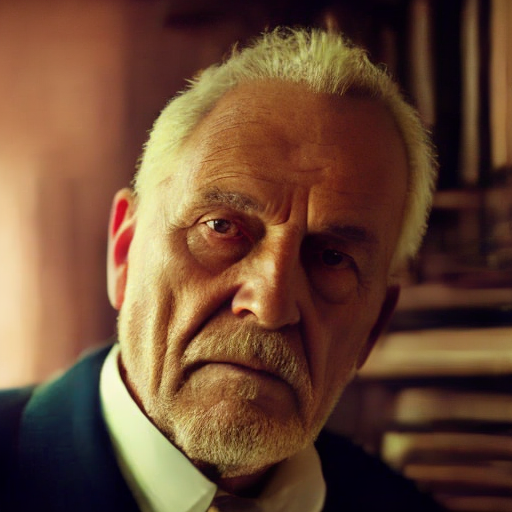}
     \\
    \end{minipage}~
    \begin{minipage}[b]{0.193\textwidth}
        \centering
        \includegraphics[width=\textwidth]{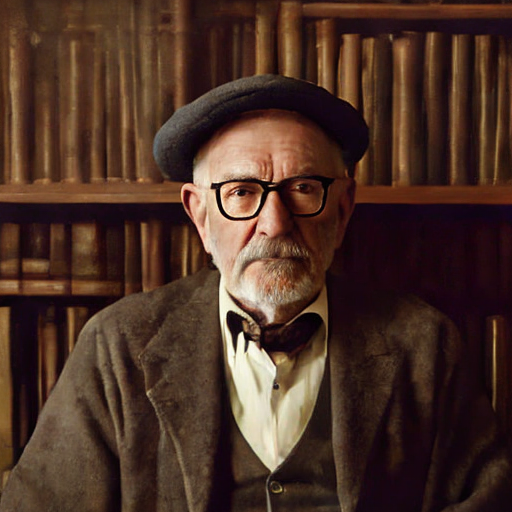}
     \\
    \end{minipage}~
    \begin{minipage}[b]{0.193\textwidth}
        \centering
        \includegraphics[width=\textwidth]{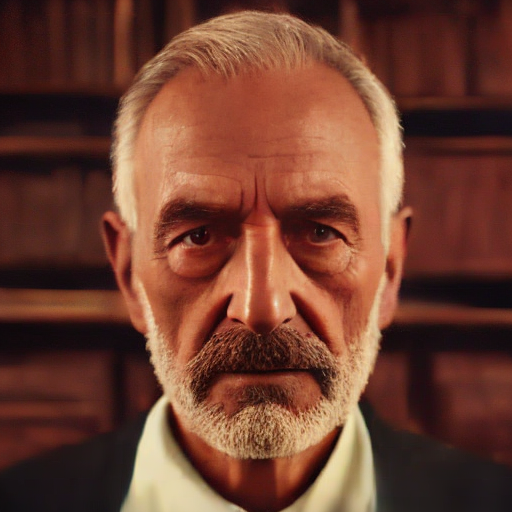}
     \\
    \end{minipage}~
    \begin{minipage}[b]{0.195\textwidth}
        \centering
        \includegraphics[width=\textwidth]{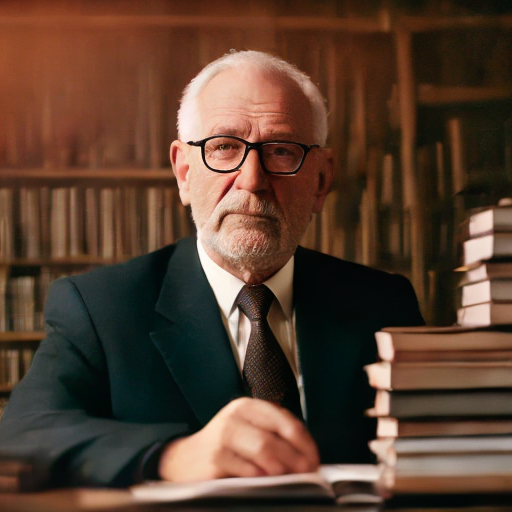}
     \\
    \end{minipage}
    \\
    \caption{\small Visual comparison of two SiD-LSG models: one preferred for FID and the other for CLIP. All images are generated from the same text prompt: ``A distinguished older gentleman in a vintage study, surrounded by books and dim lighting, his face marked by wisdom and time. 8K, hyper-realistic, cinematic, post-production." The model with a lower guidance scale of $\kappa=1.5$, which achieves a record-low one-step-generation FID of 8.15 and a competitive CLIP score of 0.304, produces images that are more diverse but align less closely with specific text details, such as ``dim lighting.'' Conversely, the model with a higher guidance scale of $\kappa=4.5$, achieving a high CLIP score of 0.322 and noted for state-of-the-art human preference scores (HPSv2) as shown in Table 2, presents a relatively high FID of 16.54, indicating less diversity but superior text alignment and visual quality.
\normalsize}
    \label{fig:qualitative_1}
\vspace{5mm}

    \centering
    SD1.5 distilled with SiD-LSG ($\kappa=1.5$): FID $=$ \textbf{8.15}, CLIP $=$ 0.304 \begin{minipage}[b]{0.193\textwidth}
        \centering
        \includegraphics[width=\textwidth]{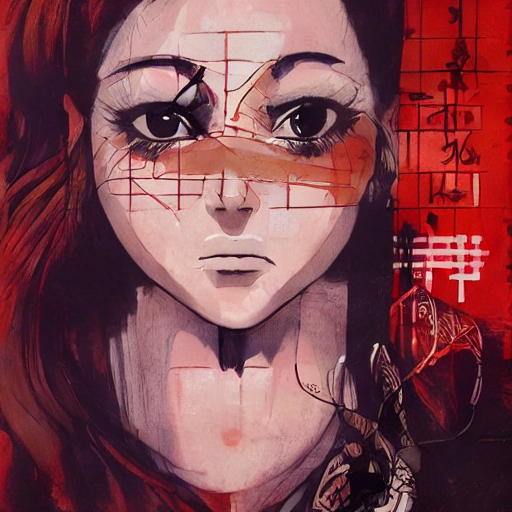}
     \\
    \end{minipage}~
    \begin{minipage}[b]{0.193\textwidth}
        \centering
        \includegraphics[width=\textwidth]{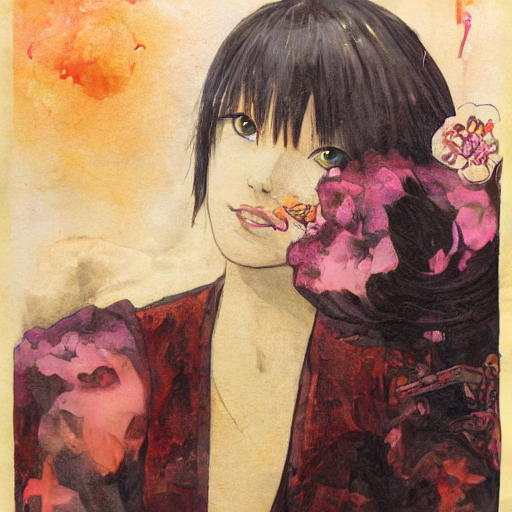}
     \\
    \end{minipage}~
    \begin{minipage}[b]{0.193\textwidth}
        \centering
        \includegraphics[width=\textwidth]{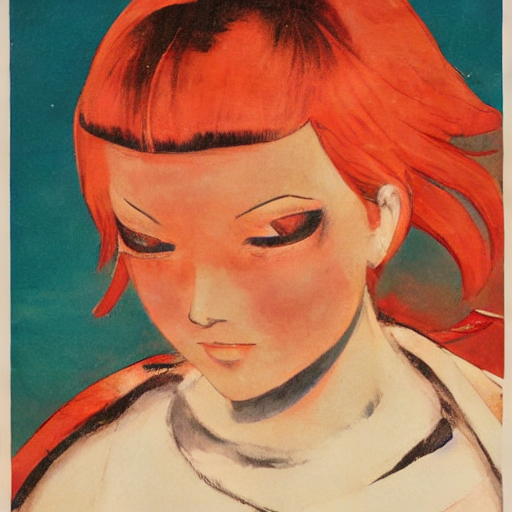}
     \\
    \end{minipage}~
    \begin{minipage}[b]{0.195\textwidth}
        \centering
        \includegraphics[width=\textwidth]{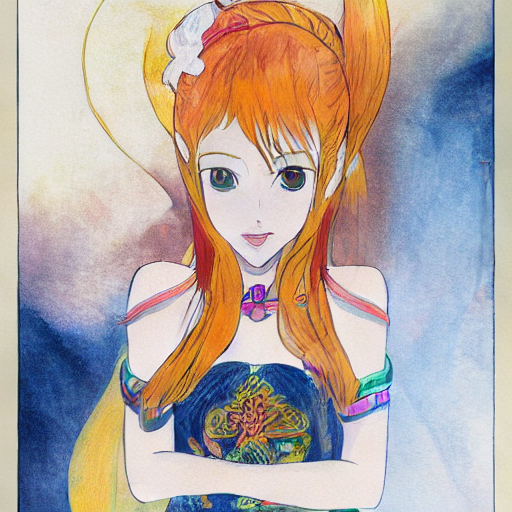}
     \\
    \end{minipage}~
    \begin{minipage}[b]{0.193\textwidth}
        \centering
        \includegraphics[width=\textwidth]{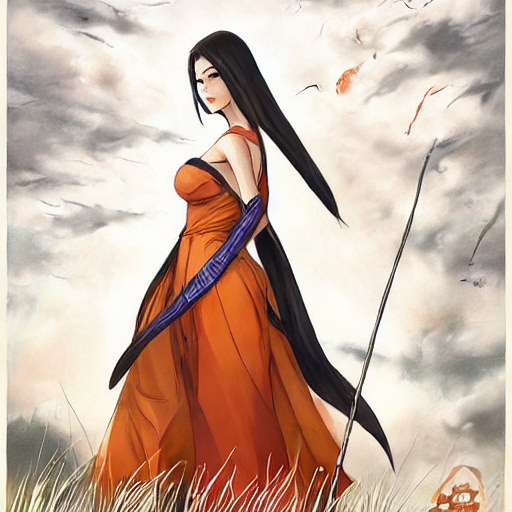}
     \\
    \end{minipage}
    \\
    SD2.1-base distilled with SiD-LSG ($\kappa=4.5$): FID = 16.54, CLIP = \textbf{0.322} \begin{minipage}[b]{0.193\textwidth}
        \centering
        \includegraphics[width=\textwidth]{sid_sd_images/figure1/000004.png}
     \\
    \end{minipage}~
    \begin{minipage}[b]{0.193\textwidth}
        \centering
        \includegraphics[width=\textwidth]{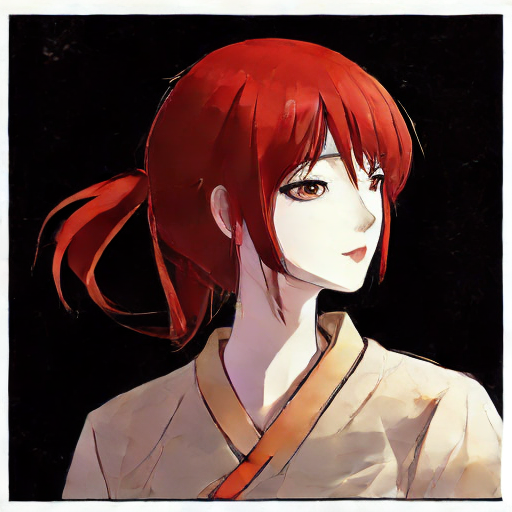}
     \\
    \end{minipage}~
    \begin{minipage}[b]{0.193\textwidth}
        \centering
        \includegraphics[width=\textwidth]{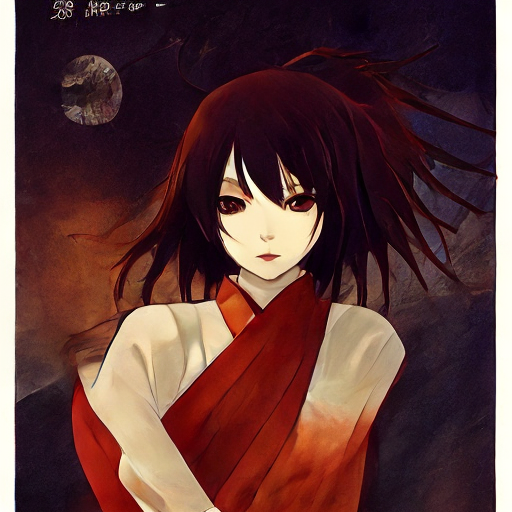}
     \\
    \end{minipage}~
    \begin{minipage}[b]{0.193\textwidth}
        \centering
        \includegraphics[width=\textwidth]{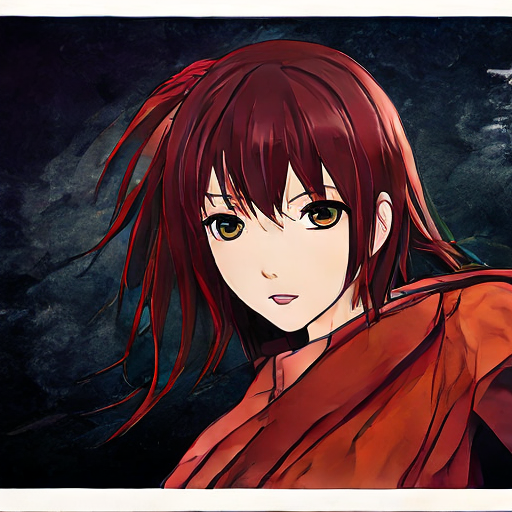}
     \\
    \end{minipage}~
    \begin{minipage}[b]{0.195\textwidth}
        \centering
        \includegraphics[width=\textwidth]{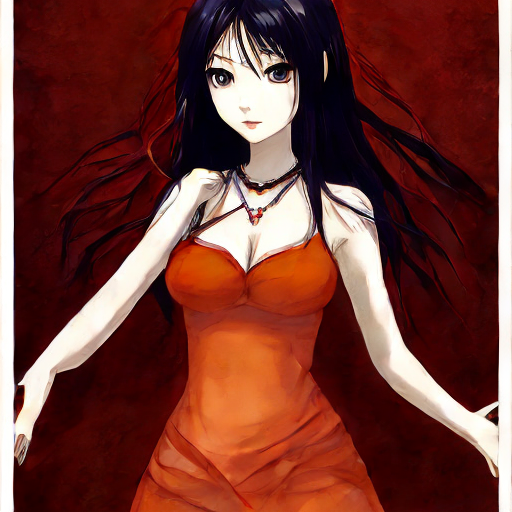}
     \\
    \end{minipage}
    \\
    \caption{\small
Analogous to Figure \ref{fig:qualitative_1}, this plot compares two rows of images generated using two distinct guidance scales, both conditioning on the same text prompt: ``poster art for the collection of the asian woman, in the style of gloomy, dark orange and white, dynamic anime, realistic watercolors, nintencore, weathercore, mysterious realism --ar 69:128 --s 750 --niji 5".
\normalsize}
    \label{fig:qualitative_2}
\end{figure}

\section{Broader impact}\label{sec:broaderimpact}
The broader impact of our work is multifaceted. On one hand, it significantly reduces the energy required to operate state-of-the-art diffusion models, contributing to more sustainable AI practices. On the other hand, the ease of distilling and deploying models that might be trained on data with questionable content or intentions presents ethical challenges. Therefore, it is crucial for the community to engage in discussions on how to minimize risks while enhancing the benefits of such advancements. This involves developing robust guidelines and frameworks to govern the use and deployment of distilled models, ensuring they are used responsibly and ethically.

\section{Related work}\label{sec:relatedwork}

Generative modeling of high-dimensional data has long been a focal point in machine learning research. This area primarily concentrates on replicating various data distributions: the original data distribution, conditional distributions influenced by labels, noisy or incomplete measurements, textual descriptions, or the joint distribution of data and other modalities. This has spurred the development of a diverse range of generative models and methodologies. Initially, these models were only capable of handling simpler, low-dimensional data such as \(28 \times 28\) grayscale or binarized MNIST digits \citep{hinton2006fast,salakhutdinov2009deep,vincent2010stacked}, vector-quantized local descriptors \citep{fei2005bayesian,chong2009simultaneous}, or patches of natural and hyperspectral images \citep{zhou2009non,xing2012dictionary,polatkan2015bayesian}. Early models often utilized neural networks with stochastic binary hidden layers or shallow hierarchical Bayesian  models, which are simpler to train but have limited capacities.

\textbf{Deep generative models. }
To tackle the generation of high-dimensional data, such as images comprising millions of pixels, substantial advancements in generative models have been made over the past~decade. 
This period has marked the emergence of diverse deep generative models, including variational auto-encoders (VAEs) \citep{kingma2013auto,rezende2014stochastic}, normalizing flows \citep{papamakarios2019normalizing}, generative adversarial networks (GANs) \citep{goodfellow2014generative,reed2016generative,karras2019style,wang2023diffusiongan}, autoregressive models \citep{gregor2015draw,mansimov2015generating},
and diffusion models \citep{sohl2015deep,song2019generative,ho2020denoising,song2020improved,song2021maximum,song2021denoising,dhariwal2021diffusion,karras2022elucidating,peebles2023scalable,zheng2024learning}. Additionally, essential resources for creating effective T2I synthesis systems, such as large language models  \citep{vaswani2017attention,devlin2018bert,radford2018improving,radford2019language,raffel2020exploring,he2020deberta,brown2020language,achiam2023gpt,touvron2023llama,jiang2023mistral}, large vision-language models \citep{radford2021learning, ilharco_gabriel_2021_5143773,cherti2023reproducible}, advanced visual tokenization and compression \citep{rolfe2016discrete,van2017neural,esser2021taming,rombach2022high,podell2024sdxl}, and extensive training datasets 
\citep{thomee2016yfcc100m,changpinyo2021cc12m,schuhmann2022laion},  have transitioned from proprietary tools of major tech companies to publicly available assets.

Previously, T2I models primarily utilized GANs and focused on small-scale, object-centric domains like flowers, birds, and face images~\citep{reed2016generative,zhang2017stackgan,xu2018attngan,Zhang2020Variational}.
Powered by text encoders that leverage pretrained large language models or vision-language models, which offer profound language comprehension and extract semantically rich latent representations, and supported by an extensive collection of text-image pairs, three main families of generative models---GANs \citep{zhou2022towards,sauer2022stylegan,kang2023scaling}, autoregressive models \citep{ramesh2021zero,zhang2021ernie,ding2021cogview,gafni2022make,yu2022scaling,chang2023muse}, and diffusion models \citep{nichol2022glide,
ramesh2022hierarchical,saharia2022photorealistic,rombach2022high,xu2022versatile,wang2023context,qin2023unicontrol}---have effectively capitalized on these technological advancements. These developments have facilitated the creation of T2I synthesis systems that demonstrate exceptional photorealism and sophisticated language understanding.

\textbf{Acceleration of diffusion-based generation. }
Early-stage diffusion models were notably slow in sampling, spurring extensive research aimed at speeding up the reverse diffusion process. Researchers have approached this by interpreting diffusion models through stochastic or ordinary differential equations and using advanced numerical solvers to enhance efficiency \citep{song2021scorebased,song2021denoising,liu2022pseudo,lu2022dpmsolver,zhang2023fast,karras2022elucidating}. Additional strategies include truncating the diffusion chain to initiate generation from more structured distributions \citep{pandey2022diffusevae,zheng2023truncated,lyu2022accelerating}, integrating these models with GANs to boost generation speed \citep{xiao2021tackling, wang2023diffusiongan}, and exploring flow matching in diffusion modeling \citep{liu2022flow,lipman2022flow,albergo2023stochastic}. 
More recently, research has pivoted towards distilling reverse diffusion chains to refine and expedite the generation process, a direction that continues to evolve with new methodologies and insights 
\citep{luhman2021knowledge,salimans2022progressive,zheng2023fast,meng2023distillation,song2023consistency,song2023improved,kim2023consistency,Sauer2023AdversarialDD,Xu2023UFOGenYF,yin2023onestepDW,luo2023diffinstruct,zhou2024score}.

\textbf{T2I diffusion distillation. } Recent efforts aim to accelerate the sampling process from pre-trained diffusion teachers like SD~\citep{rombach2022high}. \citet{Sauer2023AdversarialDD} focused on distilling diffusion models into generators capable of one or two-step operations through adversarial training. \citet{Xu2023UFOGenYF} introduced UFOGen, utilizing a time-step-dependent discriminator for generator initialization. \citet{Luo2023LatentCM} applied consistency distillation \citep{song2023consistency} to text-guided latent diffusion models \citep{ramesh2022hierarchical} for efficient, high-fidelity T2I generation. Building on the idea of using a pre-trained 2D T2I diffusion model for text-to-3D synthesis~\citep{Poole2022DreamFusionTU,wang2023prolificdreamer}, %
SwiftBrush \citep{thuan2024swiftbrush} showcases its effectiveness of distilling pre-trained stable diffusion models. Distribution Matching Distillation (DMD) by \citet{yin2023onestepDW} further enhances distillation quality by adding a regression loss term. 

\textbf{How does SiD-LSG differ from previous methods?}
SiD-LSG introduces several unique features that distinguish it from previous T2I diffusion distillation methods. Firstly, similar to SwiftBrush \citep{thuan2024swiftbrush}, SiD-LSG is a data-free distillation method, which eliminates the need for original training datasets or synthetic data generated with SDE/ODE solvers during the distillation process. However, unlike SwiftBrush, SiD-LSG necessitates gradient backpropagation through the score networks, a critical step akin to gradient backpropagation through the discriminator in methods employing adversarial losses. 
Additionally, SiD-LSG applies CFG to both the training and evaluation of the fake score network. This is a departure from previous methods, which typically apply CFG only during the evaluation of the pretrained score network. Finally, SiD-LSG aims to minimize a model-based explicit score-matching loss, a type of Fisher divergence. In contrast, previous methods often focus on minimizing losses based on KL divergence, consistency, regression, GAN-based adversarial tactics, or a combination thereof.

\section{Further Discussion on Long, Short, and Long-Short Guidance}\label{sec:discuss-LSG}

Regarding the relationship between the choice of $\kappa_2=\kappa_3$ and the categorization into long, short, and long-short guidance, we clarify that various combinations of $\kappa_{1,2,3,4}$ can define these strategies:

\textbf{Long Strategy: } This strategy involves enhancing the teacher’s CFG more than the fake score network during inference or making them equal but both larger than one. It is typically represented by $\kappa_1=\kappa_2=\kappa_3=1, \kappa_4>1$, but also includes configurations where $\kappa_1=\kappa_2=\kappa_3=\kappa_4>1$.

\textbf{Short Strategy: } This approach aims to reduce the fake-score-network’s CFG during evaluation compared to training or maintain them equal but greater than one. This strategy is exemplified by $\kappa_1=\kappa_4=1, 0<\kappa_2=\kappa_3<1$, and can also be part of $\kappa_1=\kappa_2=\kappa_3=\kappa_4>1$.

\textbf{Long-Short Strategy: } The ``simplest'' LSG configuration $\kappa_1>1, \kappa_2=\kappa_3=\kappa_4=1$ can be interpreted as incorporating both long and short strategies since increasing $\kappa_1$ for training the fake score network effectively implies that during inference, the fake score network is guided by a weaker CFG than both the teacher and its own training setting.

SiD-LSG employs a default setting where $\kappa_1= \kappa_2=\kappa_3=\kappa_4=\kappa>1$. This implies that while the teacher’s training CFG is 1 and evaluation CFG is $\kappa>1$, the fake score network's training and evaluation CFG are both $\kappa>1$, making the teacher’s CFG during inference stronger than that of the fake score network.

\begin{figure}[ht]
\centering
\includegraphics[width=0.7\textwidth]{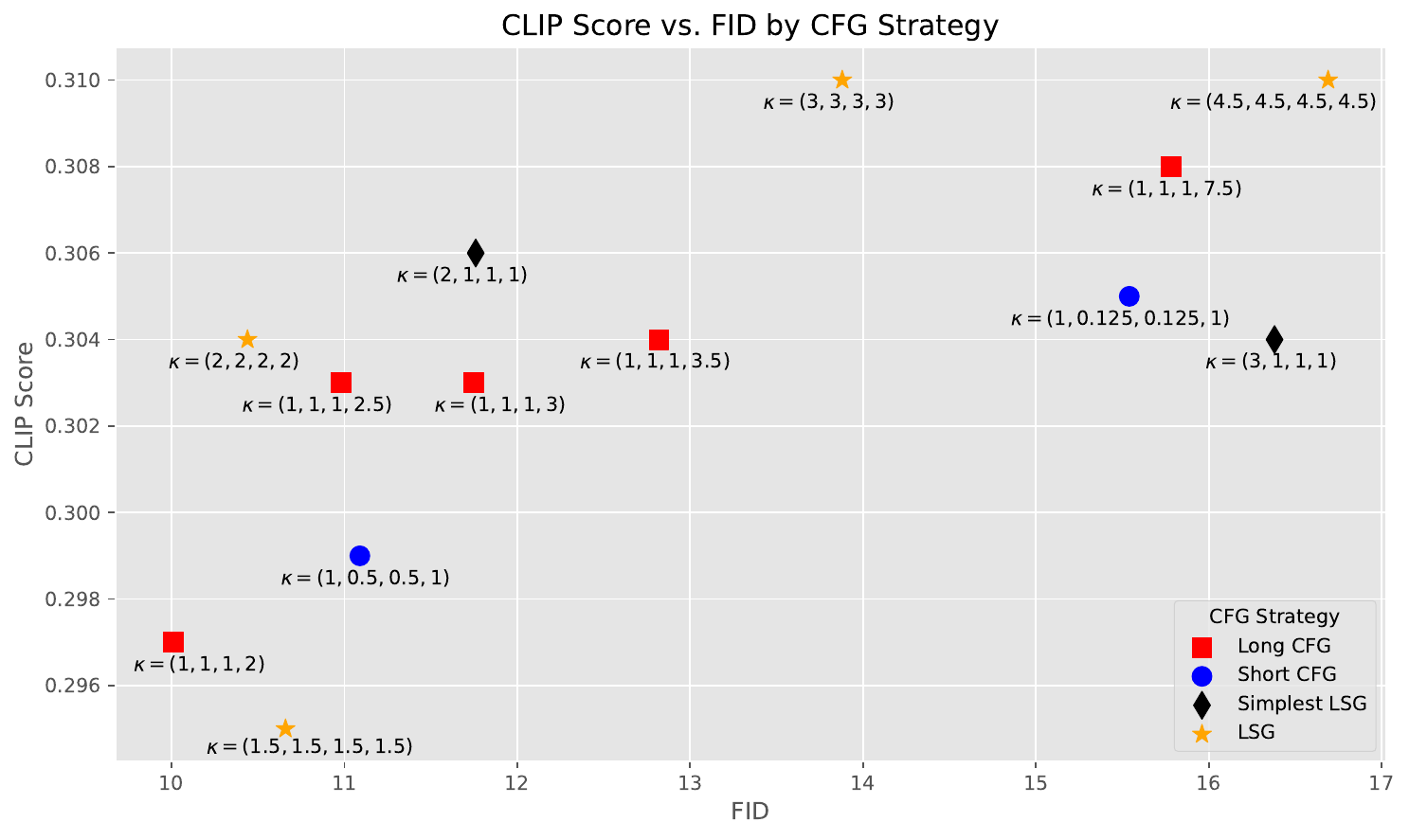} \vspace{-3mm}
\caption{  \small Comparison of different CFG strategies in terms of the best FID achieved when iterating with fewer than 2.56M images to distill the generators on Stable Diffusion 1.5 and their corresponding CLIP scores. The markers represent different CFG strategies: Long CFG is denoted by red squares, Short CFG by blue circles, the ``simplest'' LSG by black diamonds, and LSG by orange stars.}
\label{fig:fid_clip}
\end{figure}

\textbf{FID-CLIP Compromise:} We evaluate various CFG strategies based on the best FID achieved with fewer than 2.56M fake images used to distill generators on Stable Diffusion 1.5, alongside their corresponding CLIP scores. The results, whose trajectories are depicted in Figures \ref{fig:cfg_1114} and \ref{fig:lsg} and detailed in Table~\ref{tab:fid_clip} and Figure~\ref{fig:fid_clip}, indicate that the proposed LSG strategy achieves the best balance between lowering FID and enhancing CLIP scores.

\begin{table}[t]
\centering
\caption{\small  Comparison of different CFG strategies in terms of the best FID achieved when iterating with fewer than 2.56M fake images to distill the generators on Stable Diffusion 1.5 and their corresponding CLIP scores.}\vspace{-3mm}
\label{tab:fid_clip}
 \footnotesize
\begin{tabular}{@{}lcccccc@{}}
\toprule
CFG Strategy& $\kappa_1$ & $\kappa_2$ & $\kappa_3$ & $\kappa_4$ & FID & CLIP \\ \midrule
\multirow{5}{*}{Long CFG} & 1 & 1 & 1 & 2 & 10.01 & 0.297 \\
& 1 & 1 & 1 & 2.5 & 10.98 & 0.303 \\
& 1 & 1 & 1 & 3 & 11.75 & 0.303 \\
& 1 & 1 & 1 & 3.5 & 12.82 & 0.304 \\
& 1 & 1 & 1 & 7.5 & 15.78 & 0.308 \\
\midrule
\multirow{1}{*}{No CFG} & 1 & 1 & 1 & 1 & 15.49 & 0.269 \\
\midrule
\multirow{2}{*}{Short CFG} & 1 & 0.5 & 0.5 & 1 & 11.09 & 0.299 \\
& 1 & 0.125 & 0.125 & 1 & 15.54 & 0.305 \\
\midrule
\multirow{2}{*}{``Simplest'' LSG} & 2 & 1 & 1 & 1 & 11.76 & 0.306 \\
& 3 & 1 & 1 & 1 & 16.38 & 0.304 \\
\midrule
\multirow{4}{*}{LSG} & 1.5 & 1.5 & 1.5 & 1.5 & 10.66 & 0.295 \\
& 2 & 2 & 2 & 2 & 10.44 & 0.304 \\
& 3 & 3 & 3 & 3 & 13.88 & 0.310 \\
& 4.5 & 4.5 & 4.5 & 4.5 & 16.69 & 0.310 \\ \bottomrule
\end{tabular}
\end{table}

\section{Training and evaluation details} \label{sec:detail}

The hyperparameters tailored for our study are outlined in Table \ref{tab:Hyperparameters}. It's important to note that the time and memory costs reported in Table \ref{tab:Hyperparameters} do not account for those incurred during periodic evaluations of the FID and CLIP scores of the single-step generator, nor do they include the resources used to save model checkpoints during the distillation process. These costs can vary significantly depending on the computing resources used, including the versions of CUDA and Flash Attention, as well as the storage platforms employed and the frequency of operations.

\begin{algorithm}[t]
\scriptsize
\caption{\small SiD-LSG: Score identity Distillation with Long-Short classifier-free Guidance }\label{alg:sid}
\begin{algorithmic}
\STATE \textbf{Input:} Pretrained score network $f_\phi$, generator $G_\theta$, fake score network $f_\psi$, $t_{\text{init}}=625$, $t_{\min}=20$, $t_{\max}=979$, %
guidance scales $\kappa_1=\kappa_2=\kappa_3=\kappa_4=1.5$
\STATE \textbf{Initialization} $\theta \leftarrow \phi, \psi \leftarrow \phi$ 
\REPEAT

\STATE Sample $\zv \sim \cN(0,  \mathbf{I})$ and $\cv$, replacing $\cv$ with the embedding of an empty text string 10\% of the time; Define $\xv_g= G_\theta(\zv,\cv)=f_\theta(\sigma_{t_\text{init}}\zv,t_{\text{init}},\cv)$; Sample $t\in \{t_{min},\ldots,t_{\max}\}$ and $\epsilonv_t\sim \mathcal{N}(0,\mathbf{I})$, and let $\xv_t = a_t \xv_g + \sigma_t \epsilonv_t$
\STATE  Update $\psi$ with $\psi = \psi - \eta \nabla_\psi \hat{\cL}_{\psi}$, where
$$\hat{\cL}_{\psi} =\textstyle\frac{a^2_t}{\sigma^2_t} %
\|f_{\psi,\kappa_1}(\xv_t,t,\cv)-\xv_g\|_2^2 =  \|\epsilonv_{\psi,\kappa_1}(\xv_t, t,\cv)-\epsilonv_t\|_2^2$$

\STATE Sample $\zv \sim \cN(0,  \mathbf{I})$ and $\cv$, and let $\xv_g= G_\theta(\zv,\cv)=f_\theta(\sigma_{t_\text{init}}\zv,t_{\text{init}},\cv)$; Sample $t\in \{t_{min},\ldots,t_{\max}\}$ and $\epsilonv_t\sim \mathcal{N}(0,\mathbf{I})$, compute $\omega_t$ with~\eqref{eq:weight}, and let $\xv_t = a_t \xv_g + \sigma_t \epsilonv_t$
\STATE  Update $G_\theta$ with $\theta = \theta - \eta \nabla_\theta \tilde{\cL}_{\theta}$, where
\ba{\tilde{\cL}_{\theta} &=
\textstyle\frac{\omega(t)a_t^2}{{\sigma_t^4} } (f_{\phi,\kappa_4}(\xv_t, t,\cv)-f_{\psi,\kappa_2}(\xv_t, t,\cv))^T(f_{\psi,\kappa_3}(\xv_t, t,\cv)-\xv_g)\notag\\
&=\textstyle\frac{\omega(t)}{{\sigma_t^2} } (\epsilonv_{\psi,\kappa_2}(\xv_t, t,\cv)-\epsilonv_{\phi,\kappa_4}(\xv_t, t,\cv))^T(\epsilonv_t-\epsilonv_{\psi,\kappa_3}(\xv_t, t,\cv))\notag
}
\UNTIL{%
processing 10M fake images or the training budget is exhausted}
\STATE \textbf{Output:} $G_\theta$
\normalsize
\end{algorithmic}
\end{algorithm} %

\begin{table}[h]
\caption{\small Hyperparameter settings and comparison of distillation time and memory usage between different long and short guidance strategies of SiD. Note in order to run the long-short guidance (LSG) with $\kappa_1=\kappa_2=\kappa_3=\kappa_4>1$, we need to turn off the EMA network for some cases (indicated with ``no EMA''), otherwise it will be out of memory.}
\label{tab:Hyperparameters}
\begin{center}
\resizebox{\textwidth}{!}{
\begin{tabular}{cccccc}
\toprule
Computing platform & Hyperparameters & Long Strategy & Short Strategy & Long-Short & Long-Short \\
\midrule
\multirow{9}{*}{General Settings}                                 & \multirow{2}{*}{CFG scales} & $\kappa_4>1$                   & $0<\kappa_2=\kappa_3<1$ & $\kappa_1>1$ &  $\kappa_{1,2,3,4}>1$\\
                                  &                             & $\kappa_1=\kappa_2=\kappa_3=1$ & $\kappa_1=\kappa_4=1$   & $\kappa_2=\kappa_3=\kappa_4=1$ &  $\kappa_1=\kappa_2=\kappa_3=\kappa_4$\\
& Batch size & \multicolumn{4}{c}{512} \\
                                  & Learning rate & \multicolumn{4}{c}{1e-6} \\
                                  & Half-life of EMA & \multicolumn{4}{c}{50k images} \\
                                  & Optimizer under FP32 & \multicolumn{4}{c}{Adam ($\beta_1=0$, $\beta_2=0.999$, $\epsilon=1\text{e-8}$)} \\
                                  & Optimizer under FP16 & \multicolumn{4}{c}{Adam ($\beta_1=0$, $\beta_2=0.999$, $\epsilon=1\text{e-6}$)} \\
                                  & $\alpha$ & \multicolumn{4}{c}{1.0} \\
                                  & Time parameters & \multicolumn{4}{c}{$(t_{\min}, t_{\text{init}}, t_{\max}) = (20, 625, 979)$} \\
\midrule
\multirow{8}{*}{RTX-A5000 (24G), FP16} & xFormers  available and enabled & \multicolumn{4}{c}{Yes} \\
                                        & Batch size per GPU & \multicolumn{4}{c}{1} \\
                                        & \# of GPUs & \multicolumn{4}{c}{8} \\
                                        & \# of gradient accumulation round & \multicolumn{4}{c}{64} \\
                                        & Max memory in GB allocated & 22.9 & 22.9 & 22.6 & 22.0  (no EMA) \\
                                        & Max memory in GB reserved & 23.0 & 23.0 & 23.0 & 22.1 (no EMA) \\
                                        & Time in seconds per 1k images & 74 & 75 & 80 & 102 \\
                                        & Time in hours per 1M images & 21 & 21 & 22 & 29 \\
\midrule
\multirow{8}{*}{RTX-A6000 (48G), FP32} & xFormers  available and enabled & \multicolumn{4}{c}{Yes} \\
                                & Batch size per GPU & \multicolumn{4}{c}{1} \\
                                & \# of GPUs  & \multicolumn{4}{c}{8} \\
                                & \# of gradient accumulation round & \multicolumn{4}{c}{64} \\
                                & Max memory in GB allocated & 45.7 & 45.7 & 45.0  & 45.8 (no EMA)  \\
                                & Max memory in GB reserved & 45.7 & 45.7 & 45.9 & 46.0 (no EMA) \\
                                & Time in seconds per 1k images & 365 & 365 & 366 & 502  \\
                                & Time in hours per 1M images & 102 & 102 & 102  & 139  \\

\midrule
\multirow{8}{*}{H100 (80G), FP16} & xFormers  available and enabled %
& \multicolumn{4}{c}{No} \\
& Batch size per GPU & \multicolumn{4}{c}{4} \\
& \# of GPUs  & \multicolumn{4}{c}{8} \\
& \# of gradient accumulation round & \multicolumn{4}{c}{16} \\
& Max memory in GB allocated & 55.8 & 55.8 & 47.7& 63.9  \\
& Max memory in GB reserved & 57.3 & 58.3 & 49.2 & 65.4 \\
& Time in seconds per 1k images & 17 & 17 & 18  & 23  \\
& Time in hours per 1M images & 5 & 5 & 5 & 6  \\
\midrule
\multirow{8}{*}{H100 (80G), FP16} & {Flash Attention available and enabled} & \multicolumn{4}{c}{ Yes} \\
&  Batch size per GPU & \multicolumn{4}{c}{ 4} \\
&  \# of GPUs  & \multicolumn{4}{c}{ 8} \\
&  \# of gradient accumulation round & \multicolumn{4}{c}{ 16} \\
&  Max memory in GB allocated &   32.3 &   32.2 &  29.2&  35.2  \\
&  Max memory in GB reserved &   32.6 &   32.4 &  29.6 &  35.4 \\
&  Time in seconds per 1k images &   12 &   12 &  12  &  15  \\
&  Time in hours per 1M images &   3 &   3 &   3 &  4  \\
\midrule
\multirow{6}{*}{H100 (80G), FP32} & xFormers  available and enabled & \multicolumn{4}{c}{No} \\
& Batch size per GPU & \multicolumn{4}{c}{1} \\
& \# of GPUs  & \multicolumn{4}{c}{8} \\
& \# of gradient accumulation round & \multicolumn{4}{c}{64} \\
& Max memory in GB allocated & 58.9 & 57.4 & 53.4  & 62.9  \\
& Max memory in GB reserved & 60.0 & 59.0 & 54.0 & 64.0  \\
& Time in seconds per 1k images & 74 & 74 & 76  & 90  \\
& Time in hours per 1M images & 21 & 21 & 21 & 25  \\
\bottomrule
\end{tabular}
}
\vspace{-7mm}
\end{center}
\end{table}

\begin{figure}[t]
\centering
\includegraphics[width=.55\linewidth]{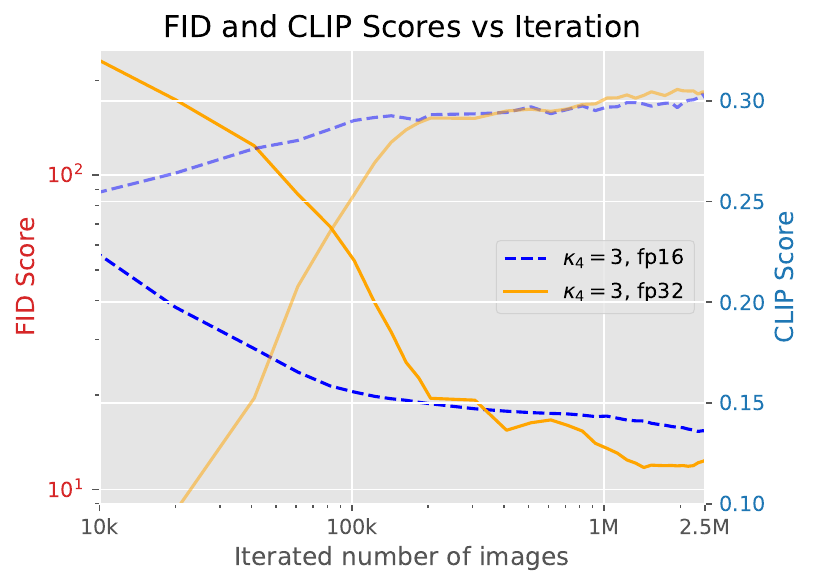}
\includegraphics[width=.55\linewidth]{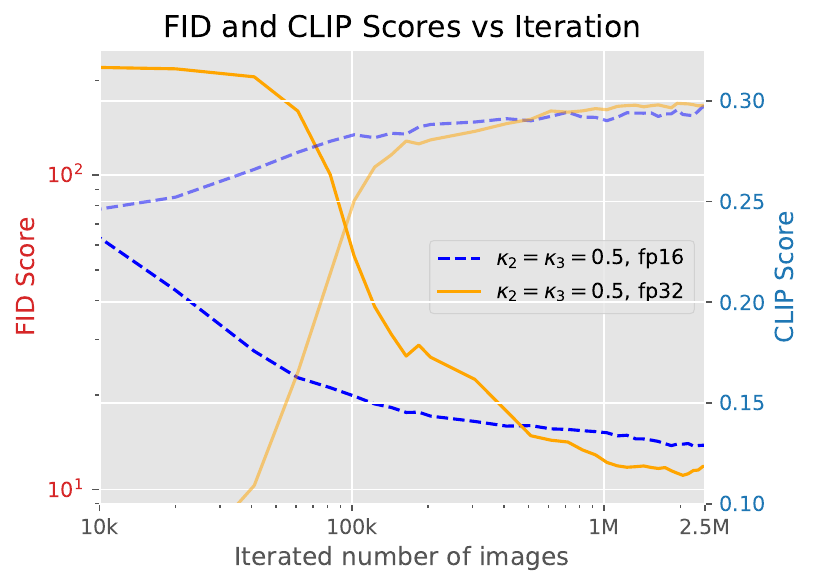}
\includegraphics[width=.55\linewidth]{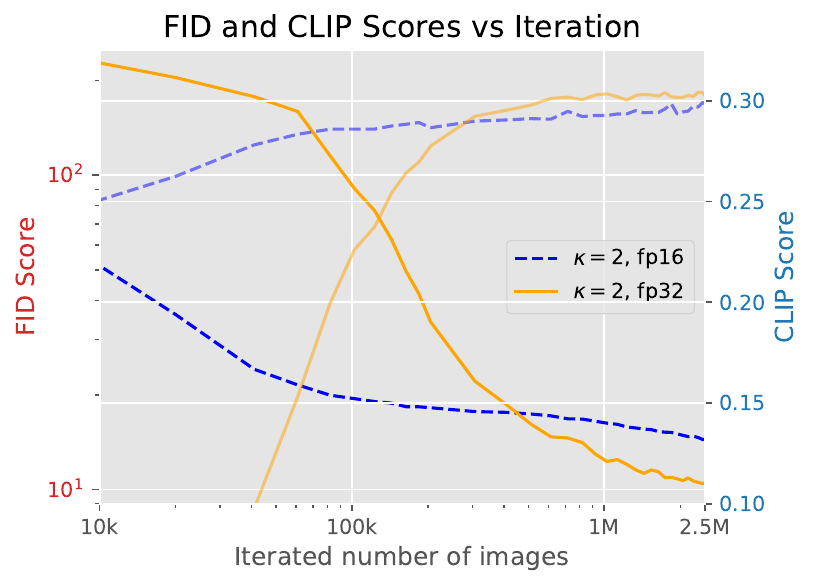}
\caption{\small Comparison between FP16 and FP32 under three different guidance strategies. Top: Long strategy with $\kappa_1=\kappa_2=\kappa_3=1$ and $\kappa_4=3$. Middle: Short strategy with $\kappa_1=\kappa_4=1$ and $\kappa_2=\kappa_3=0.5$. Bottom: Long-short guidance (LSG) with $\kappa_1=\kappa_2=\kappa_3=\kappa_4=2$.
 }
 \label{fig:fp16}
\end{figure}

\section{Limitations and computational requirements} \label{sec:3.3}

\textbf{Memory and speed. } Table~\ref{tab:Hyperparameters} in Appendix~\ref{sec:detail} offers a detailed examination of the computational resources required for the SiD distillation process employing various CFG strategies—Long, Short, ``simplest'' Long-Short, and the recommended Long-Short—across different NVIDIA GPU platforms (RTX-A5000 with 24GB, RTX-A6000 with 48GB, and H100 with 80GB). Key observations include:

SiD-LSG can be effectively operated on the RTX-A5000, which has 24GB of memory, by enabling xFormers \citep{xFormers2022} and switching to FP16 for model and gradient precision. On the RTX-A6000 with 48GB of memory, it runs using FP32 with xFormers enabled. However, on our available H100 with 80GB of memory, where xFormers were not supported at the time of our experiments, there was a noticeable increase in GPU memory consumption.

The recommended LSG strategy, which provides an improved balance between FID and CLIP scores, demands approximately 20\% more computation time per iteration and exhibits about 10\% more peak memory usage on the H100 compared to the Long or Short strategies. This underlines a trade-off between achieving performance enhancements and managing resource utilization.

\textbf{FP16 versus FP32. }
SiD-LSG can be trained under FP16 mixed precision, which significantly conserves memory and enhances processing speed, as detailed in Table~\ref{tab:Hyperparameters}. Although this reduced precision in optimization leads to quicker improvements in both FID and CLIP scores, it also restricts the potential for achieving the lowest FID and highest CLIP scores compared to results under FP32. These effects are demonstrated in the ablation study shown in Figure\,\ref{fig:fp16}. 
Additionally, FP16 operation is less stable and requires the use of gradient clipping to maintain training stability. For instance, we employ \verb|torch.nn.utils.clip_grad_value_(G.parameters(), 1)| to prevent sudden model divergence—a precaution that is not necessary with FP32.

Further investigation is needed to optimize FP16 performance to match that of FP32. This may involve refining loss scaling techniques, updating packages like Flash Attention or XFormers, or implementing an effective warmup period with FP16 before transitioning to FP32. We plan to address these challenges in our future studies.

{We note that our initial experimental platform did not provide proper support for FlashAttention \citep{dao2022flashattention,dao2023flashattention}, which adversely affected our FP16 results. Now that we have established proper support for FlashAttention, we are keen to further explore the potential of FP16 in distilling SiD-LSG, especially to determine if it can match FP32's performance at a lower cost. We have updated Table 4 displayed above to reflect significant memory reductions and a noticeable acceleration in iteration speed under FP16, facilitated by the availability of FlashAttention. This enhancement would enable us to use larger batch sizes per GPU under FP16, further improving time efficiency.
}

\textbf{Reaching a performance plateau.} \citet{zhou2024score} demonstrate through extensive comparisons that the SiD distilled one-step generator can reduce the FID at an exponential decay rate, showing a log-log linear relationship between the number of iterations and FID. This approach can match or even surpass the performance of both unconditional and label-conditional teacher diffusion models trained under the EDM framework \citep{karras2022elucidating}, provided sufficient training.

However, results shown in Figures~\ref{fig:lsg} and \ref{fig:ablation} indicate that the SiD-LSG distilled one-step generator on SD1.5 quickly reaches a performance plateau in reducing FID and/or increasing CLIP scores, particularly when the LSG scale exceeds 2. Additionally, data from Table~\ref{tab:comparison} demonstrate that after processing 10M images (approximately 20k iterations at a batch size of 512), SiD-LSG still does not match the text-image alignment performance of the teacher model, which achieves higher CLIP scores after 50 generation steps. 
Notably, by reducing the LSG scale to 1.5 and doubling the training duration, SiD-LSG achieves a record-low data-free FID of 8.15, establishing a new benchmark among all data-free diffusion distillation methods. This achievement also surpasses the teacher model's FID of 8.78, which was obtained with a CFG scale of 3 and 50 generation steps. However, this reduction in the LSG scale to 1.5 also results in a significant decline in its CLIP score.

This suggests significant potential for further performance enhancements, possibly by extending beyond single-step generation, increasing the model size, or incorporating real data and additional regression or adversarial losses into the distillation process. These avenues for improvement will be one of the focuses of future studies.

\begin{figure}[!t]
\centering
\includegraphics[width=.7\linewidth]{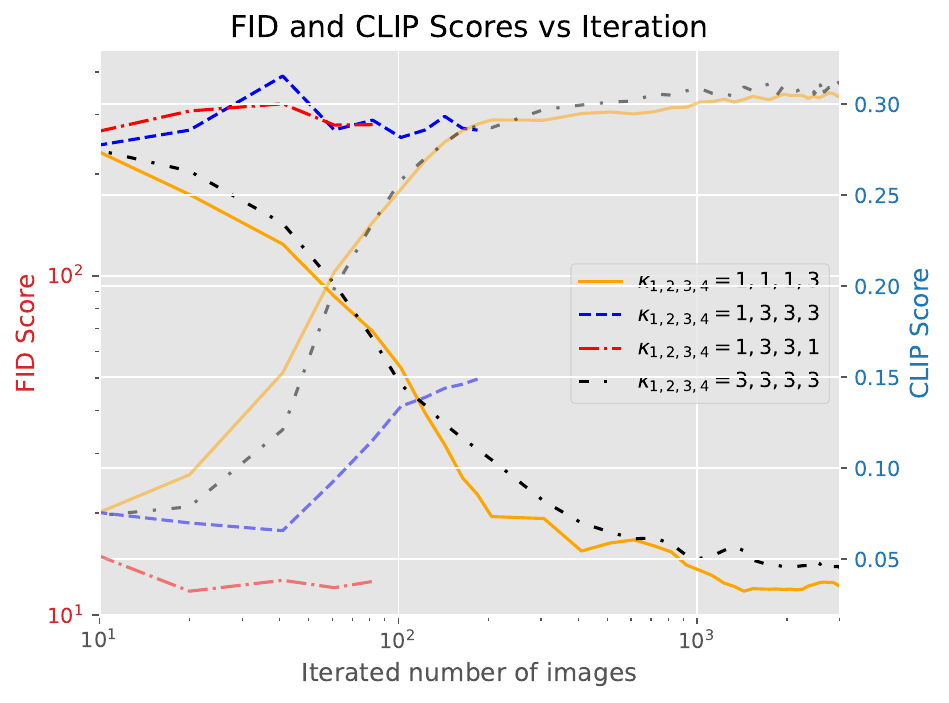}
 \vspace{-2mm}
\caption{\small  
This figure shows the progression of FID and CLIP scores during an ablation study of distilling SD1.5 using SiD-LSG, featuring four different sets of $\kappa_{1,2,3,4}$ values. Settings that converged too slowly were terminated early. %
 }
 \label{fig:ablation_kappa}
  \vspace{-3mm}
\end{figure}

{
\textbf{Additional ablation study of the guidance strategies. } At the beginning of our research, we evaluated various combinations of $\kappa_{1,2,3,4}$, quickly discarding those that converged too slowly or diverged. Specifically, the setting $\kappa_1=1$, $\kappa_2=\kappa_3>1$, and $\kappa_4 \in \{\kappa_2,1\}$ was eliminated early due to its suboptimal performance, as depicted in Figure \ref{fig:ablation_kappa}. }

\section{Prompts and additional example images}\label{sec:prompts}
We list the prompts used in Figure \ref{fig:teaser} as follows:
\begin{enumerate}
    \item A distinguished older gentleman in a vintage study, surrounded by books and dim lighting, his face marked by wisdom and time. 8K, hyper-realistic, cinematic, post-production.
    \item saharian landscape at sunset , 4k ultra realism, BY Anton Gorlin, trending on artstation, sharp focus, studio photo, intricate details, highly detailed, by greg rutkowski.
    \item chinese red blouse, in the style of dreamy and romantic compositions, floral explosions --ar 24:37 --stylize 750 --v 6
    \item Digital 2D, Miyazaki's style, ultimate detailed, tiny finnest details, futuristic, sci-fi, magical dreamy landscape scenery, small cute girl living alone with plushified friendly big tanuki in the gigantism of wilderness, intricate round futuristic simple multilayered architecture, habitation cabin in the trees, dramatic soft lightning, rule of thirds, cinematic.
    \item poster art for the collection of the asian woman, in the style of gloomy, dark orange and white, dynamic anime, realistic watercolors, nintencore, weathercore, mysterious realism --ar 69:128 --s 750 --niji 5
    \item A fantasy-themed portrait of a female elf with golden hair and violet eyes, her attire shimmering with iridescent colors, set in an enchanted forest. 8K, best quality, fine details.
    \item 'very beautiful girl in bright leggings, white short top, charismatic personality, professional photo, style of jessica drossin, super realistic photo, hyper detail, great attention to skin and eyes, professional photo.
    \item (steampunk atmosphere, a stunning girl with a mecha musume aesthetic, adorned in intricate cyber gogle,) digital art, fractal, 32k UHD high resolution, highres, professional photography, intricate details, masterpiece, perfect anatomy, cinematic angle , cinematic lighting, (dynamic warrior pose:1)
    \item (Pirate ship sailing into a bioluminescence sea with a galaxy in the sky), epic, 4k, ultra.
    \item tshirt design, colourful, no background, yoda with sun glasses, dancing at a festival, ink splash, 8k.
\end{enumerate}

We list the prompts used in Figure \ref{fig:imagenet_progress}, which are taken from the COCO-2014 validation set,  as follows:
\begin{enumerate}
\item many cars stuck in traffic on a high way
\item an old blue car with a surfboard on top', 3. `a sole person sits in the front pew of a large church.
\item a shot of the hollywood sign at santa monica blvd.
\item a bunch of flowers, in front of a forest.
\item there is some sort of vegetables in a bowl
\item a man in a pink shirt stands staring against a green wall.
\item small girl in green shirt holding a slice of pizza to her face
\item two dogs sitting in the back seat of a car looking out the windwo
\end{enumerate}
We list the prompts used in Figure \ref{fig:qualitative} as follows:
\begin{enumerate}
    \item Half-length head portrait of the goddess of autumn with wheat ears on her head, depicted as dreamy and beautiful, by wlop.
    \item Walter White dressed as a medieval-style king.
    \item A serene meadow with a tree, river, bridge, and mountains in the background under a slightly overcast sunrise sky.
    \item A closeup portrait of a gray owl with spreaded wings attacking in cinematic lighting, digital painting by Greg Rutkowski used as album cover art on Artstation.
    \item A hyena fursona sitting in the grass in a savannah at sunset.
    \item A puppy staring through a red sectioned window.
\end{enumerate}

\begin{figure}[!h]
    \centering
    \begin{minipage}[b]{0.24\textwidth}
        \centering
        \includegraphics[width=\textwidth]{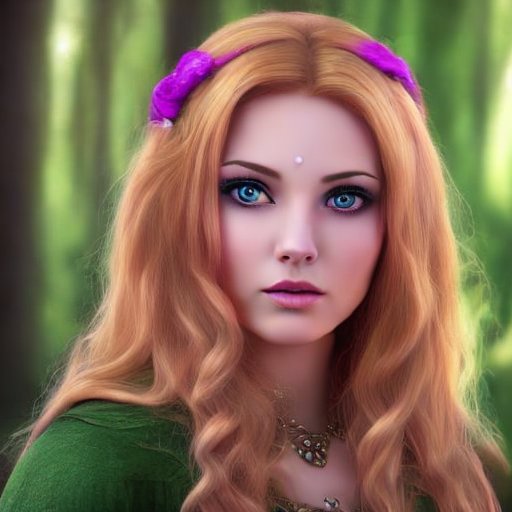}
     \\\vspace{-3mm}\caption*{\tiny A fantasy-themed portrait of a female elf
with golden hair and violet eyes, [...]
}
    \end{minipage}~~
    \begin{minipage}[b]{0.24\textwidth}
        \centering
        \includegraphics[width=\textwidth]{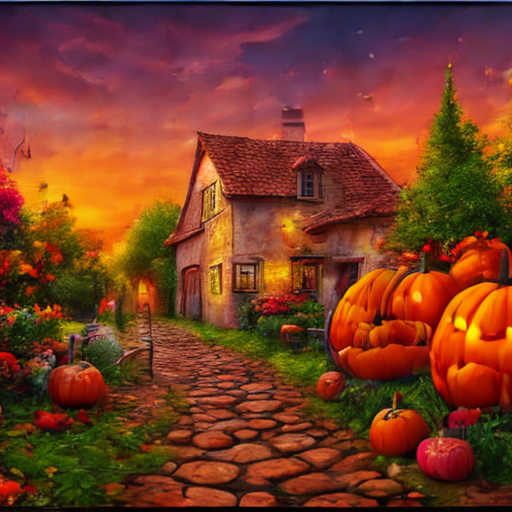}
        \\\vspace{-3mm}\caption*{\tiny pumpkins, autumn sunset in the old village,
cobblestone houses, streets,  [...]
 }
    \end{minipage}~~
    \begin{minipage}[b]{0.24\textwidth}
        \centering
        \includegraphics[width=\textwidth]{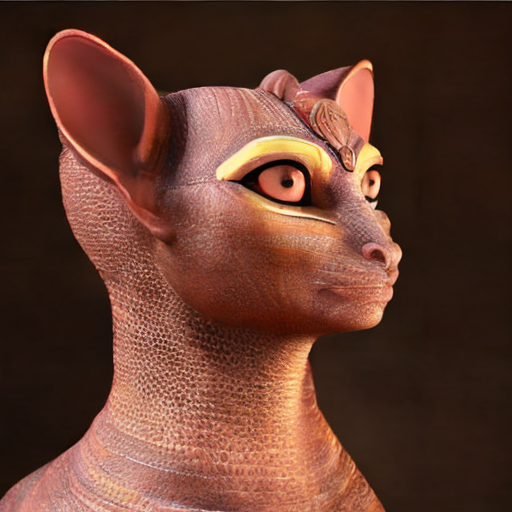}
        \\\vspace{-3mm}\caption*{\tiny Highly detailed mysterious egyptian (sphynx
cat), skindentation:1.2, [...]
}
    \end{minipage}~~
    \begin{minipage}[b]{0.24\textwidth}
        \centering
        \includegraphics[width=\textwidth]{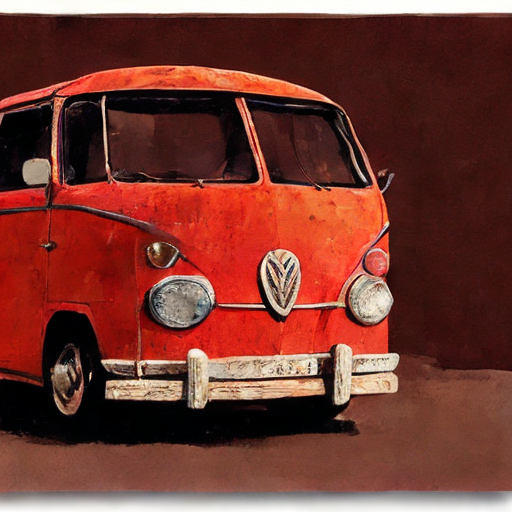}
        \\\vspace{-3mm}\caption*{\tiny vw bus, canvas art, abstract art printing, in the
style of brian mashburn [...]
}
    \end{minipage}~~\\
    \begin{minipage}[b]{0.24\textwidth}
        \centering
        \includegraphics[width=\textwidth]{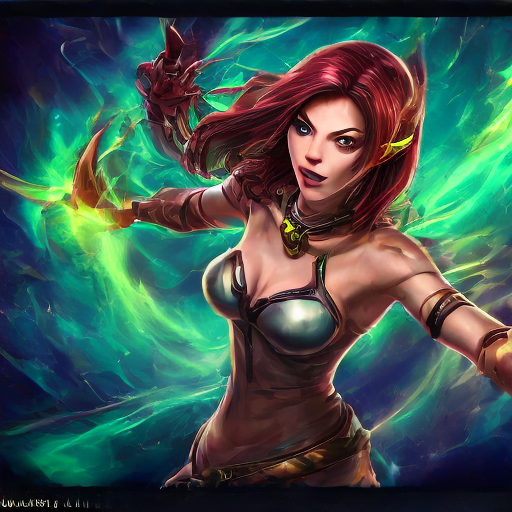}
        \\\vspace{-3mm}\caption*{\tiny painterly style, seductive female League of
legends Jinx character fighting at [...]
}
    \end{minipage}~~
    \begin{minipage}[b]{0.24\textwidth}
        \centering
        \includegraphics[width=\textwidth]{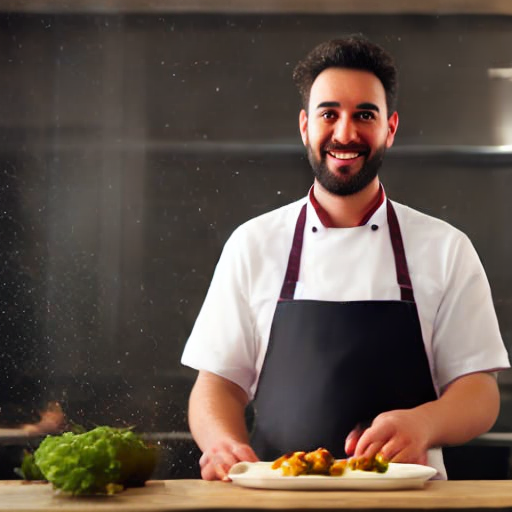}
        \\\vspace{-3mm}\caption*{\tiny A charismatic chef in a bustling kitchen, his
apron dusted with flour, smiling [...]
}
    \end{minipage}~~
    \begin{minipage}[b]{0.24\textwidth}
        \centering
        \includegraphics[width=\textwidth]{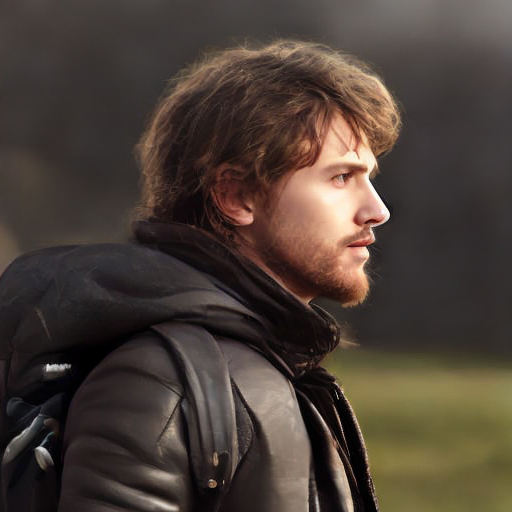}
        \\\vspace{-3mm}\caption*{\tiny A young adventurer with tousled hair and bright
eyes, wearing a leather  [...]
}
    \end{minipage}~~
    \begin{minipage}[b]{0.24\textwidth}
        \centering
        \includegraphics[width=\textwidth]{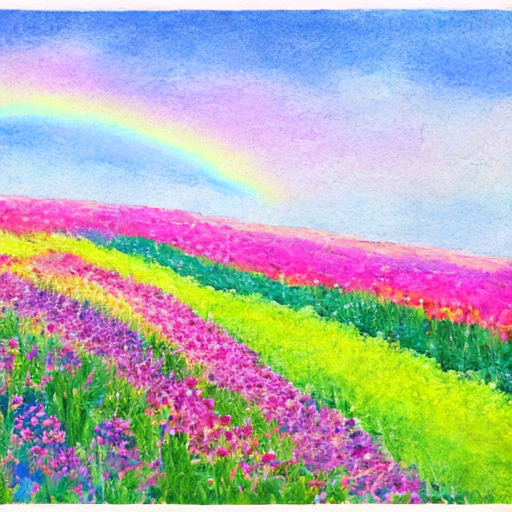}
        \\\vspace{-3mm}\caption*{\tiny A watercolor painting of a vibrant flower field in
spring, with a rainbow of [...]
}
    \end{minipage}~~\\
    \begin{minipage}[b]{0.24\textwidth}
        \centering
        \includegraphics[width=\textwidth]{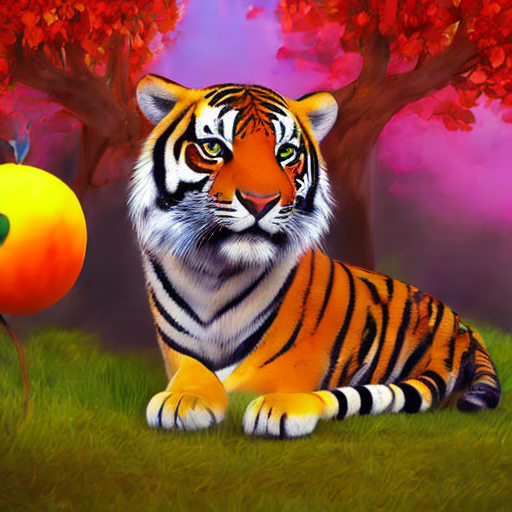}
        \\\vspace{-3mm}\caption*{\tiny digital art of a beautiful tiger pokemon under an
apple tree, cartoon style, [...]
}
    \end{minipage}~~
    \begin{minipage}[b]{0.24\textwidth}
        \centering
        \includegraphics[width=\textwidth]{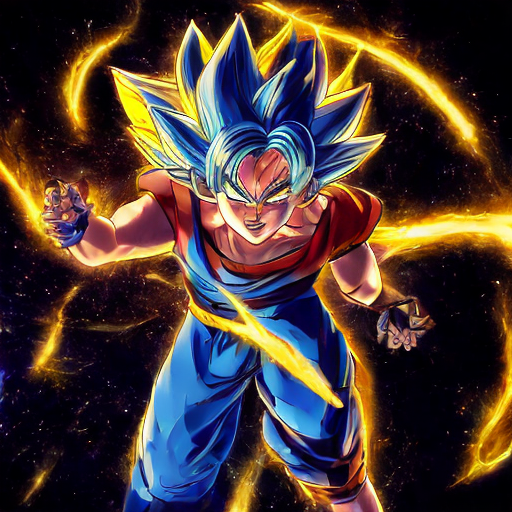}
        \\\vspace{-3mm}\caption*{\tiny painterly style, Goku fighting at war, raging, blue
hair, character design, body [...]
}
    \end{minipage}~~
    \begin{minipage}[b]{0.24\textwidth}
        \centering
        \includegraphics[width=\textwidth]{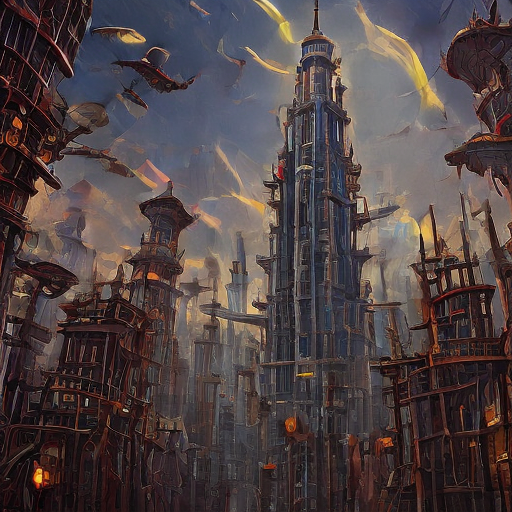}
        \\\vspace{-3mm}\caption*{\tiny A stunning steampunk city with towering
skyscrapers and intricate [...]
}
    \end{minipage}~~
    \begin{minipage}[b]{0.24\textwidth}
        \centering
        \includegraphics[width=\textwidth]{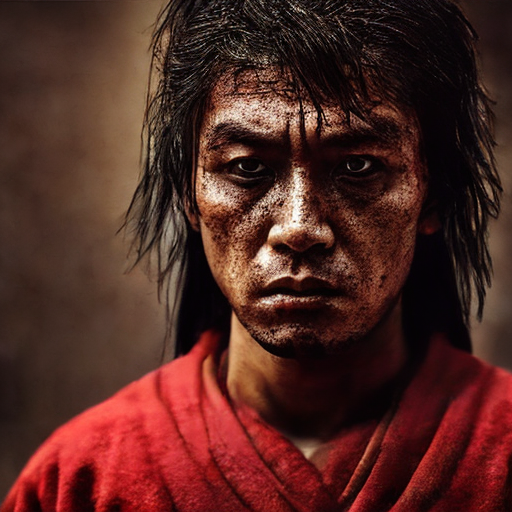}
        \\\vspace{-3mm}\caption*{\tiny Samurai looks at the enemy, stands after the
battle, fear and horror on his [...]
}
    \end{minipage}~~\\
    \begin{minipage}[b]{0.24\textwidth}
        \centering
        \includegraphics[width=\textwidth]{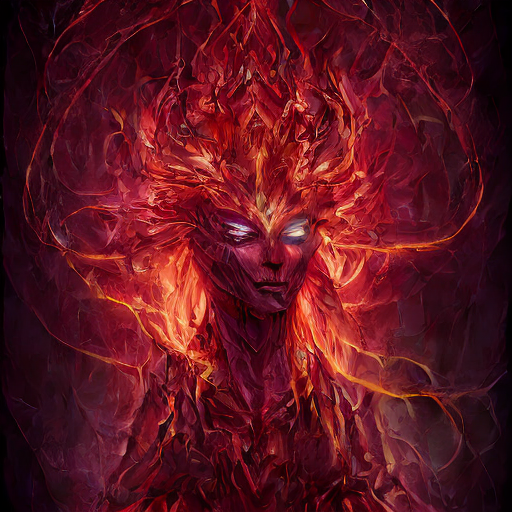}
        \\\vspace{-3mm}\caption*{\tiny A portrait of an elemental entity with strong rim
lighting and intricate  [...]
}
    \end{minipage}~~
    \begin{minipage}[b]{0.24\textwidth}
        \centering
        \includegraphics[width=\textwidth]{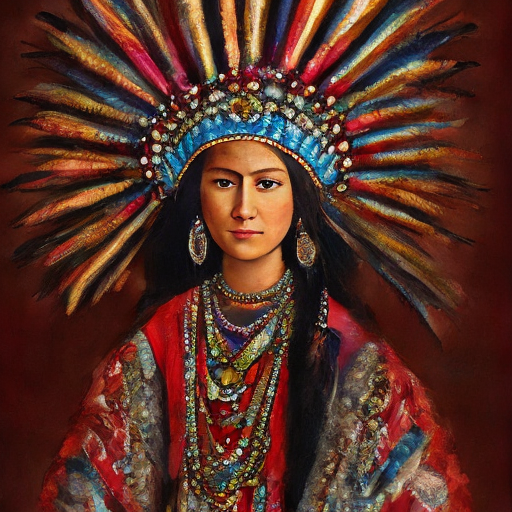}
        \\\vspace{-3mm}\caption*{\tiny A regal female portrait with an ornate headdress
decorated with gemstones[...]
}
    \end{minipage}~~
    \begin{minipage}[b]{0.24\textwidth}
        \centering
        \includegraphics[width=\textwidth]{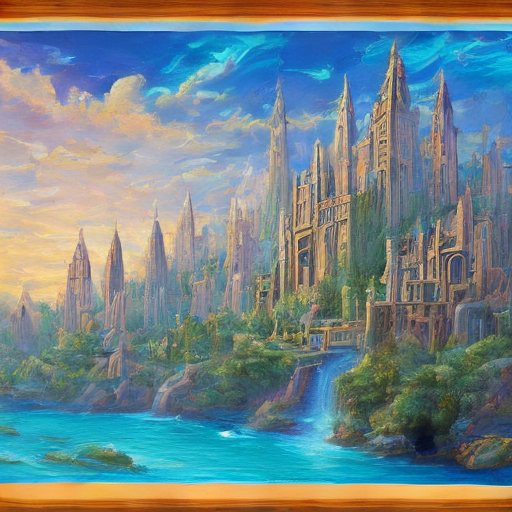}
        \\\vspace{-3mm}\caption*{\tiny A detailed painting of Atlantis by multiple
artists, featuring intricate [...]
}
    \end{minipage}~~
    \begin{minipage}[b]{0.24\textwidth}
        \centering
        \includegraphics[width=\textwidth]{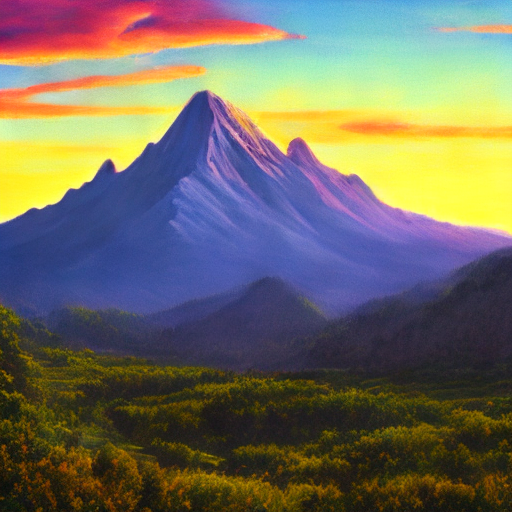}
        \\\vspace{-3mm}\caption*{\tiny \tiny A landscape featuring mountains, a valley,
sunset light, wildlife and a  [...]}
    \end{minipage}~~\\
\caption{More examples from the one-step generator distilled from Stable Diffusion 2.1-base using the proposed method: Score identity Distillation with Long-Short Guidance.}
    \label{fig:more_sid_examples}
\end{figure}

We list the prompts used in Figure \ref{fig:more_sid_examples} as follows:
\begin{enumerate}

    \item A fantasy-themed portrait of a female elf with golden hair and violet eyes, her attire shimmering with iridescent colors, set in an enchanted forest. 8K, best quality, fine details.
    \item pumpkins, autumn sunset in the old village, cobblestone houses, streets, plants, flowers, entrance, realistic, stunningly beautiful
    \item "Highly detailed mysterious egyptian  (sphynx cat), skindentation:1.2, bright eyes,  ancient egypt pyramid background, photorealistic, (hyper-realistic:1.2), cinematic, masterpiece:1.1, cinematic lighting"
    \item "vw bus, canvas art, abstract art printing, in the style of brian mashburn, light red and light brown, theo prins, charming character illustrations, pierre pellegrini, vintage cut-and-paste, rusty debris --ar 73:92 --stylize 750 --v 6"
    \item painterly style, seductive female League of legends Jinx character fighting at war, raging, crazy smile, crazy eyes, rocket lancher, guns, crazy face expression, character design, body is adorned with glowing golden runes, intense green aura around her, body dynamic epic action pose, intricate, highly detailed, epic and dynamic composition, dynamic angle, intricate details, multicolor explosion, blur effect, sharp focus, uhd, hdr, colorful shot, stormy weather, tons of flying debris around her, dark city background, modifier=CarnageStyle, color=blood\_red, intensity=1.6
    \item A charismatic chef in a bustling kitchen, his apron dusted with flour, smiling as he presents a beautifully prepared dish. 8K, hyper-realistic, cinematic, post-production.

    \item A young adventurer with tousled hair and bright eyes, wearing a leather jacket and a backpack, ready to explore distant lands. 8K, hyper-realistic, cinematic, post-production.

    \item "A watercolor painting of a vibrant flower field in spring, with a rainbow of blossoms under a bright blue sky. 8K, best quality, fine details.",

    \item "digital art of a beautiful tiger pokemon under an apple tree, cartoon style,Matte Painting,Magic Realism,Bright colors,hyper quality,high detail,high resolution, --video --s 750 --v 6.0 --ar 1:2" 
    \item "painterly style, Goku fighting at war, raging, blue hair, character design, body is adorned with glowing golden runes, yellow aura around him, body dynamic epic action pose, intricate, highly detailed, epic and dynamic composition, dynamic angle, intricate details, multicolor explosion, blur effect, sharp focus, uhd, hdr, colorful shot, stormy weather, tons of flying debris around him, dark city background, modifier=CarnageStyle, color=blood\_red, intensity=1.6"
    \item A stunning steampunk city with towering skyscrapers and intricate clockwork mechanisms, gears and pistons move in a complex symphony, steam billows from chimneys, airships navigate the bustling skylanes, a vibrant metropolis
    \item "Samurai looks at the enemy, stands after the battle, fear and horror on his face, tired and beaten, sand on his face mixed with sweat, an atmosphere of darkness and horror, hyper realistic photo, In post - production, enhance the details, sharpness, and contrast to achieve the hyper - realistic effect"
    \item A portrait of an elemental entity with strong rim lighting and intricate details, painted digitally by Alvaro Castagnet, Peter Mohrbacher, and Dan Mumford 
    \item "A regal female portrait with an ornate headdress decorated with colorful gemstones and feathers, her robes rich with intricate designs and bright hues. 8K, best quality, fine details.",
    \item "A detailed painting of Atlantis by multiple artists, featuring intricate detailing and vibrant colors.",
    \item "A landscape featuring mountains, a valley, sunset light, wildlife and a gorilla, reminiscent of Bob Ross's artwork.
\end{enumerate}

\end{document}